\documentclass[twoside,11pt]{article}

\usepackage{jmlr2e}
\usepackage{bbm}
\usepackage{amsmath}
\usepackage{enumerate}
\usepackage{amssymb}
\usepackage{mathtools}
\usepackage{cite}
\usepackage{algorithm}
\usepackage{color,xcolor}
\usepackage{algorithmic}
\usepackage{tikz}
\usepackage{graphics,graphicx}
\usepackage{subfigure,epic}
\usepackage{bm}

\usepackage{multirow}
\usepackage{array}
\usepackage{makecell}
\usepackage{booktabs}

\usetikzlibrary{arrows,shapes,chains}

\newtheorem{assumption}{Assumption}
\usepackage{color,xcolor}

\usepackage{lastpage}
\jmlrheading{?}{2019}{1-\pageref{LastPage}}{?/?; Revised
?/?}{?/?}{?-0?}{Shao-Bo Lin, Di Wang and Ding-Xuan Zhou}
\ShortHeadings{Distributed Kernel Ridge Regression with Communications}{Lin, Wang and Zhou}
\firstpageno{1}

\begin{document}

\title{Distributed Kernel Ridge Regression with Communications}

\author{\name Shao-Bo Lin \email sblin1983@gmail.com \\
       \addr Center of Intelligent Decision-Making
        and Machine Learning\\
        School of Management\\
       Xi'an Jiaotong University\\
       Xi'an, China
       \AND
       \name Di Wang\thanks{Corresponding author} \email wangdi@wzu.edu.cn \\
       \addr Center of Intelligent Decision-Making
        and Machine Learning\\
        School of Management\\
       Xi'an Jiaotong University\\
       Xi'an, China
       \AND
       \name Ding-Xuan Zhou \email mazhou@cityu.edu.hk \\
       \addr School of Data Science and Department of Mathematics\\
        City University of Hong Kong\\
        Kowloon, Hong Kong, China}

%\author{\name Shao-Bo Lin$^{1,3}$ \email sblin1983@gmail.com\\
%\name Di Wang$^2$\thanks{Corresponding author} \email wangdi@wzu.edu.cn \\
%\name Ding-Xuan Zhou$^3$ \email mazhou@cityu.edu.hk\\
%\addr $1$.School of Management, Xi'an Jiaotong University, Xi'an, China\\
%$2$.Department of Computer Science and Artificial Intelligence, Wenzhou University, Wenzhou, China\\
%$3$.School of Data Science and Department of Mathematics, City University of Hong Kong,  Kowloon, Hong Kong, China}

\editor{}

\maketitle

\begin{abstract}%   <- trailing '%' for backward compatibility of .sty file
This paper focuses on generalization performance analysis for distributed algorithms in the
framework of learning theory.  Taking distributed kernel ridge regression (DKRR) for example,
we succeed in deriving its optimal learning rates in expectation and providing  theoretically
optimal ranges of the number of local processors. Due to the gap between theory and experiments,
we also deduce optimal   learning rates for DKRR in probability to essentially reflect the
generalization performance and limitations of DKRR. Furthermore, we propose a communication
strategy to improve  the learning performance of DKRR and demonstrate the power of communications
in DKRR via both theoretical assessments and numerical experiments.
\end{abstract}

\begin{keywords}
learning theory, distributed learning, kernel ridge regression, communication
\end{keywords}
\section{Introduction}

Commonly in this era, data of huge size are stored in numerous machines and cannot be shared for protecting data privacy. Typical examples include the clinical data in  medicine where  medical data are collected in different hospitals  and  financial data in business where commercial  data are generated in different companies. These distributively stored data bring a new challenge for machine learning in the sense that every one would like to use the other's data but is unwilling to share his  own data. Nonparametric distributed  learning (NDL) \citep{Zhang2015,Lin2017} presents   a preferable approach to conquer this challenge by means of combining  the prediction results from many local processors without sharing individual  data each other.
\begin{figure*}[t]
    \centering
	\subfigure[Flow of training]{\includegraphics[width=6cm,height=6cm]{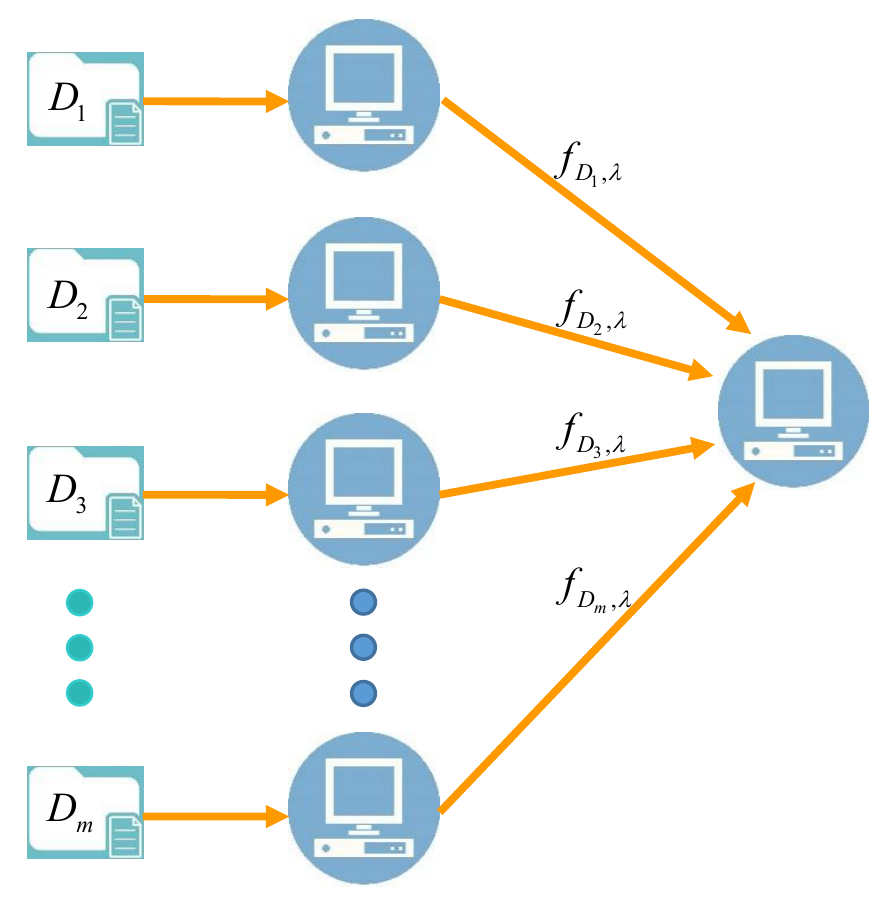}}\hspace{0.3in}
    \subfigure[Flow of testing]{\includegraphics[width=6cm,height=6cm]{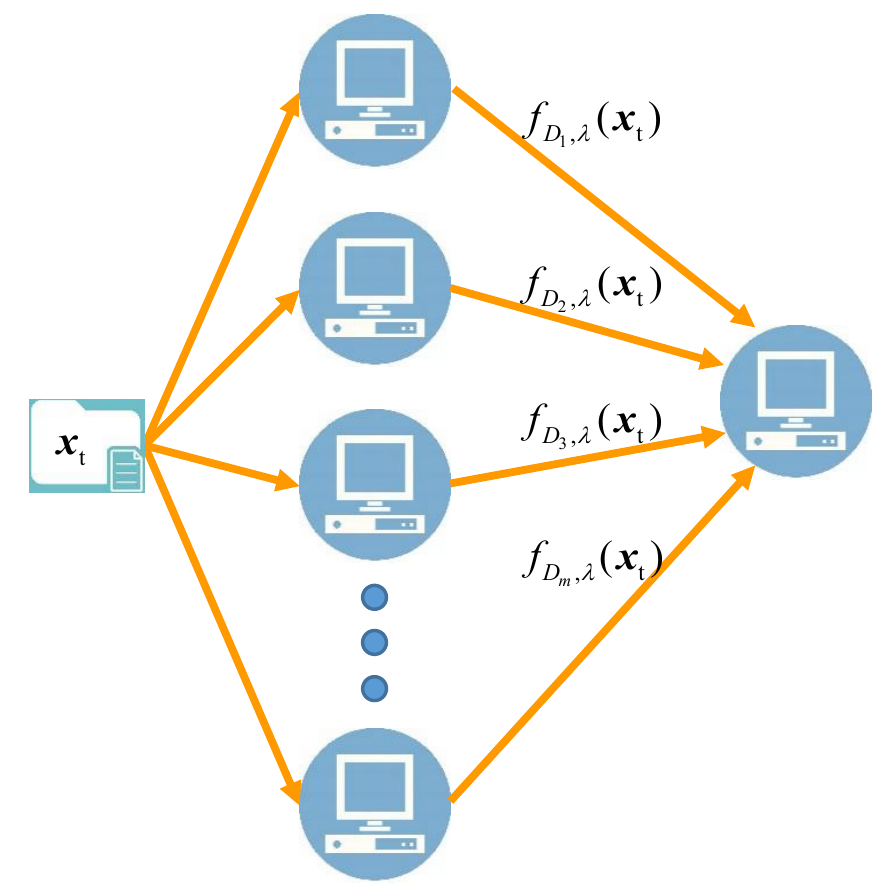}}
	\caption{Training and testing flows of  the divide-and-conquer learning }
\label{Fig:Flow1}
\end{figure*}

There are three ingredients of
NDL: local processing, communication, and synthesization. The local processing issue refers to applying a particular learning algorithm such as the kernel ridgel regression \citep{Zhang2015}, local average regression \citep{Chang2017a}, multi-penalty regularization \citep{Guo2019a}, {  coefficient-based regularization \citep{Pang2018,Shi2018}, and spectral algorithms \citep{Guo2017,Blanchard2018,Linj2018}} to tackle the data subset  in a local  machine  and produce a local estimator. The communication issue focuses on  exchanging  exclusive information  such as the data \citep{Bellet2015}, gradients \citep{Zeng2018} and local estimator \citep{Huang2015} between different local machines. The synthesization issue devotes to producing a global estimator by combining   local estimators  and communicated information on the global machine, typical strategies of which are the majority voting \citep{Mann2009}, weighted average \citep{Chang2017a} and gradient-based algorithms \citep{Bellet2015}.

One of the most popular NDL approaches is    the divide-and-conquer learning, in which the communication  is not required and the weighted average is utilized in the synthesization issue.  Figure  \ref{Fig:Flow1} presents the training and testing flows of  the divide-and-conquer learning.
As shown in  Figure \ref{Fig:Flow1}, to give a prediction of a query point $x_t$, only a real number $f_{D_j,\lambda}(x_t)$  is communicated to the global machine, which succeeds in protecting the data privacy of each local machine.
Generalization performances of  the divide-and-conquer learning   have been  proved to be similar to running the corresponding algorithm processing the whole data on {  a single but large enough machine} \citep{Zhang2015,Chang2017a,Lin2017,Guo2017,Blanchard2018,Linj2018,Pang2018,Shi2018}. The theoretical problem is, however, that there is a strict restriction on   the number of local machines to guarantee the optimal generalization performance, which is difficult to be satisfied in real applications.

In this paper, taking the distributed kernel ridge regression (DKRR) to be the specific algorithm,  we aim at enlarging the number of  local machines by considering  communications among different local machines. There are three purposes in our study. At first, we improve the
existing results for DKRR in expectation in the sense of removing
the eigen-function assumption in \citep{Zhang2015} and relaxing the
regularity assumption  in \citep{Lin2017}.
% In particular, we provide optimal
%learning rates for DKRR when the regression function belongs to the
%reproducing kernel Hilbert space without any eigen-function
%assumption.
Our main tool to achieve this goal is a tight operator product estimate  based on a recently
developed concentration inequality for self-adjoint operators
\citep{Minsker2017}. These estimates improve the results in
\citep{Lin2017,Guo2017}, where the second order decomposition of
operator differences and a classical concentration inequality in
\citep{Pinelis1994} are used.

Since generalization error estimates in probability quantify the generalization performance of DKRR in a single trial while estimates in expectation describe   the average error, it is highly desired to deduce optimal learning rates for DKRR in probability. However,
almost all existing
results for DKRR are established in expectation \citep{Zhang2015,Lin2017,Chang2017}.
The main reason is
that the power of averaging in DKRR can be directly reflected by expectation, provided the samples are assumed to be drawn independently and identically according to some distribution. Our second purpose is to derive optimal learning rates for DKRR in probability, by means of a novel
error decomposition  technique motivated by \citep{Lin2018}.
  Since the advantage of  averaging cannot be directly
utilized, the restriction on the number of local machines is a bit strict. Our estimates in probability support numerical observations that cannot be seen from the estimates in expectation.

Our last purpose is to
develop a communication strategy  to improve the performance of DKRR.  Combining the recently developed integral approach \citep{Lin2017,Guo2017} with a Newton-Raphson iteration, we design a communication strategy to enlarge the number of local machines to guarantee optimal learning rates for DKRR. Our basic idea is to communicate  gradients of each local machine and utilize the Newton-Raphson iteration in the global machine to synthesize the global machine. Both theoretical analysis and numerical results are conducted to verify the power of   communications. Theoretically, we prove that, in the sense of probability, DKRR with communications can reach the optimal learning rates, while the restriction to the number of local machines is the same as that in expectation. Numerically, we exhibit that communications   enlarge the number of local machines of DKRR and thus essentially improve its learning performance.

The rest of the paper is organized as follows. In the next section, we present the communication  strategy as well as its motivation. In Section \ref{Sec.Main Results}, theoretical results including optimal learning rates for DKRR in expectation, optimal learning rates for DKRR in probability and optimal learning rates for DKRR with communications in probability are given.   Section \ref{Sec.Realted-work} makes some comparisons between our results and related work. In Section \ref{Sec.Operator representations}, we provide the main tool in our analysis, where a novel concentration inequality is used to bound the difference between integral operators and their empirical counterparts and some novel error decomposition strategies are adopted to quantify the generalization error. In Section \ref{Sec.Proof}, we prove our theoretical results presented in Section \ref{Sec.Main Results}. In the last section, we conduct a series of numerical studies to verify the outperformance of DKRR with communications.

\section{DKRR with Communications}\label{Sec.Algorithm}
In this section, we propose a novel communication strategy for DKRR to improve the learning performance.

\subsection{Limitations of DKRR without communications}
Let $m$ be the number of local machines,
$D_j=\{(x_{i,j},y_{i,j})\}_{i=1}^{|D_j|}$ be the data subset stored
in the $j$-th local machine  with $1\leq j\leq m$ and
$D=\bigcup_{j=1}^mD_j$ be the disjoint union of $\{D_j\}_{j=1}^m$,
where $|D_j|$ denotes the cardinality of $D_j$. Write
$D_j(x):=\{x:(x,y)\in D_j\}$. Let $({\mathcal H}_K, \|\cdot\|_K)$ be
the reproduced kernel Hilbert space (RKHS) induced by a Mercer
kernel $K$ on a compact metric (input)  space ${\mathcal X}$.  DKRR
  is  defined \citep{Zhang2015} with a
regularization parameter $\lambda>0$ by
\begin{equation}\label{DKRR}
     \overline{f}^0_{D,\lambda}= \sum_{j=1}^m \frac{|D_j|}{|D|} f_{D_j,
       \lambda},
\end{equation}
where
\begin{equation}\label{KRR}
    f_{D,\lambda} =\arg\min_{f\in \mathcal{H}_{K}}
    \left\{\frac{1}{|D|}\sum_{(x, y)\in
    D}(f(x)-y)^2+\lambda\|f\|^2_{K}\right\}.
\end{equation}

\begin{figure*}[t]
    \centering	 \subfigure{\includegraphics[width=8cm,height=6.5cm]{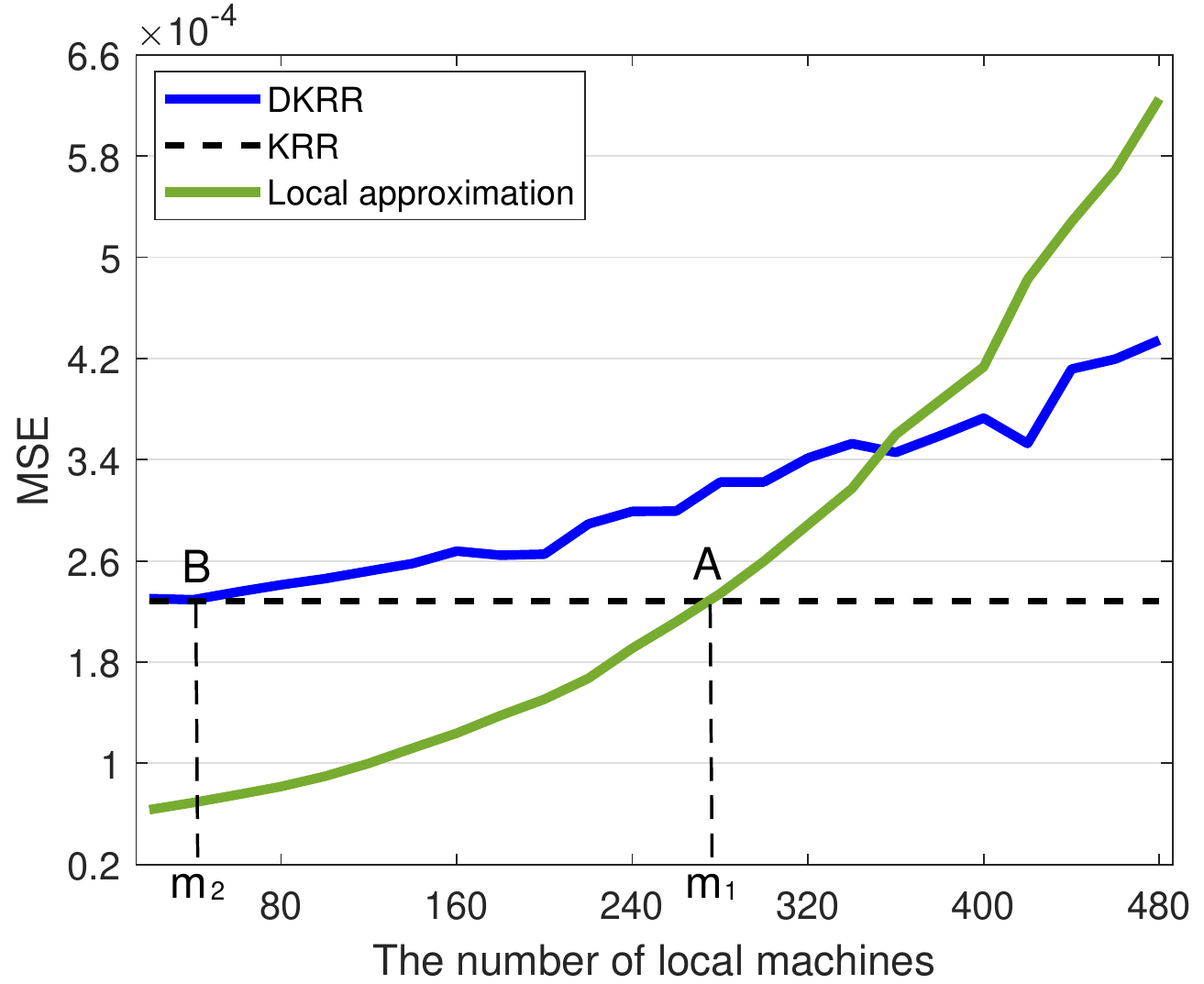}}
	\caption{ The kernel is $K(t,t')=1+\min\{t,t'\}$ for $t,t'\in [0,1]$. Data $\{(t_i,y_i)\}_{i=1}^{10000}$ are generated with $\{t_i\}_{i=1}^{10000}$ being drawn
i.i.d. according to the uniform distribution on $[0,1]$ and $y=f(t)+\varepsilon$, where $f(t)=\min\{t,1-t\}$ and $\varepsilon\sim\mathcal N(0,0.2)$. The green line
(local approximation) denotes the performance of running KRR {on} a local machine {with} a noiseless data subset (of size $10000/m$), $\{(t_i,f(t_i))\}$.} \label{Fig:Motivation}
\end{figure*}

The limitations of DKRR were studied in \citep{Shang2017} and \citep{Liu2018} by presenting a sharp upper bound of $m$ to guarantee the comparable  performances  for DKRR and KRR. Their core idea is that the weighted average in (\ref{DKRR}) cannot improve the approximation ability of KRR in  each local machine.
The representer theorem shows
$$
     f_{D_j,\lambda}\in\mathcal H_{K,D_j}:=\left\{\sum_{x\in D_j(x)}a_x K_x: a_x\in\mathbb R\right\},
$$
where $K_x:=K(x,\cdot)$. Since $\mathcal H_{K,D_j}$ is a $|D_j|$-dimensional linear space, its approximation ability becomes worse   when $m$ increases,  just as the trend  of green line in   Figure \ref{Fig:Motivation} shows. Therefore, it is impossible to derive comparable generalization errors of DKRR (blue line in Figure \ref{Fig:Motivation}) and KRR with whole data (black line in Figure \ref{Fig:Motivation}) when $m$ is larger than $m_1$.  An ideal range of $m$   to guarantee the optimal generalization performance of DKRR is $[1,m_1]$. However, as shown in Figure \ref{Fig:Motivation}, the practical range $[1,m_2]$  is much narrower than $[1,m_1]$. This phenomenon says that the main bottleneck for DKRR is not due to the approximation ability  but the fact that the weighted averaging is not good enough to compensate the loss of samples. Thus, efficient communication strategies and synthesization methods are required to enlarge the range of $m$ to guarantee the best  generalization performance of distributed learning.

\subsection{Motivations from operator representations}

Before presenting our communication strategy, we give the motivation first.  Let $S_{D}:\mathcal H_K\rightarrow\mathbb R^{|D|}$ be the sampling
operator \citep{Smale2007} defined by
$$
         S_{D}f:=(f(x))_{(x,y)\in D}.
$$
 Its scaled adjoint $S_D^T:\mathbb R^{|D|}\rightarrow
\mathcal H_K$  is
given by
$$
       S_{D}^T{\bf c}:=\frac1{|D|}\sum_{i=1}^{|D|}c_iK_{x_i},\qquad {\bf
       c}:=(c_1,c_2,\dots,c_{|D|})^T
       \in\mathbb
       R^{|D|}.
$$
Define
$$
         L_{K,D}f:=S_D^TS_Df=\frac1{|D|}\sum_{(x,y)\in
         D}f(x)K_x.
$$
Then, it can be  found in  {  \citep{Smale2007} and \citep{Lin2017} respectively} that
\begin{equation}\label{operator KRR}
    f_{D,\lambda}=\left(L_{K,D}+\lambda
    I\right)^{-1}S_{D}^Ty_D
\end{equation}
and
\begin{equation}\label{operator DKRR}
    \overline{f}^0_{D,\lambda}=\sum_{j=1}^m\frac{|D_j|}{|D|}
    \left(L_{K,D_j}+\lambda
    I\right)^{-1}S_{D_j}^Ty_{D_j},
\end{equation}
where $y_D:=(y_1,\dots,y_{|D|})^T$.
For an arbitrary $f\in\mathcal H_K$, we have
\begin{equation}\label{NRKRR}
    f_{D,\lambda}= f -\left(L_{K,D}+\lambda
    I\right)^{-1}[\left(L_{K,D}+\lambda
    I\right)f-S_{D}^Ty_D],
\end{equation}
and
\begin{equation}\label{motivation2}
    \overline{f}^0_{D,\lambda}= f -\sum_{j=1}^m\frac{|D_j|}{|D|}\left(L_{K,D_j}+\lambda
    I\right)^{-1}[\left(L_{K,D_j}+\lambda
    I\right)f-S_{D_j}^Ty_{D_j}].
\end{equation}
Since the (half) gradient of the empirical risk in (\ref{KRR}) over $\mathcal H_K$ on $f$ is
\begin{equation}\label{gradient}
   G_{D,\lambda,f}= \frac{1}{|D|}\sum_{(x, y)\in
    D}(f(x)-y)K_x+\lambda f=(L_{K,D}+\lambda
        I)f-S_{D}^Ty_{D}
\end{equation}
and the Hessian is
\begin{equation}\label{Hessian}
   H_{D,\lambda}=\frac{1}{|D|}\sum_{(x, y)\in
    D}\langle \cdot,K_x\rangle_KK_x+\lambda I=L_{K,D}+\lambda I,
\end{equation}
(\ref{NRKRR}) and (\ref{motivation2}) can be regarded as the well known Newton-Raphson iteration.  Comparing (\ref{NRKRR}) with (\ref{motivation2}) and noting that the global  gradients can be achieved via communications, i.e.,
$
    G_{D,\lambda,f}=\sum_{j=1}^m\frac{|D_j|}{|D|}G_{D_j,\lambda,f},
$
 we aim at designing a communication strategy
  via the Newton-Raphson iteration formed as
\begin{equation}\label{operator DKRRwcl}
    \overline{f}^\ell_{D,\lambda}= \overline{f}^{\ell-1}_{D,\lambda} -\sum_{j=1}^m\frac{|D_j|}{|D|}\left(L_{K,D_j}+\lambda
    I\right)^{-1}[\left(L_{K,D}+\lambda
    I\right)\overline{f}^{\ell-1}_{D,\lambda} -S_{D}^Ty_{D}],\qquad \ell\in\mathbb N.
\end{equation}

\subsection{DKRR with communications}

In order to derive an estimator with the operator representation (\ref{operator DKRRwcl}), we propose a communication strategy for DKRR($\ell$)  by iterating the following procedure for $\ell=1,\dots,L$.

Step 1. Communicate the global estimator $\overline{f}^{\ell-1}_{D,\lambda}$  to
local machines and get the local gradient function
$G_{D_j,\lambda,\ell}:=G_{D_j,\lambda,\overline{f}^{\ell-1}_{D,\lambda}}$.

Step 2. Communicate back $\{G_{D_j,\lambda,\ell}:j=1,\dots,m\}$
 to the global machine and synthesize the global
gradient by
$
      G_{D,\lambda,\ell}:=\sum_{j=1}^m\frac{|D_j|}{|D|}G_{D_j,\lambda,\ell}.
$

Step 3. Communicate the gradient function ${G}_{D,\lambda,\ell}$ to each local machine and generate the gradient data $G_{j,\ell}=\{(x,{G}_{D,\lambda,\ell}(x)):x\in D_j(x)\}$. Then
  run  KRR on the data $G_{j,\ell}$ to obtain a function
\begin{equation}\label{GKRR}
    g_{D_j,\lambda,\ell}:=\arg\min_{f\in \mathcal{H}_{K}}
    \left\{\frac{1}{|D_j|}\sum_{(x, y)\in
    G_{j,\ell}}(f(x)-y)^2+\lambda\|f\|^2_{K}\right\}, \qquad j=1, \ldots, m.
\end{equation}

Step 4. Communicate back $g_{D_j,\lambda,\ell}$ to the global machine
and get
\begin{equation}\label{DKRRl}
    \overline{f}^{\ell}_{D,\lambda}=
    \overline{f}^{\ell-1}_{D,\lambda}-
    \frac1\lambda\left[{G}_{D,\lambda,\ell}-\sum_{j=1}^m\frac{|D_j|}{|D|}
    g_{D_j,\lambda,\ell}\right].
\end{equation}

\begin{figure*}[t]
    \centering
	\subfigure[Flow of training]{\includegraphics[width=7.5cm,height=5cm]{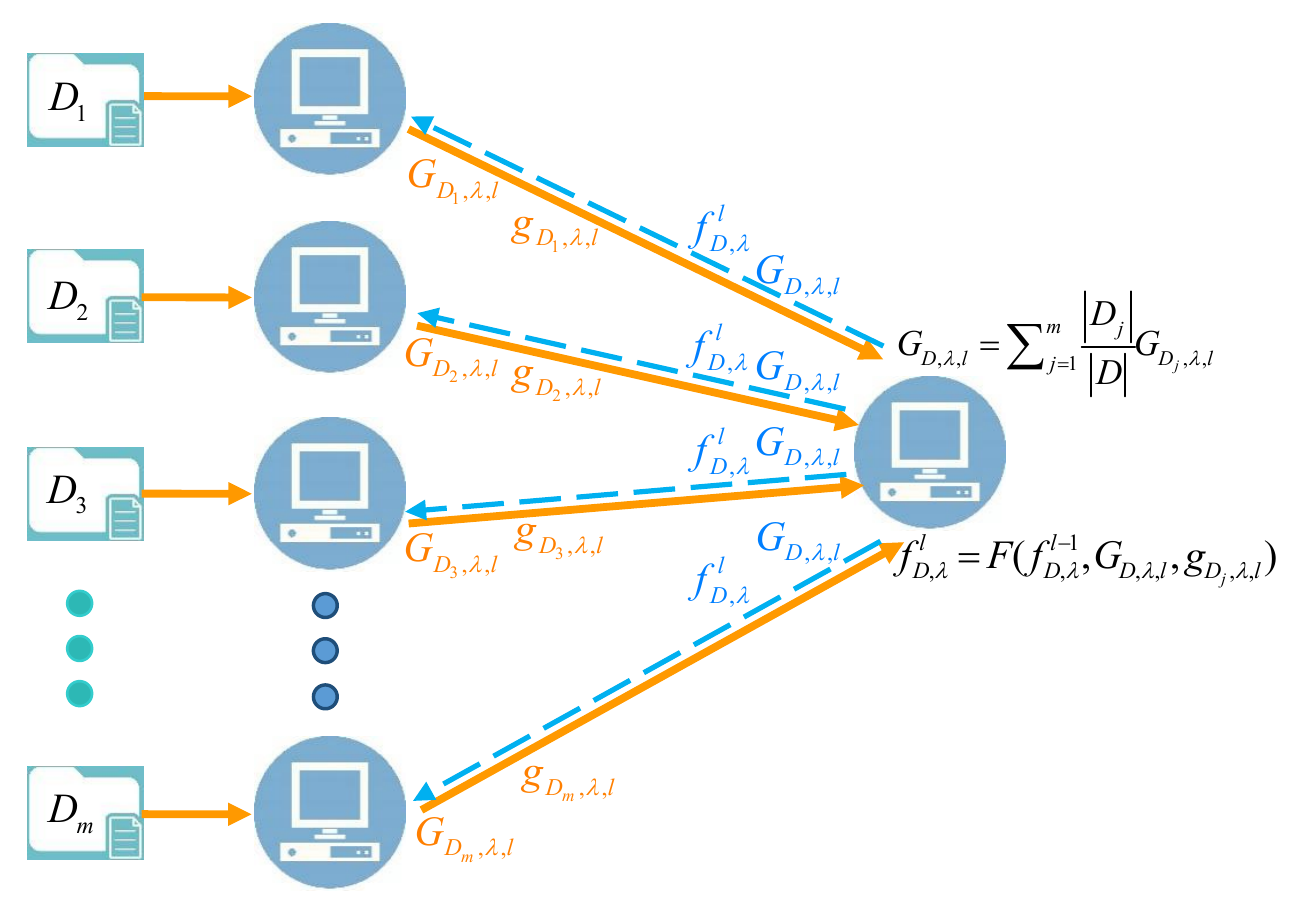}}\hspace{0.1in}
    \subfigure[Flow of testing]{\includegraphics[width=6.5cm,height=5cm]{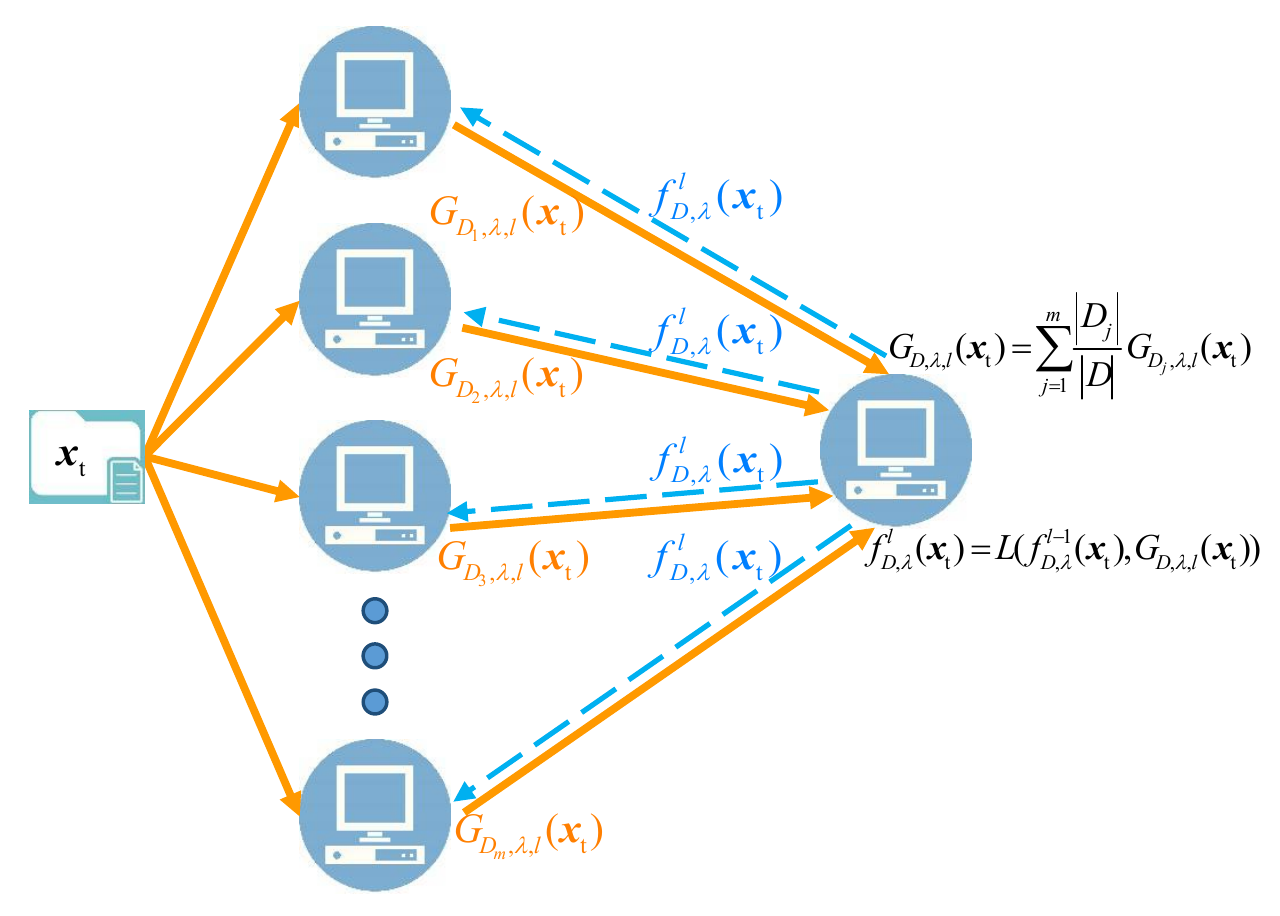}}
	\caption{Training and testing of distributed learning with communications. }\label{Figure:flow-of-c}
\end{figure*}

Due to (\ref{operator KRR}) and (\ref{GKRR}), we have
$$
      g_{D_j,\lambda,\ell}=   (L_{K,D_j}+\lambda I)^{-1}L_{K,D_j}G_{D,\lambda,\ell}.
$$
This together with  the identity
$$
   \frac1\lambda[I-(L_{K,D_j}+\lambda I)^{-1}L_{K,D_j}]G_{D,\lambda,\ell}
   =(L_{K,D_j}+\lambda I)^{-1}G_{D,\lambda,\ell}
$$
and (\ref{DKRRl}) yields (\ref{operator DKRRwcl}). The training and testing flows of  DKRR($\ell$) are exhibited in Figure \ref{Figure:flow-of-c}. Comparing Figure \ref{Fig:Flow1} with Figure \ref{Figure:flow-of-c}, communications are required in both  training and testing stages of DKRR($\ell$). Noticing that communicating functions are infeasible in practice, Appendix B presents a simple realization for DKRR($\ell$) by communicating input data. We believe that there are other efficient implementations of DKRR($\ell$) and leave it as   future studies.

\section{Main Results}\label{Sec.Main Results}

In this section, we analyze the generalization performances of the proposed algorithm as well as DKRR in a standard regression setting
\citep{Cucker2007}.  Let  a sample $D=\{(x_i,y_i)\}_{i=1}^N$ be
independently drawn according to $\rho$,  a Borel probability
measure  on ${\mathcal Z} :={\mathcal X}\times  {\mathcal Y}$ with  $\mathcal Y=\mathbb R$ . The
primary objective is the regression function defined by
$$
         f_\rho(x)=\int_{\mathcal Y} y d\rho(y|x), \qquad x\in\mathcal X,
$$
where $\rho(y|x)$ denotes the conditional distribution at $x$
induced by $\rho$.
Throughout the paper, we assume that $K$ is a Mercer kernel and
$\mathcal X$ is compact, which implies
$\kappa:=\sqrt{\sup_{x\in\mathcal X}K(x,x)}<\infty.$

\subsection{Optimal learning rates for DKRR in expectation}

  To derive   optimal learning rates for
DKRR, we need   some assumptions on the decay of the outputs,
regularity of the regression function and capacity of $\mathcal
H_K$.

\begin{assumption}\label{Assumption:bounded for output}
 We
assume
  $\int_{\mathcal Y}
 y^2d\rho<\infty$ and
\begin{equation}\label{Boundedness for output}
        \int_{\mathcal
        Y}\left(e^{\frac{|y-f_\rho(x)|}M}-\frac{|y-f_\rho(x)|}M-1\right)d\rho(y|x)\leq
        \frac{\gamma^2}{2M^2}, \qquad \forall x\in\mathcal X,
\end{equation}
where $M$ and $\gamma$ are positive constants.
\end{assumption}

   Condition (\ref{Boundedness for output}) is
 satisfied  if the noise is uniformly bounded,
Gaussian or sub-Gaussian \citep{Caponnetto2007}. Let $\rho_X$ be the marginal distribution of $\rho$ and
$L^2_{\rho_{_X}}$ be the Hilbert space of $\rho_X$ square integrable
functions on $\mathcal X$, with norm denoted by $\|\cdot\|_\rho$.
The Mercer kernel $K: {\mathcal X}\times {\mathcal X}
\rightarrow \mathcal R$ defines an integral operator $L_K$ on
${\mathcal H}_K$ (or $L_{\rho_X}^2$) by
$$
         L_Kf =\int_{\mathcal X} K_x f(x)d\rho_X, \qquad f\in {\mathcal
          H}_K \quad (\mbox{or}\ f\in L_{\rho_X}^2).
$$
Our second assumption is the capacity assumption measured by the
effective dimension \citep{Guo2017,Lin2017},
$$
        \mathcal{N}(\lambda)={\rm Tr}((\lambda I+L_K)^{-1}L_K),  \qquad \lambda>0.
$$

\begin{assumption}\label{Assumption:effective dimension}
 There exists some $s\in(0,1]$ such that
\begin{equation}\label{assumption on effect}
      \mathcal N(\lambda)\leq C_0\lambda^{-s},
\end{equation}
where $C_0\geq 1$ is  a constant independent of $\lambda$.
\end{assumption}
{  Condition} (\ref{assumption on
effect}) with $s=1$ is always satisfied by taking the constant
$C_0=\mbox{Tr}(L_K)\leq\kappa^2$. For $0<s<1$,   it was shown in
\citep[Page 7]{Guo2017} that (\ref{assumption on effect}) is slightly
more general than the eigenvalue decaying assumption in the
literature \citep{Caponnetto2007} and has been employed in
\citep{Blanchard2016,Guo2017,Lin2017,Chang2017,Blanchard2018} to
derive optimal learning rates for kernel-based algorithms.

\begin{assumption}\label{Assumption:regularity}
For $r>0$, assume
\begin{equation}\label{regularitycondition}
         f_\rho=L_K^r h_\rho,~~{\rm for~some}  ~ h_\rho\in L_{\rho_X}^2,
\end{equation}
where $L_K^r$ denotes the $r$-th power of $L_K: L_{\rho_X}^2 \to
L_{\rho_X}^2$ as a compact and positive operator.
\end{assumption}

The regularity condition (\ref{regularitycondition}) describes the
regularity of $f_\rho$ and has been adopted in a large literature to
quantify  learning rates for some algorithms
\citep{Smale2007,Caponnetto2007,Blanchard2016,Guo2017,Lin2017}. Based
on the above three assumptions, we can derive the following optimal
learning rates for DKRR in expectation.

\begin{theorem}\label{Theorem:optimal dkrr in expectation}
  Under Assumptions 1-3 with  $\frac12\leq r\leq 1$ and
 $0<s\leq 1$, if $\lambda=|D|^{-\frac{1}{2r+s}}$, $|D_1|=\dots=|D_m|$ and
\begin{equation}\label{Restiction on m for DKRR}
    m\leq C_1|D|^\frac{2r+s-1}{2r+s}(\log|D|)^{-3},
\end{equation}
 then
\begin{equation}\label{optimal krr in p l}
    E[ \|\overline{f}^0_{D,\lambda}-f_\rho\|_\rho^2]\leq C_2
     |D|^{-\frac{2r}{2r+s}},
\end{equation}
where  $C_1,C_2$ are constants independent of $|D|$ or $m$, whose
 values will be given explicitly  in the proof.
\end{theorem}

Theorem \ref{Theorem:optimal dkrr in expectation} exhibits optimal
learning rates for DKRR in expectation under the restriction
(\ref{Restiction on m for DKRR}). In previous studies
\citep{Lin2017,Guo2017,Blanchard2018}, optimal learning rates for
DKRR are built upon the restriction
\begin{equation}\label{Restiction on m for DKRR classical}
    m\leq |D|^\frac{2r-1}{2r+s}.
\end{equation}
A direct consequence is that if $r=1/2$, then {  DKRR with $m=|D|^\theta$ for an arbitrarily small $\theta>0$ may not achieve the optimal learning, according to the existing work. In particular, it was shown in \citep{Lin2017} that a parameter value for $\lambda$ larger than $|D|^{-1/(2r+s)}$ is required under this circumstance, which leads to a sub-optimal learning rate.} Comparing (\ref{Restiction
on m for DKRR classical}) with (\ref{Restiction on m for DKRR}), we
relax the restriction on $m$ so that optimal learning rates for
DKRR also hold  for $r=1/2$.
 The restriction (\ref{Restiction on m
for DKRR}) with $r=1/2$ is similar {  to} that in \citep{Zhang2015}.
However, we remove  the eigenfunction assumption in \citep{Zhang2015}
and derive optimal learning rates for DKRR under Assumption
\ref{Assumption:regularity} with $\frac12\leq r\leq 1$. {  It should be mentioned that removing the eigenfunction assumption in \citep{Zhang2015} was already made in a series of previous papers \citep{Lin2017,Guo2017,Blanchard2018,Linj2018}. However, an additional level of regularity, $r>1/2$ is imposed due to (\ref{Restiction on m for DKRR classical}), excluding $r=1/2$ for \citep{Zhang2015}.  Our study in Theorem \ref{Theorem:optimal dkrr in expectation} successfully fills this gap.}

%For Soblev space $W^{\tau}(B(\mathbb{R}^d))$ with the smoothness index $\tau>d/2$ on a ball $B(\mathbb{R}^d)$
%of the Euclidean space $\mathbb{R}^d$ when the marginal distribution $\rho_X$ is
%the uniform distribution on $B(\mathbb{R}^d)$, it is easy to check \citep{Lin2017} that Assumption  \ref{Assumption:effective dimension}  holds with $s=\frac{d}{2\tau}$. Thus, under  Assumptions 1-3 with  $r=1/2$ and
%$0<s\leq 1$, if $\lambda =|D|^{-\frac{2\tau}{2\tau+d}}$ and $m\leq |D|^\frac{d}{2\tau+d}$, then  the optimal learning rate holds \citep[Theorem 3.2]{Gyorfi2002}:
% \begin{equation}\label{a-e-1}
%    E[ \|\overline{f}^0_{D,\lambda}-f_\rho\|_\rho^2]\leq C_2
%     |D|^{-\frac{2\tau}{2\tau+d}}.
%\end{equation}
%For $m\sim|D|^\frac{d}{2\tau+d}$, the dimension of the assumption space in each local machine is $\mathcal O(|D|^\frac{d}{2\tau+d})$, implying the best approximation error for $f_\rho\in W^{\tau}(B(\mathbb{R}^d)$ is $\mathcal O(|D|^{-\frac{\tau}{2\tau+d}})$ \citep{Pinkus1985}.  Note that the learning rate exhibited in ({a-e-1}) follows an approximation rate of order $|D|^{-\frac{\tau}{2\tau+d}}$ in each local machine and the weighted average cannot improve the approximation ability of local machines \citep{Shang2017}, the derived restrictions on $m$ in (\ref{Restiction on m for DKRR classical}) cannot be improved further.

\subsection{Optimal learning rates for DKRR in probability}
Theorem \ref{Theorem:optimal dkrr in expectation} presented optimal learning rates   for DKRR in expectation. However, the expectation describes the average information for multiple trails and fails to capture the learning performance of DKRR for a single trail.  This explains  the inconsistency between the theoretical result in Theorem \ref{Theorem:optimal dkrr in expectation} and numerical observation in Figure \ref{Fig:Motivation}, where $m_2$ is much smaller than $m_1$.
In the following theorem, we deduce   learning rates for DKRR in
probability.

\begin{theorem}\label{Theorem:optimal dkrr in probability}
Let $0<\delta<1$. Under Assumptions  1-3  with
  $\frac12\leq r\leq 1$ and
$0<s\leq 1$, if $\lambda=|D|^{-\frac{1}{2r+s}}$,
$|D_1|=\dots=|D_m|$,
\begin{equation}\label{Restiction on m for DKRR in prob}
    m\leq\frac{|D|^\frac{2r+s-1}{4r+2s}}{C_3\log^3|D|},
\end{equation}
and
\begin{equation}\label{Restriction on delta 111}
    16|D|^{\frac{2r+s-1}{4r+2s}}\exp\left\{- C_4|D|^{\frac{2r+s-1}{8r+4s}}\log|D|\right\}
    \leq\delta,
\end{equation}
 then with confidence at least
$1-\delta$, there holds
\begin{equation}\label{optimal krr in p}
     \|\overline{f}^0_{D,\lambda}-f_\rho\|_\rho\leq C_5
     |D|^\frac{-r}{2r+s}\log^2\frac{8}\delta,
\end{equation}
where $C_3,C_4,C_5$ are constants independent of $|D|$, $m$ or
$\delta$, whose values will be given explicitly in the proof.
\end{theorem}

It has been a difficult task to derive optimal learning rates for DKRR in probability as shown in  (\ref{optimal krr in p}).  Compared with the  classical error decomposition in expectation \citep{Chang2017}, where the generalization error is decomposed into approximation error, sample error and distributed error, the error decomposition in probability  is totally different. In particular, it is {  not easy to separate a} distributed error in probability to control the number of local machines. As a consequence,  the upper bound of  (\ref{Restiction on m for DKRR in prob}) is  tighter than that of (\ref{Restiction on m for DKRR}), showing a stricter restriction on $m$ to guarantee the optimal learning rate in probability. Neglecting the logarithmic fact, we have $|D|^\frac{2r+s-1}{4r+2s}=\sqrt{|D|^\frac{2r+s-1}{2r+s}}$. Noting that $m_2\sim \sqrt{m_1}$ in Figure \ref{Fig:Motivation}, the error estimate in probability coincides with the numerical results, showing the power of estimate in probability.  Based on the confidence-based error estimate in Theorem \ref{Theorem:optimal dkrr in probability}, we can   derive   almost sure
convergence of DKRR.

\begin{corollary}\label{Corollary: almost surely convergence}
Under Assumption \ref{Assumption:bounded for output}, Assumption
\ref{Assumption:regularity} with $\frac12\leq r\leq 1$ and
Assumption \ref{Assumption:effective dimension}   with $0<s\leq 1$,
if $\lambda=|D|^{-\frac{1}{2r+s}}$, $|D_1|=\dots=|D_m|$ and
(\ref{Restiction on m for DKRR in prob}) holds,
 then for any $\varepsilon>0$, there holds
$$
       \lim_{|D|\rightarrow\infty}|D|^{\frac{r}{2r+s}(1-\varepsilon)}\|\overline{f}_{D,\lambda}^0-f_\rho\|_\rho=0.
$$
\end{corollary}

\subsection{Optimal learning rates for DKRR with communications}

In the previous  subsection, we presented  theoretical limitations of DKRR in terms of a small range of $m$ to guarantee the optimal learning rate in probability. In the following  theorem, we show that our proposed communication strategy can improve the performance of DKRR.

\begin{theorem}\label{Theorem:optimal l}
Let $0<\delta<1$.  Under Assumptions 1-3
  with $\frac12\leq r\leq 1$ and
  $0<s\leq 1$, if
$\lambda=|D|^{-\frac{1}{2r+s}}$, $|D_1|=\dots=|D_m|$,
\begin{equation}\label{Restiction on m for l}
    m\leq\frac{|D|^\frac{(2r+s-1)(\ell+1)}{(2r+s)(\ell+2)}}{C_6\log^3|D|},
\end{equation}
and
\begin{equation}\label{Restriction on delta for l}
     \frac{|D|^{\frac{(2r+s-1)(\ell+1)}{(2r+s)(\ell+2)}}}{C_7\log^3|D|}
     \exp\{-|D|^\frac{2r+s-1}{2(2r+s)(\ell+2)}\log|D|\}\leq\delta<1,
\end{equation}
 then with confidence at least
$1-\delta$, there holds
\begin{equation}\label{optimal for l}
     \|\overline{f}^\ell_{D,\lambda}-f_\rho\|_\rho\leq
     C_8
     |D|^\frac{-r}{2r+s}\log^{\ell+2}\frac{4}\delta,
\end{equation}
where $C_6,C_7,C_8$ are constants independent of $m$, $\delta$ or
$|D|$, whose value will be given  explicitly in the proof.
\end{theorem}

Comparing (\ref{Restiction on m for l}) with (\ref{Restiction on m
for DKRR in prob}), the proposed communication strategy  relaxes
the restriction on $m$ from order $|D|^\frac{2r+s-1}{4r+2s}$  to $|D|^\frac{(2r+s-1)(\ell+1)}{(2r+s)(\ell+2)}$. Furthermore, the upper bound of $m$ is monotonically increasing with the number of communications,   showing the {  power}  of {  communications} in DKRR. As $\ell\rightarrow\infty$,  up to a logarithmic factor, the restriction tends to  the best one in (\ref{Restiction on m for DKRR}).  {  ¡¡At the first glance, the restriction  (\ref{Restiction on m for l}) is always worse than (\ref{Restiction on m for DKRR}), contradicting our assertions on the outperformance of communications.   However, it should be highlighted that (\ref{Restiction on m for DKRR}) only guarantees the error bound in expectation. This means that if $m$ satisfies (\ref{Restiction on m for DKRR}), we cannot conduct feasibility analysis of DKRR via a single (or  finite many) trial.  Figure 2 numerically shows the drawback of analysis in expectation.  Theorem \ref{Theorem:optimal l} conducts the error analysis in probability, a totally different theoretical framework from Theorem \ref{Theorem:optimal dkrr in expectation}. In this framework, Theorem \ref{Theorem:optimal l} shows that communications improve the performance of DKRR since \eqref{Restiction on m for l} is better than \eqref{Restiction on m for DKRR in prob}. It will be shown in Proposition \ref{Proposition: error decomposition for l} below that under \eqref{Restiction on m for l}, the error of DKRR($\ell$) converges exponentially fast with respect to the number of communications, meaning that only a small $\ell$ in DKRR($\ell$) is required to get a satisfactory error bound in probability.
Based on Theorem \ref{Theorem:optimal l},   we  present   almost sure
convergence of DKRR with communications.}

\begin{corollary}\label{Corollary: almost surely convergence l}
Under Assumptions \ref{Assumption:bounded for output}-
\ref{Assumption:regularity} with  $\frac12\leq r\leq 1$   and $0<s\leq 1$,
if $\lambda=|D|^{-\frac{1}{2r+s}}$, $|D_1|=\dots=|D_m|$ and
(\ref{Restiction on m for l}) holds,
 then for
any $\varepsilon>0$, there holds
$$
       \lim_{|D|\rightarrow\infty}|D|^{\frac{r}{2r+s}(1-\varepsilon)}\|\overline{f}_{D,\lambda}^\ell-f_\rho\|_\rho=0.
$$
\end{corollary}

\section{Related Work and Discussions}\label{Sec.Realted-work}
Kernel ridge regression (KRR) is a classical learning algorithm for regression and has been extensively studied in statistics and learning theory. Optimal learning rates for KRR were   established in \citep{Caponnetto2007,SteinwartHS,Lin2017,Lin2019}.
For DKRR,   optimal learning rates were deduced in \citep{Zhang2015} under Assumption  1, Assumption 3  with $r=1/2$,  some eigenvalue decaying assumption that is similar to Assumption 2, and an additional boundedness assumption for the eigenfunctions. In our   paper \citep{Lin2017}, we removed the eigenfunction  assumption by introducing a novel integral operator approach as well as a second-order decomposition for operator difference \citep{Lin2017}. However,   optimal learning rates for DKRR \citep{Lin2017} were derived under Assumption 3 with $\frac12<r\leq 1$, excluding the most popular case $r=1/2$, i.e. $f_\rho\in\mathcal H_K$.  Although several recent work \citep{Guo2017,Chang2017,Blanchard2018,Linj2018,Shi2018} focused on conquering this theoretical drawback, there is  no essential improvement in presenting a good bound for  the number of local machines according to the theory of \citep{Shang2017,Liu2018}. In this paper, we succeed in deriving a tight bound for the number of local machines as (\ref{Restiction on m for DKRR}) by applying the  concentration inequality established in \citep{Minsker2017} to describe the similarity   of different operators (see the next section for detailed descriptions). Different from \citep{Lin2017},  Theorem \ref{Theorem:optimal dkrr in expectation} in this paper removes the eigenfunction assumption of \citep{Zhang2015} without presenting additional regularity assumption.

Previous optimal learning rates for DKRR \citep{Zhang2015,Lin2017,Guo2017,Chang2017,Blanchard2018,Linj2018,Shi2018} were built in  expectation. Technically, the generalization error in expectation can be divided into the   approximation
error, sample error and distributed error \citep{Chang2017} by using the unbiasedness property $E[y|x]=f_\rho(x)$.
 The approximation error,
independent of the sample, describes the approximation capability of the hypothesis space.
 The sample error connects the synthesized estimator
(\ref{DKRR}) with the estimator (\ref{KRR}) by showing an additional
$\frac{|D_j|}{|D|}$ in the sample error for local estimators.
The distributed error   measures  the  limitation of the distributed
learning algorithm (\ref{DKRR}) and presents the upper bound of $m$ to guarantee optimal learning rates for DKRR.   Our error estimate in expectation also follows from this classical error decomposition (see  Lemma \ref{Lemma: dis error decomposition} below). However,  this widely used   error decomposition is  not applicable to   DKRR in probability  since there lacks an expectation operator to realize the  unbiasedness $E[y|x]=f_\rho(x)$.
Thus, it requires novel approaches to deduce  optimal  learning rates for DKRR in probability. According to an explicit operator representation for the kernel-based gradient descent algorithm, optimal learning rates as well as a novel error decomposition based on its operator representation for distributed gradient descent algorithms were established in probability in \citep{Lin2018}. Using the similar error decomposition as \citep{Lin2018}, a   minimum error entropy algorithm with   distributed gradient descents was proposed in \citep{Hu2019} and a tight learning rate was derived.  However, DKRR requires the computation  of the inverse matrix (or operator),   the error decomposition of gradient descent algorithm in \citep{Lin2018,Hu2019}  is not suitable for DKRR. In this paper, as shown in Proposition \ref{Proposition:Error decomposition in pro} below, we succeed in deriving a novel error decomposition for DKRR by introducing some measurements to quantify the difference between integral operators and their empirical counterparts. Then, applying the recently developed concentration inequalities for positive operators \citep{Minsker2017}, we derive optimal learning rates for DKRR in probability under much looser restriction on $m$ than \citep{Lin2018,Hu2019}.

Numerous communication strategies \citep{Li2014,Shamir2014,Bellet2015,Huang2015} were proposed to improve the learning performance of distributed learning algorithms in the framework of parametric regression (linear regression). To the best of our knowledge, our proposed communication strategy  is the first work focusing on improving the performance of learning algorithms in nonparametric regression.  As shown in Figure \ref{Fig:Flow1}, nonparametric regression transmits function values and protects the privacy of local machines,  while parametric regression \citep{Zhang2013} transmits coefficients that disclose the detailed information for local estimators. The most related work is \citep{Huang2015}, where a communication strategy based on Newton-Raphson iterations is proposed to equip ridge regression in linear regression. Our work differs from \citep{Huang2015} in the following three aspects. Firstly, our analysis is carried out in nonparametric regression rather than linear regression. Secondly, the communication strategy is based on the operator representation, which is exclusive for kernel approaches. Thirdly, our theory focuses on enlarging the range of number of local machines rather than improving the learning rate  of distributed learning algorithms, since DKRR without communications is already optimal for not so large $m$ based on previous studies \citep{Lin2017}.
It would be interesting to extend our communication strategy to other distributed learning schemes such as distributed learning with convolutional neural networks in deep learning \citep{Zhou2018Deep,Zhou2018Uni,Zhou2018Dis}.

\section{Operator Similarities and Error Decomposition}\label{Sec.Operator representations}
We analyze the learning performance of DKRR($\ell$) by using the
integral operator approach presented in
 \citep{Smale2007,Lin2017,Guo2017,Guo2018}. Our novelty in analysis
is   tight bounds on quantifying the similarity between different
operator. These bounds together with the exclusive error
decomposition yield   optimal learning rates for DKRR and show the
advantage of communications in distributed learning.

\subsection{Similarities of operators}

The similarity between $\overline{f}^{\ell}_{D,\lambda}$ and
$f_{D,\lambda}$ depends heavily on  that between the
operator $L_{K}$ and $L_{K,D}$.
The classical method for analyzing similarity between $L_K$ and
$L_{K,D}$ is to bound the norm of operator difference $L_K-L_{K,D}$.
By using a concentration inequality in Hilbert spaces  from
\citep{Pinelis1994}, it can be found in {  \citep{Caponnetto2007,Blanchard2016}} that for any
$\delta\in(0,1),$ with confidence at least $1-\delta,$ there holds
\begin{equation}\label{Define S}
         \mathcal S_{D,\lambda}:=\|(L_K+\lambda I)^{-1/2}(L_K-L_{K,D})\|\leq
         2\kappa(\kappa+1)\mathcal A_{D,\lambda}\log\frac2\delta,
\end{equation}
where
\begin{equation}\label{Def.A}
    \mathcal A_{D,\lambda}:=\frac{1}{\sqrt{|D|}}\left(\frac{1}{\sqrt{|D|\lambda}}
           +\sqrt{\mathcal{N}(\lambda)}\right).
\end{equation}

 The bound in (\ref{Define S}) is tight. However, in estimating
the difference between $f_{D,\lambda}$ and
$\overline{f}^{\ell}_{D,\lambda}$, one also needs to estimate
\begin{equation}\label{Define R}
     \mathcal R_{D,\lambda}:=\|(L_K+\lambda I)^{-1/2}(L_K-L_{K,D})(L_K+\lambda I)^{-1/2}\|.
\end{equation}
 A
classical approach \citep{Lin2017,Guo2017} is to use
$
     \mathcal R_{D,\lambda}
      \leq
         \frac1{\sqrt{\lambda}}\mathcal S_{D,\lambda}
$ and get that
\begin{equation}\label{bound R 1}
     \mathcal R_{D,\lambda}
     \leq
     2\kappa(\kappa+1)\lambda^{-1/2}\mathcal A_{D,\lambda}\log\frac{2}{\delta}
\end{equation}
holds with confidence $1-\delta$. The leading term in (\ref{bound R
1}) is $\frac{\sqrt{\mathcal N(\lambda)}}{\sqrt{|D|\lambda}}$ for
$\lambda\geq |D|^{-1}$. In the following lemma, which will be proved
in Appendix A, we  reduce the leading
term for bounding $\mathcal R_{D,\lambda}$  from
$\frac{\sqrt{\mathcal N(\lambda)}}{\sqrt{|D|\lambda}}$ to
$\frac{\sqrt{\log\mathcal N(\lambda)}}{\sqrt{|D|\lambda}}$ by using a new concentration inequality for
self-adjoint operators \citep{Minsker2017}.

\begin{lemma}\label{Lemma:operator difference}
Let $0<\delta\leq 1$. If $0<\lambda\leq1$ and $\mathcal
N(\lambda)\geq 1$, then
  with confidence $1-\delta,$ there
holds
\begin{equation}\label{operator difference concentration 1}
         \mathcal R_{D,\lambda}
         \leq C_1^*\mathcal B_{D,\lambda}\log\frac4\delta,
\end{equation}
where $C_1^*:=\max\{(\kappa^2+1)/3,2\sqrt{\kappa^2+1}\}$ and
\begin{equation}\label{Def.B}
     \mathcal B_{D,\lambda}:=
           \frac{1+\log\mathcal
      N(\lambda)}{\lambda|D|}+ \sqrt{\frac{1+\log\mathcal
      N(\lambda)}{\lambda|D|}}.
\end{equation}
\end{lemma}

Besides the differences between $\mathcal R_{D,\lambda}$ and $\mathcal
S_{D,\lambda}$, another quantity to measure the similarity between
$L_{K,D}$ and $L_K$ is the operator product $\|(L_K+\lambda
I)^{1/2}(L_{K,D}+\lambda I)^{-1/2}\|$. A recently developed second
order decomposition for positive operators \citep{Lin2017,Guo2017}
  asserts that  if $A$ and $B$ are
invertible operators on a Banach space, then
$$
          A^{-1}-B^{-1}
          =
          B^{-1}(B-A)B^{-1}(B-A)A^{-1}+B^{-1}(B-A)B^{-1}.
$$
This implies the following decomposition of the operator product
\begin{equation}\label{operator product based on second order}
     BA^{-1}
   =(B-A)B^{-1}(B-A)A^{-1}+(B-A)B^{-1}+I.
\end{equation}
Inserting $A=L_{K,D}+\lambda I$ and $B=L_{K}+\lambda I$ to
(\ref{operator product based on second order}) and noting
(\ref{Define S}), it is easy to derive the following upper bound for
$\|(L_K+\lambda I)(L_{K,D}+\lambda I)^{-1}\|$ (e.g., \citep{Guo2017}):
  for any $0<\delta<1,$ with confidence at least
$1-\delta,$ there holds
$$
          \|(L_K+\lambda I)(L_{K,D}+\lambda I)^{-1}\|\le
          2\left[\left(\frac{2\kappa}{\sqrt{|D|\lambda}}\left(\frac{\kappa}{\sqrt{|D|\lambda}}
           +\sqrt{\mathcal{N}(\lambda)}\right)\log\frac{2}{\delta} \right)^2  +1\right].
$$
Hence, according to the {  Cordes inequality \citep{Bathia1997} }
\begin{equation}\label{Codes inequality}
         \|A^\tau B^\tau\|\le\|AB\|^\tau, \qquad 0<\tau\leq 1,
\end{equation}
  we have
\begin{eqnarray}\label{Define Q}
          \mathcal Q_{D,\lambda}
          &:=&\|(L_K+\lambda I)^{1/2}(L_{K,D}+\lambda I)^{-1/2}\|\nonumber\\
          &\leq&
          \sqrt{2\left[\left(\frac{2\kappa}{\sqrt{|D|\lambda}}\left(\frac{\kappa}{\sqrt{|D|\lambda}}
           +\sqrt{\mathcal{N}(\lambda)}\right)\log\frac{2}{\delta} \right)^2  +1\right]}.
\end{eqnarray}
The leading term of the  right-hand side of (\ref{Define Q}) is
$\frac{\sqrt{\mathcal N(\lambda)}}{\sqrt{|D|\lambda}}$.  In the
following lemma, whose proof is postponed to Appendix A,  we improve
(\ref{Define Q}) by using Lemma \ref{Lemma:operator difference}.

\begin{lemma}\label{Lemma:operator product}
Assume   $0<\lambda\leq 1$ and $\mathcal N(\lambda)\geq 1$.
 For $\delta\geq 4\exp\{-1/(2C_1^*\mathcal B_{D,\lambda})\}$ , with
confidence $1-\delta,$ there holds
\begin{equation}\label{operator product}
         \mathcal Q_{D,\lambda}\leq\sqrt{2}.
\end{equation}
\end{lemma}

In (\ref{Define Q}), to guarantee the boundedness of $\mathcal
Q_{D,\lambda}$, it requires $\frac{\sqrt{\mathcal
N(\lambda)}}{\sqrt{|D|\lambda}}\log\frac{2}\delta\leq C_2^* $ for
some $C_2^*>0$ depending only on $\kappa$. However, in Lemma
\ref{Lemma:operator product}, recalling (\ref{Def.B}), it is
sufficient that
  $\frac{\log(\mathcal
N(\lambda))}{\sqrt{|D|\lambda}}\log\frac{2}\delta\leq C_3^*$ for
some $C_3^*>0$ depending only on $\kappa$.

\subsection{Error decomposition for DKRR in expectation}

We use the error decomposition for DKRR in \citep{Chang2017}, where
the data-free limit and noise-free
  version  of $f_{D_j,\lambda}$,
\begin{equation}\label{data free limit}
    f_{\lambda} =\arg\min_{f\in \mathcal{H}_{K}}
    \left\{\int_{\mathcal X}(f(x)-f_\rho(x))^2d\rho_X+\lambda\|f\|^2_{K}\right\}
    =(L_K+\lambda I)^{-1}L_Kf_\rho.
\end{equation}
and
\begin{equation}\label{noise free version}
  f^{\diamond}_{D_j,\lambda}:=E[f_{D_j,\lambda}|D_j(x)]=(L_{K,D_j}+\lambda
  I)^{-1}L_{K,D_j}f_\rho
\end{equation}
  are utilized.
The following lemma can be found in \citep{Chang2017}.

\begin{lemma}\label{Lemma: dis error decomposition}
 Let $\overline{f}^0_{D,\lambda}$ be defined by (\ref{DKRR}).
 We have
 \begin{equation}\label{dis error decomp rho}
        \frac12E\left[\|\overline{f}_{D,\lambda}^0-f_\rho\|_\rho^2\right]
        \leq
         \|f_\lambda-f_\rho\|_\rho^2+
         \sum_{j=1}^m\frac{|D_j|^2}{|D|^2}E\left[\|f_{D_j,\lambda}
         -f_\lambda\|_\rho^2\right]
        +\sum_{j=1}^m\frac{|D_j|}{|D|}
        E\left[\left\|f^{\diamond}_{D_j,\lambda}-f_\lambda\right\|_\rho^2\right].
\end{equation}
\end{lemma}

The three  terms on the  right-hand side of (\ref{dis error decomp
rho})  are respectively the approximation error, sample error and distributed
error. Based on Lemma \ref{Lemma: dis error decomposition}, we can
derive the following error decomposition for DKRR in expectation.

\begin{proposition}\label{Proposition:Error decomposition in ex}
Let $\overline{f}_{D,\lambda}^0$ be defined by (\ref{DKRR}). If
$f_\rho\in\mathcal H_K$, then
\begin{eqnarray}\label{Error decomp for DKRR in expectation}
        \frac12E\left[\|\overline{f}_{D,\lambda}^0-f_\rho\|_\rho^2\right]
        &\leq&
         \|f_\lambda-f_\rho\|_\rho^2+
         \sum_{j=1}^m\frac{|D_j|^2}{|D|^2}E\left[ \mathcal Q_{D_j,\lambda}^4(\mathcal P_{D_j,\lambda}+\mathcal
        S_{D_j,\lambda}\|f_\lambda\|_K) ^2  \right]  \nonumber\\
        &+&\sum_{j=1}^m\frac{|D_j|}{|D|}
        E\left[\mathcal Q_{D_j,\lambda}^4\mathcal R^2_{D_j,\lambda}\|(L_K+\lambda
    I)^{1/2}(f_\lambda-f_\rho)\|_K^2\right],
\end{eqnarray}
where
\begin{equation}\label{Define P}
 \mathcal P_{D,\lambda}:=
         \left\|(L_K+\lambda
          I)^{-1/2}(L_{K}f_\rho-S^T_{D}y_D)\right\|_K.
\end{equation}
\end{proposition}

{\bf Proof.}  We  first bound the sample error. It
follows from (\ref{operator KRR}) and (\ref{data free limit}) that
\begin{eqnarray*}
   &&f_{D,\lambda}-f_\lambda
   =
   (L_{K,D}+\lambda I)^{-1}S_D^Ty_D
   -
   (L_K+\lambda I)^{-1}L_Kf_\rho \\
   &=&
   (L_{K,D}+\lambda I)^{-1}(S^T_Dy_D-L_Kf_\rho)+[(L_{K,D}+\lambda I)^{-1}
   -
   (L_K+\lambda I)^{-1}]L_Kf_\rho\nonumber\\
   &=&
   (L_{K,D}+\lambda I)^{-1/2}(L_{K,D}+\lambda I)^{-1/2}(L_K+\lambda I)^{1/2}(L_K+\lambda
   I)^{-1/2}(S^T_Dy_D-L_Kf_\rho)\nonumber\\
   &+&
   (L_{K,D}+\lambda I)^{-1/2}(L_{K,D}+\lambda I)^{-1/2}(L_K+\lambda
   I)^{1/2}
   (L_K+\lambda
   I)^{-1/2}(L_K-L_{K,D})f_\lambda. \nonumber
\end{eqnarray*}
So
\begin{eqnarray}\label{sample for KRR}
      \|(L_K+\lambda I)^{1/2}(f_{D_j,\lambda}-f_\lambda)\|_K
        \leq
        \mathcal Q_{D_j,\lambda}^2(\mathcal P_{D_j,\lambda}+\mathcal
        S_{D_j,\lambda}\|f_\lambda\|_K).
\end{eqnarray}
Then, we bound the distributed error. Due to
(\ref{operator KRR}) and (\ref{noise free version}), we get
\begin{eqnarray*}
  &&  f^\diamond_{D,\lambda}-f_\lambda
  =
   (L_{K,D}+\lambda I)^{-1}L_{K,D}f_\rho-(L_K+\lambda
   I)^{-1}L_Kf_\rho\\
   &=&
   (L_{K,D}+\lambda I)^{-1}(L_{K,D}-L_{K})f_\rho
  +[(L_{K,D}+\lambda I)^{-1}-(L_{K}+\lambda I)^{-1}]L_Kf_\rho\\
  &=&
  (L_{K,D}+\lambda I)^{-1}(L_{K,D}-L_{K})f_\rho+(L_{K,D}+\lambda I)^{-1}(L_K-L_{K,D})
  f_\lambda \\
  &=&
  (L_{K,D}+\lambda I)^{-1}(L_{K,D}-L_{K})(f_\rho-f_\lambda)\\
  &=&
   (L_{K,D}+\lambda I)^{-1}(L_K+\lambda I)^{1/2} (L_K+\lambda I)^{-1/2}(L_{K,D}-L_{K})(L_K+\lambda I)^{-1/2}\\
   &&(L_K+\lambda I)^{1/2}(f_\rho-f_\lambda).
\end{eqnarray*}
Combining this with $f_\rho\in\mathcal H_K$ yields
\begin{equation}\label{Distributed error estimate}
    \|(L_K+\lambda I)^{1/2}(f^\diamond_{D_j,\lambda}-f_\lambda)\|_K
    \leq \mathcal Q_{D_j,\lambda}^2\mathcal R_{D_j,\lambda}\|(L_K+\lambda
    I)^{1/2}(f_\lambda-f_\rho)\|_K.
\end{equation}
Plugging (\ref{Distributed error estimate}) and (\ref{sample for
KRR}) into (\ref{dis error decomp rho}), we have
\begin{eqnarray*}
        \frac12E\left[\|\overline{f}_{D,\lambda}^0-f_\rho\|_\rho^2\right]
        &\leq&
         \|f_\lambda-f_\rho\|_\rho^2+
         \sum_{j=1}^m\frac{|D_j|^2}{|D|^2}E\left[ \mathcal Q_{D_j,\lambda}^4(\mathcal P_{D_j,\lambda}+\mathcal
        S_{D_j,\lambda}\|f_\lambda\|_K) ^2  \right]\\
        &+&\sum_{j=1}^m\frac{|D_j|}{|D|}
        E\left[\mathcal Q_{D_j,\lambda}^4\mathcal R^2_{D_j,\lambda}\|(L_K+\lambda
    I)^{1/2}(f_\lambda-f_\rho)\|_K^2\right].
\end{eqnarray*}
This completes the proof of Proposition \ref{Proposition:Error
decomposition in ex}. $\Box$

\subsection{Error decomposition for DKRR in probability}
 To deduce learning rates for
DKRR in probability, we need the following error decomposition. It holds for
$\|\overline{f}_{D,\lambda}^0-f_{D,\lambda}\|_\rho$, which is
totally different from Proposition \ref{Proposition:Error
decomposition in ex} focusing  on the expectation.

\begin{proposition}\label{Proposition:Error decomposition in pro}
Let $\overline{f}_{D,\lambda}^0$ be defined by (\ref{DKRR}). Then
\begin{eqnarray}\label{Error decomp for DKRR}
   &&\|\overline{f}_{D,\lambda}^0-f_{D,\lambda}\|_\rho\leq \|(L_K+\lambda I)^{1/2}
   (\overline{f}_{D,\lambda}^0-f_{D,\lambda})\|_K\nonumber\\
    &\leq&
        \mathcal Q_{D,\lambda}^2\sum_{j=1}^m\frac{|D_j|}{|D|}(\mathcal
   R_{D_j,\lambda}+\mathcal
   R_{D,\lambda}) \mathcal Q_{D_j,\lambda}^2(\mathcal P_{D_j,\lambda}+\mathcal
        S_{D_j,\lambda}\|f_\lambda\|_K).
\end{eqnarray}
\end{proposition}

{\bf Proof.}
 From the definition of $\overline{f}_{D,\lambda}^0$, we see
\begin{eqnarray*}
      && \overline{f}_{D,\lambda}^0-f_{D,\lambda}
      =
      \sum_{j=1}^m\frac{|D_j|}{|D|}(L_{K,D_j}+\lambda
      I)^{-1}S_{D_j}^Ty_{D_j} -(L_{K,D}+\lambda I)^{-1}S_D^Ty_D\\
      &=&
      \sum_{j=1}^m\frac{|D_j|}{|D|}\left[(L_{K,D_j}+\lambda
      I)^{-1}-(L_{K,D}+\lambda
      I)^{-1}\right]S_{D_j}^Ty_{D_j}\\
      &=&
      \sum_{j=1}^m\frac{|D_j|}{|D|}(L_{K,D}+\lambda
      I)^{-1}(L_{K,D}-L_{K,D_j})f_{D_j,\lambda}\\
      &=&
      \sum_{j=1}^m\frac{|D_j|}{|D|}(L_{K,D}+\lambda
      I)^{-1}(L_{K,D}-L_K)(f_{D_j,\lambda}-f_\lambda)\\
      &+&
      \sum_{j=1}^m\frac{|D_j|}{|D|}(L_{K,D}+\lambda
      I)^{-1}(L_{K}-L_{K,D_j})(f_{D_j,\lambda}-f_\lambda)\\
      &=&(L_{K,D}+\lambda
      I)^{-1}(L_K+\lambda I)^{1/2}(L_K+\lambda
      I)^{-1/2}(L_{K,D}-L_K)(L_K+\lambda
      I)^{-1/2}\\
      &&\sum_{j=1}^m\frac{|D_j|}{|D|}(L_K+\lambda
      I)^{1/2}(f_{D_j,\lambda}-f_\lambda)\\
      &+&
     (L_{K,D}+\lambda
      I)^{-1}(L_K+\lambda I)^{1/2}\\
      &&\sum_{j=1}^m\frac{|D_j|}{|D|}(L_K+\lambda
      I)^{-1/2}(L_{K,D}-L_K)(L_K+\lambda I)^{-1/2}(L_K+\lambda
      I)^{1/2}(f_{D_j,\lambda}-f_\lambda).
\end{eqnarray*}
Then,
\begin{eqnarray*}
  &&
   \|(L_K+\lambda I)^{1/2}
   (\overline{f}_{D,\lambda}^0-f_{D,\lambda})\|_K
  \leq
  \mathcal Q_{D,\lambda}^2\mathcal R_{D,\lambda}
   \sum_{j=1}^m\frac{|D_j|}{|D|}\|(L_K+\lambda
  I)^{1/2}(f_{D_j,\lambda}-f_\lambda)\|_K\\
  &+&
  \mathcal Q_{D,\lambda}^2
    \sum_{j=1}^m\frac{|D_j|}{|D|}\mathcal R_{D_j,\lambda}\|(L_K+\lambda
  I)^{1/2}(f_{D_j,\lambda}-f_\lambda)\|_K.
\end{eqnarray*}
This together with (\ref{sample for KRR}) gives
\begin{eqnarray*}
   \| \overline{f}_{D,\lambda}^0-f_{D,\lambda}\|_\rho
   &\leq&
   \mathcal Q_{D,\lambda}^2\mathcal
   R_{D,\lambda}\sum_{j=1}^m\frac{|D_j|}{|D|} \mathcal Q_{D_j,\lambda}^2(\mathcal P_{D_j,\lambda}+\mathcal
        S_{D_j,\lambda}\|f_\lambda\|_K)\\
        &+&
        \mathcal Q_{D,\lambda}^2\sum_{j=1}^m\frac{|D_j|}{|D|}\mathcal
   R_{D_j,\lambda} \mathcal Q_{D_j,\lambda}^2(\mathcal P_{D_j,\lambda}+\mathcal
        S_{D_j,\lambda}\|f_\lambda\|_K).
\end{eqnarray*}
This completes the proof of Proposition \ref{Proposition:Error
decomposition in pro}. $\Box$

\subsection{Error decomposition for DKRR($\ell$)}
In this subsection, we derive an error decomposition for
DKRR($\ell$). At first, we   show  in the following proposition the
power of communications.

\begin{proposition}\label{Proposition: power for c}
Let $\ell\geq0$. We have
\begin{eqnarray}\label{power for c}
    &&\|(L_K+\lambda I)^{1/2} (\overline{f}_{D,\lambda}^{\ell}-
    f_{D,\lambda})\|_K    \\
    &\leq&
    \left(\sum_{j=1}^m\frac{|D_j|}{|D|}\mathcal Q_{D_j,\lambda}^2
    (\mathcal
    R_{D_j,\lambda}+
    \mathcal R_{D,\lambda})\right)^{\ell}\|(L_K+\lambda
    I)^{1/2}(\overline{f}_{D,\lambda}^{0}- f_{D,\lambda})\|_K.
    \nonumber
\end{eqnarray}
\end{proposition}

{\bf Proof.} Since
$$
      f_{D,\lambda}=\overline{f}_{D,\lambda}^{\ell-1}-(L_{K,D}+\lambda I)^{-1}[(L_{K,D}+\lambda
      I)\overline{f}_{D,\lambda}^{\ell-1}-S_D^Ty_D],
$$
we have
\begin{eqnarray*}
     &&   f_{D,\lambda}-\overline{f}_{D,\lambda}^{\ell}
     =\overline{f}^{\ell-1}_{D,\lambda}
    -
    \sum_{j=1}^m\frac{|D_j|}{|D|}(L_{K,D_j}+\lambda I)^{-1}[(L_{K,D}+\lambda
        I)\overline{f}^{\ell-1}_{D,\lambda}-S_{D}^Ty_{D}]\\
        &-&
        \overline{f}_{D,\lambda}^{\ell-1}+(L_{K,D}+\lambda I)^{-1}[(L_{K,D}+\lambda
      I)\overline{f}_{D,\lambda}^{\ell-1}-S_D^Ty_D]\\
      &=&
      \sum_{j=1}^m\frac{|D_j|}{|D|}[(L_{K,D_j}+\lambda I)^{-1}
      -(L_{K,D}+\lambda I)^{-1}][(L_{K,D}+\lambda
        I)\overline{f}^{\ell-1}_{D,\lambda}-S_{D}^Ty_{D}]\\
        &=&
        \sum_{j=1}^m\frac{|D_j|}{|D|} (L_{K,D_j}+\lambda
        I)^{-1}(L_{K,D}-L_{K,D_j})
       (L_{K,D}+\lambda I)^{-1}[(L_{K,D}+\lambda
        I)\overline{f}^{\ell-1}_{D,\lambda}-S_{D}^Ty_{D}]\\
        &=&
       \sum_{j=1}^m\frac{|D_j|}{|D|} (L_{K,D_j}+\lambda
        I)^{-1}(L_{K,D}-L_{K,D_j})
        ( \overline{f}^{\ell-1}_{D,\lambda}-f_{D,\lambda})\\
        &=&
         \sum_{j=1}^m\frac{|D_j|}{|D|} (L_{K,D_j}+\lambda
        I)^{-1}(L_{K,D}-L_K)
        ( \overline{f}^{\ell-1}_{D,\lambda}-f_{D,\lambda})\\
        &+&
        \sum_{j=1}^m\frac{|D_j|}{|D|} (L_{K,D_j}+\lambda
        I)^{-1}(L_K-L_{K,D_j})
        ( \overline{f}^{\ell-1}_{D,\lambda}-f_{D,\lambda})\\
        &=:&
        \mathcal U_{D,\lambda,1}+\mathcal U_{D,\lambda,2}.
\end{eqnarray*}
Since
\begin{eqnarray*}
       &&(L_K+\lambda I)^{1/2}\mathcal U_{D,\lambda,1}\\
       &=&
       \sum_{j=1}^m\frac{|D_j|}{|D|}(L_K+\lambda I)^{1/2} (L_{K,D_j}+\lambda
        I)^{-1}(L_K+\lambda I)^{1/2} (L_K+\lambda
        I)^{-1/2}(L_{K,D}-L_K)\\
        &\times&
        (L_K+\lambda I)^{-1/2}(L_K+\lambda I)^{1/2}( \overline{f}^{\ell-1}_{D,\lambda}-f_{D,\lambda}),
\end{eqnarray*}
we have
$$
    \|(L_K+\lambda I)^{1/2}\mathcal U_{D,\lambda,1}\|_K
    \leq \sum_{j=1}^m\frac{|D_j|}{|D|}\mathcal Q_{D_j,\lambda}^2 \mathcal R_{D,\lambda}\|(L_K+\lambda
    I)^{1/2}(\overline{f}_{D,\lambda}^{\ell-1}- f_{D,\lambda})\|_K.
$$
Similarly, we find
$$
    \|(L_K+\lambda I)^{1/2}\mathcal U_{D,\lambda,2}\|_K
    \leq \sum_{j=1}^m\frac{|D_j|}{|D|}\mathcal Q_{D_j,\lambda}^2 \mathcal R_{D_j,\lambda}\|(L_K+\lambda
    I)^{1/2}(\overline{f}_{D,\lambda}^{\ell-1}- f_{D,\lambda})\|_K.
$$
Then,
\begin{eqnarray*}
    &&\|(L_K+\lambda I)^{1/2} (\overline{f}_{D,\lambda}^{\ell}-
    f_{D,\lambda})\|_K\\
    &\leq&
    \left(\sum_{j=1}^m\frac{|D_j|}{|D|}\mathcal Q_{D_j,\lambda}^2
    \mathcal R_{D_j,\lambda}+ \sum_{j=1}^m\frac{|D_j|}{|D|}\mathcal Q_{D_j,\lambda}^2
    \mathcal R_{D,\lambda}\right)\|(L_K+\lambda
    I)^{1/2}(\overline{f}_{D,\lambda}^{\ell-1}- f_{D,\lambda})\|_K\\
    &\leq&
    \left(\sum_{j=1}^m\frac{|D_j|}{|D|}\mathcal Q_{D_j,\lambda}^2
   \mathcal  R_{D_j,\lambda}+ \sum_{j=1}^m\frac{|D_j|}{|D|}\mathcal Q_{D_j,\lambda}^2
   \mathcal  R_{D,\lambda}\right)^{\ell}\|(L_K+\lambda
    I)^{1/2}(\overline{f}_{D,\lambda}^{0}- f_{D,\lambda})\|_K.
\end{eqnarray*}
This completes the proof of Proposition \ref{Proposition: power for
c}. $\Box$

Combining Proposition \ref{Proposition: power for c} with
Proposition \ref{Proposition:Error decomposition in pro}, we can
derive the following error decomposition for DKRR($\ell$).

\begin{proposition}\label{Proposition: error decomposition for l}
Let $\ell\geq0$. We have
\begin{eqnarray}\label{error decomposition for l}
    &&\|(L_K+\lambda I)^{1/2} (\overline{f}_{D,\lambda}^{\ell}-
    f_{D,\lambda})\|_K
     \leq
    \left(\sum_{j=1}^m\frac{|D_j|}{|D|}\mathcal Q_{D_j,\lambda}^2
    (\mathcal R_{D_j,\lambda}+\mathcal R_{D,\lambda})
    \right)^{\ell}\nonumber\\
    &\times&  \mathcal Q_{D,\lambda}^2\sum_{j=1}^m\frac{|D_j|}{|D|}(\mathcal
   R_{D,\lambda}+\mathcal
   R_{D_j,\lambda}) \mathcal Q_{D_j,\lambda}^2(\mathcal P_{D_j,\lambda}+\mathcal
        S_{D_j,\lambda}\|f_\lambda\|_K).
    \nonumber
\end{eqnarray}
\end{proposition}

\section{Proofs}\label{Sec.Proof}
In this section, we present proofs of our main results.

\subsection{Optimal learning rates for DKRR in expectation}

In this subsection, we prove optimal learning rates for DKRR in
expectation. We need the following general theorem based
on Assumption \ref{Assumption:bounded for output} and Assumption
\ref{Assumption:regularity}
\begin{theorem}\label{Theorem:rate for DKRR in expectation}
Under Assumption \ref{Assumption:regularity} with $\frac12\leq r\leq
1$ and Assumption \ref{Assumption:bounded for output}, if
$0<\lambda\leq1$ and $\mathcal N(\lambda)\geq 1$, then
\begin{eqnarray}\label{rate for dkrr in expectation}
         E\left[\|\overline{f}_{D,\lambda}^0-f_\rho\|_\rho^2\right]
        &\leq&
         2\lambda^{2r}+
         24\sum_{j=1}^m\frac{|D_j|^2}{|D|^2} \left(\exp\{-1/(2C_1^*\mathcal B_{D_j,\lambda})\}
   +
    \tilde{C}_1^2\mathcal
    A^2_{D_j,\lambda}\right)
           \nonumber\\
        &+&\sum_{j=1}^m\frac{|D_j|}{|D|}
        \left(16\exp\{-1/(2C_1^*\mathcal B_{D_j,\lambda})\}
   +
   32\tilde{C}_2^2\mathcal
    B^2_{D_j,\lambda}\lambda^{2r}\right).
\end{eqnarray}
where $\tilde{C}_1:=2(2(\kappa
        M+\gamma)(\kappa+1)+2\kappa^{2r}(2\kappa+1)\|h_\rho\|_\rho)$ and
        $\tilde{C_2}:=4\kappa(\kappa+1)\|h_\rho\|_\rho$.
\end{theorem}

{\bf Proof.} Due to  Assumption \ref{Assumption:regularity} with
$r\geq 1/2$, we obtain
\begin{equation}\label{bound flambda}
       \|f_\lambda\|_K=\|(L_K+\lambda
       I)^{-1}L_KL_K^rh_\rho\|\leq \kappa^{2r-1}\|L_K^{1/2}h_\rho\|_K
       =\kappa^{2r-1}\|h_\rho\|_\rho.
\end{equation}
Moreover  (\ref{Boundedness for output}) implies
\citep{Blanchard2016,Lin2018} that with confidence at least
$1-\delta$, there holds
\begin{equation}\label{difference 1}
   \mathcal P_{D_j,\lambda}    \leq   2(\kappa M+\gamma)(\kappa+1)\mathcal A_{D_j,\lambda}
   \log\frac2\delta.
 \end{equation}
Thus, for $\delta\geq 12\exp\{-1/(2C_1^*\mathcal
B_{D_j,\lambda})\}$, it follows from Lemma \ref{Lemma:operator
product}, (\ref{difference 1}), (\ref{Define S}) and (\ref{bound
flambda}) that with confidence $1-\delta$, there holds
\begin{equation}\label{DKRR:bound 1}
    \mathcal Q_{D_j,\lambda}^2\left(\mathcal P_{D_j,\lambda}+\mathcal
        S_{D_j,\lambda}\|f_\lambda\|_K\right)
        \leq
        \tilde{C}_1\mathcal
        A_{D_j,\lambda}\log\frac{6}{\delta},\qquad\forall\
        j=1,\dots,m.
\end{equation}
 Using the probability to expectation formula
\begin{equation}\label{expectation formula}
               E[\xi] =\int_0^\infty P\left[\xi > t\right] d t
\end{equation}
  to the positive random variable $\xi_{1,j} =
 \mathcal Q^4_{D_j,\lambda}(\mathcal
         P_{D_j,\lambda}+ \mathcal
         S_{D_j,\lambda}\|f_{\lambda}\|_K)^2$ for any
         $j=1,\dots,m$,
we have
\begin{eqnarray*}
    &&E[\xi_{1,j}]=\int_0^\infty P\left[\xi_{1,j} > t\right] d t\\
    &=&
    \int_0^{12\exp\{-1/(2C_1^*\mathcal B_{D_j,\lambda})\}} P\left[\xi_{1,j} > t\right] d t
  +
  \int_{12\exp\{-1/(2C_1^*\mathcal B_{D_j,\lambda})\}}^\infty P\left[\xi_{1,j} > t\right] d
   t\\
   &\leq&
   12\exp\{-1/(2C_1^*\mathcal B_{D_j,\lambda})\}
   +
    \int_{12\exp\{-1/(2C_1^*\mathcal B_{D_j,\lambda})\}}^\infty
    P[ \xi_{1,j}>t]dt.
\end{eqnarray*}
When $t\geq 12\exp\{-1/(2C_1^*\mathcal B_{D_j,\lambda})\}$, it
follows from (\ref{DKRR:bound 1}) that
$$
      P[ \xi_{1,j}>t]
         \leq
         6\exp\{-\tilde{C}_1^{-1}\mathcal
         A_{D_j,\lambda}^{-1}t^{1/2}\},\qquad\forall\ j=1,\dots,m,
$$
and
\begin{eqnarray*}
    &&\int_{12\exp\{-1/(2C_1^*\mathcal B_{D_j,\lambda})\}}^\infty
   P[\xi_{1,j}>t]dt
        \leq
    6\int_0^\infty\exp\{-\tilde{C}_1^{-1}\mathcal
         A_{D_j,\lambda}^{-1}t^{1/2}\}dt\\
    &\leq&
    12\tilde{C}_1^2\mathcal
    A^2_{D_j,\lambda}\int_0^\infty ue^{-u}du=12\tilde{C}_1^2\mathcal
    A^2_{D_j,\lambda}.
\end{eqnarray*}
Thus,
\begin{equation}\label{DKRR.2 expectation}
     E\left[ \mathcal Q_{D_j,\lambda}^4(\mathcal P_{D_j,\lambda}+\mathcal
        S_{D_j,\lambda}\|f_\lambda\|_K) ^2  \right]
        \leq
       12\exp\{-1/(2C_1^*\mathcal B_{D_j,\lambda})\}
   +
   12\tilde{C}_1^2\mathcal
    A^2_{D_j,\lambda}.
\end{equation}
Due to   Assumption \ref{Assumption:regularity} with $r\geq1/2$, we
have
\begin{equation}\label{Approximation error}
      \|(L_K+\lambda I)^{1/2}(f_\lambda-f_\rho)\|_K
      \leq\lambda\|(L_K+\lambda I)^{-1/2}L_K^{r+1/2}\|\|h_\rho\|_\rho
      \leq \lambda^r\|h_\rho\|_\rho.
\end{equation}
Then, Lemma \ref{Lemma:operator difference} and Lemma
\ref{Lemma:operator product} with
$\delta\geq8\exp\{-1/(2C_1^*\mathcal B_{D_j,\lambda})\}$ yield that
 with confidence
$1-\delta$, there holds
\begin{equation}\label{DKRR:bound 2}
   \mathcal Q_{D_j,\lambda}^2 \mathcal
    R_{D_j,\lambda}\|(L_K+\lambda
    I)^{1/2}(f_\lambda-f_\rho)\|_K\leq
    \tilde{C}_2\mathcal B_{D_j,\lambda}
    \lambda^r\log\frac{8}\delta,\qquad\forall \ j=1,\dots,m.
\end{equation}
 Then, applying (\ref{expectation formula}) to $\xi_{2,j} =\mathcal
Q_{D_j,\lambda}^4 \mathcal
    R^2_{D_j,\lambda}\|(L_K+\lambda
    I)^{1/2}(f_\lambda-f_\rho)\|_K^2$ and using the same approach as
    above,
we can derive for any $j=1,\dots,m$,
\begin{equation}\label{DKRR.3 expectation}
   E[\mathcal Q_{D_j,\lambda}^4
   \mathcal
    R^2_{D_j,\lambda}\|(L_K+\lambda
    I)^{1/2}(f_\lambda-f_\rho)\|_K^2] \leq
   8\exp\{-1/(2C_1^*\mathcal B_{D_j,\lambda})\}
   +
   16\tilde{C}_2^2\mathcal
    B^2_{D_j,\lambda}\lambda^{2r}.
\end{equation}
Plugging (\ref{Approximation error}), (\ref{DKRR.2 expectation}) and
(\ref{DKRR.3 expectation}) into (\ref{Error decomp for DKRR in
expectation}), we get (\ref{rate for dkrr in expectation}) directly.
This completes the proof of Theorem \ref{Theorem:rate for DKRR in
expectation}. $\Box$

Based on Theorem \ref{Theorem:rate for DKRR in expectation}, we can
prove Theorem \ref{Theorem:optimal dkrr in expectation} as follows.

{\bf Proof of Theorem \ref{Theorem:optimal dkrr in expectation}.}
For
$C_1:=\min\left\{\frac{2r+s}{2s},\frac{(2r+s)^3}{2rs\max\{(\kappa^2+1)/3,2\sqrt{\kappa^2+1}\}}\right\}$,
due to (\ref{Def.B}), $|D_1|=\dots=|D_m|$,
$\lambda=|D|^{-\frac1{2r+s}}$ and (\ref{assumption on effect}), we
have for any $j=1,\dots,m$,
\begin{eqnarray}\label{Bound Bj}
    \mathcal B_{D_j,\lambda}
    &\leq&
    \frac{2s}{2r+s}m|D|^{-\frac{2r+s-1}{2r+s}}\log|D|
    +
    \sqrt{\frac{2s}{2r+s}m|D|^{-\frac{2r+s-1}{2r+s}}\log|D|}.
\end{eqnarray}
Noting $C_1^*=\max\{(\kappa^2+1)/3,2\sqrt{\kappa^2+1}\}$, we see that
(\ref{Restiction on m for DKRR}) implies
\begin{eqnarray}\label{Bound Bj1}
     \mathcal B_{D_j,\lambda}\leq \frac{2r+s}{4rC_1^*\log|D|}
\end{eqnarray}
and
\begin{equation}\label{bound exponential}
    \exp\{-1/(2C_1^*\mathcal B_{D_j,\lambda})\}\leq
    |D|^{-\frac{2r}{2r+s}},\qquad \forall j=1,\dots,m.
\end{equation}
According to (\ref{Def.A}), $|D_1|=\dots=|D_m|$,
$\lambda=|D|^{-\frac1{2r+s}}$ and (\ref{assumption on effect}), we
also get
\begin{eqnarray}\label{Bound Aj}
    \mathcal A_{D_j,\lambda}
     \leq
    m|D|^{-\frac{4r+2s-1}{4r+2s}}+\sqrt{m}|D|^{-\frac{r}{2r+s}},\qquad\forall j=1,\dots,m.
\end{eqnarray}
Inserting (\ref{Bound Aj}), (\ref{Bound Bj1}) and (\ref{bound
exponential}) into (\ref{rate for dkrr in expectation}) and noting
$\lambda=|D|^{-\frac{1}{2r+s}}$ and (\ref{Restiction on m for
DKRR}), we have
$$
       E[\|\overline{f}_{D,\lambda}^0-f_\rho\|_\rho^2]
       \leq
       C_1|D|^{-\frac{2r}{2r+s}},
$$
where
$C_2:=42+48\tilde{C}_1^2+32\tilde{C}_2^2\left(\frac{2r+s}{4rC_1^*}\right)^2.$
This completes the proof of Theorem \ref{Theorem:optimal dkrr in
expectation}. $\Box$

\subsection{Optimal learning rates for DKRR in probability}

In this subsection, we prove Theorem \ref{Theorem:optimal
dkrr in probability}. To this end, we need the following theorem for
DKRR in probability.

\begin{theorem}\label{Theorem:rate for DKRR}
  Under Assumption \ref{Assumption:bounded for
output}   and  Assumption \ref{Assumption:regularity} with
$\frac12\leq r\leq 1$, if $0<\lambda\leq1$, $\mathcal N(\lambda)\geq
1$, and
\begin{equation}\label{restion on delta in theorem5a}
  16m\exp\{-1/(2C_1^*\mathcal B_{D_j,\lambda})\}<1,\qquad\forall
  j=1,\dots,m,
\end{equation}
  then for
\begin{equation}\label{restion on delta in theorem5b}
   16m\exp\{-1/(2C_1^*\mathcal B_{D_j,\lambda})\}\leq\delta<1, \qquad\forall
  j=1,\dots,m,
\end{equation}
 with confidence $1-\delta$ there holds
\begin{equation}\label{rate for DKRR}
      \|\overline{f}_{D,\lambda}^0-f_{D,\lambda}\|_\rho
   \leq \tilde{C_3}\log^2(4m)\sum_{j=1}^m\frac{|D_j|}{|D|}\mathcal
    A_{D_j,\lambda}\mathcal B_{D_j,\lambda}\log^2\frac{4}\delta,
\end{equation}
  where $\tilde{C}_3=16C_1^*\tilde{C}_1.$
\end{theorem}

{\bf Proof.} Let $j\in\{1,\dots,m\}$ be fixed. It follows from Lemma
\ref{Lemma:operator difference} and $\mathcal B_{D,\lambda}\leq
\mathcal B_{D_j,\lambda}$ that with confidence $1-\frac\delta4$,
there holds
\begin{equation}\label{DKRR.3}
      \mathcal R_{D,\lambda}+\mathcal R_{D_j,\lambda}
      \leq
      2C_1^*\mathcal B_{D_j,\lambda}\log\frac{16}\delta.
\end{equation}
Since $\delta\geq 16\exp\{-1/(2C_1^*\mathcal B_{D_j,\lambda})\}$,
Lemma \ref{Lemma:operator product} together with (\ref{DKRR:bound
1}) shows that
   with confidence $1-3\delta/4$, there holds
\begin{equation}\label{DKRR:bound 3}
    \mathcal Q_{D,\lambda}^2\mathcal Q_{D_j,\lambda}^2\left(\mathcal P_{D_j,\lambda}+\mathcal
        S_{D_j,\lambda}\|f_\lambda\|_K\right)
        \leq
         2\tilde{C}_1\mathcal
        A_{D_j,\lambda}\log\frac{16}{\delta}.
\end{equation}
Plugging (\ref{DKRR:bound 3}) and (\ref{DKRR.3}) into (\ref{Error
decomp for DKRR}), for fixed $j\in \{1,\dots,m\}$,
we see with confidence $1-\delta$, there holds
\begin{equation}\label{DKRR.7}
     \mathcal Q_{D,\lambda}^2 (\mathcal
   R_{D_j,\lambda}+\mathcal
   R_{D,\lambda}) \mathcal Q_{D_j,\lambda}^2(\mathcal P_{D_j,\lambda}+\mathcal
        S_{D_j,\lambda}\|f_\lambda\|_K)
        \leq
        \tilde{C_3}\log^2\frac{16}\delta\mathcal
    A_{D_j,\lambda}\mathcal B_{D_j,\lambda}.
\end{equation}
Thus, for $\delta\geq 16\exp\{-1/(2C_1^*\mathcal
B_{D_j,\lambda})\}$, the above estimate  implies that   with
confidence at least $1-m\delta$, there holds
$$
     \max_{1\leq j\leq m}\mathcal Q_{D,\lambda}^2 (\mathcal
   R_{D_j,\lambda}+\mathcal
   R_{D,\lambda}) \mathcal Q_{D_j,\lambda}^2(\mathcal P_{D_j,\lambda}+\mathcal
        S_{D_j,\lambda}\|f_\lambda\|_K)
        \leq
        \tilde{C_3}\log^2\frac{16}\delta\mathcal
    A_{D_j,\lambda}\mathcal B_{D_j,\lambda}.
$$
Scaling $m\delta$ to $\delta$, for $\delta\geq
16m\exp\{-1/(2C_1^*\mathcal B_{D_j,\lambda})\}$, with confidence
$1-\delta$, there holds
$$
    \|\overline{f}_{D,\lambda}^0-f_{D,\lambda}\|_\rho
   \leq
    4C_1^*\tilde{C}_1\log^2\frac{16m}\delta\sum_{j=1}^m\frac{|D_j|}{|D|}\mathcal
    A_{D_j,\lambda}\mathcal B_{D_j,\lambda}.
$$
Note that
$$
   \log\frac{16m}\delta=\log\frac4\delta+\log(4m)\leq
   (\log(4m)+1)\log\frac4\delta\leq2\log(4m)\log\frac4\delta,
$$
which follows from (\ref{rate for DKRR}).  Then
  the proof of Theorem \ref{Theorem:rate for DKRR} is complete.
$\Box$

{\bf Proof of Theorem \ref{Theorem:optimal dkrr in probability}.}
Due to the triangle inequality, we have
\begin{equation}\label{triangle1}
    \|\overline{f}_{D,\lambda}^0-f_\rho\|_\rho
    \leq\|f_{D,\lambda}-\overline{f}_{D,\lambda}^0\|_\rho+\|f_{D,\lambda}-f_\rho\|_\rho.
\end{equation}
To bound $\|f_{D,\lambda}-f_\rho\|_\rho$, under
(\ref{regularitycondition}) with $\frac12\leq r\leq 1$, we obtain
from (\ref{Approximation error}) and (\ref{sample for KRR}) that
$$
   \|f_{D,\lambda}-f_\rho\|_\rho
   \leq
   \lambda^r+\mathcal Q_{D,\lambda}^2(\mathcal P_{D,\lambda}+\mathcal
        S_{D,\lambda}\|f_\lambda\|_K).
$$
Then, (\ref{Restiction on m for DKRR in prob}) with
$C_3:=\max\{4C_1^*(4\log2+1),16\log^22\}$ and (\ref{Restriction on
delta 111}) with $C_4:=4C_1^*$ imply $\delta\geq
\exp\{2C_1^*\mathcal B_{D,\lambda}\}$, and then (\ref{DKRR:bound 2})
with $D_j$  replaced by $D$ shows that with confidence
$1-\delta/2$, we know that there holds
\begin{eqnarray}\label{estimate for KRR}
    \|f_{D,\lambda}-f_\rho\|_\rho
    \leq
      \lambda^r+\tilde{C}_1\mathcal
        A_{D,\lambda}\log\frac{16}{\delta}.
\end{eqnarray}
Noting (\ref{Restiction on m for DKRR in prob}) and
(\ref{Restriction on delta 111}) imply (\ref{restion on delta in
theorem5a}) and (\ref{restion on delta in theorem5b}), and plugging
(\ref{estimate for KRR}) and (\ref{rate for DKRR}) into
(\ref{triangle1}), with confidence $1-\delta$, there holds
\begin{equation}\label{DKRR bound. 5}
   \|\overline{f}_{D,\lambda}^0-f_\rho\|_\rho
   \leq
   \lambda^r+\tilde{C}_1\mathcal
        A_{D,\lambda}\log\frac{16}{\delta}
        + \tilde{C_3}\log^2(4m)\log^2\frac{8}\delta\sum_{j=1}^m\frac{|D_j|}{|D|}\mathcal
    A_{D_j,\lambda}\mathcal B_{D_j,\lambda}.
\end{equation}
Since $|D_1|=\dots=|D_m|$, $\lambda=|D|^{-\frac1{2r+s}}$ and
$2s/(2r+s)\leq 1$, we have from (\ref{Bound Aj}) and (\ref{Bound
Bj}) that
\begin{eqnarray}\label{Bound AtimesB}
     && \mathcal
    A_{D_j,\lambda}\mathcal B_{D_j,\lambda}
     \leq
    \left(m|D|^{-\frac{4r+2s-1}{4r+2s}}+\sqrt{m}|D|^{-\frac{r}{2r+s}}\right) \nonumber\\
    &\times&
    \left(m|D|^{-\frac{2r+s-1}{2r+s}}\log|D|
    +
    \sqrt{m|D|^{-\frac{2r+s-1}{2r+s}}\log|D|}\right).
\end{eqnarray}
This together with $(\ref{Restiction on m for DKRR in prob})$ shows
$$
   \log^2(4m) \mathcal
    A_{D_j,\lambda}\mathcal B_{D_j,\lambda}
    \leq 4|D|^{-\frac{r}{2r+s}}.
$$
Inserting the above inequality into (\ref{DKRR bound. 5}) and
noting
\begin{equation}\label{bound ADlambda}
    \mathcal A_{D,\lambda}
     \leq
    |D|^{-\frac{4r+2s-1}{4r+2s}}+|D|^{-\frac{r}{2r+s}}\leq 2|D|^{-\frac{r}{2r+s}},
\end{equation}
we see with confidence $1-\delta$, there holds
$$
   \|\overline{f}_{D,\lambda}^0-f_\rho\|_\rho
   \leq
   C_5|D|^{-\frac{r}{2r+s}}\log^2\frac{8}{\delta},
$$
where we use $\log\frac{16}\delta\leq\log^2\frac{8}{\delta}$ and
$$
    C_5=1+2\tilde{C}_1+4\tilde{C}_3.
$$
This completes the proof of Theorem \ref{Theorem:optimal dkrr in
probability}. $\Box$

To prove Corollary \ref{Corollary: almost surely convergence}, we
need the following  Borel-Cantelli Lemma  \citep[page
262]{Dudley2002}. The Borel-Cantelli Lemma asserts  for a sequence
$\{\eta_n\}_n$ of events that if the sum of the probabilities is
finite, i.e., $\sum_{n=1}^\infty P[\eta_n]<\infty$, then the probability
that infinitely many of them occur is $0$.

\begin{lemma}\label{BOREL CANTELLI LEMMA}
Let $\{\eta_n\}$ be a sequence of events in some probability space
and $\{\varepsilon_n\}$ be a sequence of positive numbers satisfying
$\lim_{n\rightarrow\infty}\varepsilon_n=0$. If
$$
        \sum_{n=1}^\infty
        \mbox{Prob}[|\eta_n-\eta|>\varepsilon_n]<\infty,
$$
then $\eta_n$ converges to $\eta$ almost surely.
\end{lemma}

\noindent {\it Proof of Corollary \ref{Corollary: almost surely
convergence}.} Let $N:=|D|$, it is easy to check that
$\delta=\delta_N=\tilde{C}_4 N^{-2} $  satisfies (\ref{Restriction
on delta 111}) for some $\tilde{C}_4>0$ independent of $N$.
 Set
$\Psi_N=N^{-\frac{r}{2r+s}}$. By Theorem \ref{Theorem:optimal dkrr
in probability}, if $\lambda=|D|^{-\frac{1}{2r+s}}$,
$|D_1|=\dots=|D_m|$ and
 (\ref{Restiction on m for DKRR in prob}) holds, then
 for any $N$ and $\varepsilon>0$,
$$
      P\left[\Psi_N^{-1+\varepsilon}\|\overline{f}^0_{D,\lambda}-f_\rho\|_\rho
      >
      C_5\Psi_N^\varepsilon\left(\log\frac{8}{\delta_N}\right)^{2}\right]
      \leq\delta_N.
$$
 Denote
$\mu_N=C_5\Psi_N^\varepsilon\left(\log\frac{8}{\delta_N}\right)^{2}$.
Obviously,
$$
           \sum_{N=2}^\infty P\left[\Psi_N^{-1+\varepsilon}\|
           \overline{f}_{t,D}-f_\rho\|_\rho>\mu_N\right]
           \leq\sum_{N=2}^\infty\delta_N<\infty
$$
and $\mu_N\rightarrow0$ when $N\rightarrow0$. Then our conclusion
follows from Lemma \ref{BOREL CANTELLI LEMMA}. This completes the
proof of Corollary \ref{Corollary: almost surely convergence}.
  \hfill  $\Box$

\subsection{Optimal learning rates for DKRR($\ell$)}

In this subsection, we present the proof of Theorem
\ref{Theorem:optimal l}. To this end, we  prove the following
theorem.

\begin{theorem}\label{Theorem:rate for l}
  Under Assumption \ref{Assumption:bounded for
output}   and  Assumption \ref{Assumption:regularity} with
$\frac12\leq r\leq 1$, if $0<\lambda\leq1$, $\mathcal N(\lambda)\geq
1$, and (\ref{restion on delta in theorem5a}) holds,
  then for $\delta$ satisfying (\ref{restion on delta in
  theorem5b}),
 with confidence $1-\delta$
  there holds
\begin{equation}\label{rate for DKRR l}
      \|\overline{f}_{D,\lambda}^{\ell}-
    f_{D,\lambda}\|_\rho
    \leq
    2\tilde{C}_3 (4C_1^*)^\ell
    \left(\log(4m)\log\frac4\delta\right)^{\ell+2}
   \left(\sum_{j=1}^m
     \frac{|D_j|}{|D|}\mathcal
     B_{D_j,\lambda}\right)^\ell
     \sum_{j=1}^m
     \frac{|D_j|}{|D|}\mathcal A_{D_j,\lambda}\mathcal
     B_{D_j,\lambda}.
\end{equation}
\end{theorem}

{\bf Proof.} Under (\ref{restion on delta in theorem5a}) and
(\ref{restion on delta in theorem5b}), we obtain from Lemma
\ref{Lemma:operator product}, (\ref{DKRR.3}) and (\ref{DKRR.7}) that
with confidence $1-m\delta$, there holds
$$
     \max_{1\leq j\leq m} \mathcal Q_{D_j,\lambda}^2
    (\mathcal R_{D_j,\lambda}+\mathcal R_{D,\lambda})
    \leq
    4C_1^*\mathcal B_{D_j,\lambda}\log\frac{16}\delta,
$$
and
\begin{eqnarray*}
   \max_{1\leq j\leq m}\mathcal Q_{D,\lambda}^2(\mathcal
   R_{D,\lambda}+\mathcal
   R_{D_j,\lambda}) \mathcal Q_{D_j,\lambda}^2(\mathcal P_{D_j,\lambda}+\mathcal
        S_{D_j,\lambda}\|f_\lambda\|_K)
         \leq
         \tilde{C_3}\log^2\frac{16}\delta\mathcal
    A_{D_j,\lambda}\mathcal B_{D_j,\lambda}.
\end{eqnarray*}
Scaling $m\delta$ to  $\delta$, we obtain from (\ref{error
decomposition for l}) that
\begin{eqnarray*}
     &&\|\overline{f}_{D,\lambda}^{\ell}-
    f_{D,\lambda}\|_\rho\leq
    \|(L_K+\lambda I)^{1/2} (\overline{f}_{D,\lambda}^{\ell}-
    f_{D,\lambda})\|_K\\
     &\leq& \left(4C_1^*\log\frac{16m}\delta\right)^\ell \left(\sum_{j=1}^m
     \frac{|D_j|}{|D|}\mathcal
     B_{D_j,\lambda}\right)^\ell\tilde{C_3}\log^2\frac{16m}\delta
    \sum_{j=1}^m
     \frac{|D_j|}{|D|}\mathcal A_{D_j,\lambda}\mathcal
     B_{D_j,\lambda}.
\end{eqnarray*}
 Noting further $\log\frac{16}{\delta}\leq 2\log 4m$,  we have with
 confidence $1-\delta$,
$$
   \|\overline{f}_{D,\lambda}^{\ell}-
    f_{D,\lambda}\|_\rho
    \leq
    2\tilde{C}_3 (4C_1^*)^\ell \log^{\ell+2}(4m)\log^{\ell+2}\frac4\delta
   \left(\sum_{j=1}^m
     \frac{|D_j|}{|D|}\mathcal
     B_{D_j,\lambda}\right)^\ell
     \sum_{j=1}^m
     \frac{|D_j|}{|D|}\mathcal A_{D_j,\lambda}\mathcal
     B_{D_j,\lambda}.
$$
This completes the proof of Theorem \ref{Theorem:rate for l}. $\Box$

{\bf Proof of Theorem \ref{Theorem:optimal l}.}    Since
$|D_1|=\dots=|D_m|$ and $\lambda=|D|^{-\frac1{2r+s}}$, it  follows
from (\ref{Bound Bj}) and $2s/(2r+s)\leq 1$ that
$$
   \left(\sum_{j=1}^m\frac{|D_j|}{|D|} \mathcal B_{D_j,\lambda}
    \right)^{\ell}\leq
    \left(m|D|^{-\frac{2r+s-1}{2r+s}}\log|D|)
    +
    \sqrt{m\log|D|}|D|^{-\frac{2r+s-1}{4r+2s}}\right)^\ell.
$$
This together with (\ref{Bound AtimesB}) shows that with confidence
$1-\delta$, there holds
\begin{eqnarray*}
    &&\left(\sum_{j=1}^m\frac{|D_j|}{|D|} \mathcal B_{D_j,\lambda}
    \right)^{\ell}\sum_{j=1}^m\frac{|D_j|}{|D|}\mathcal B_{D_j,\lambda}\mathcal
        A_{D_j,\lambda}\\
        &\leq&
     \left(m|D|^{-\frac{2r+s-1}{2r+s}}\log|D|
    +
    \sqrt{m\log|D|}|D|^{-\frac{2r+s-1}{4r+2s}}\right)^{\ell+1}
     \left(m|D|^{-\frac{4r+2s-1}{4r+2s}}+\sqrt{m}|D|^{-\frac{r}{2r+s}}\right).
\end{eqnarray*}
Since (\ref{Restiction on m for l}) with $C_6:=(4C_1^*(2+4\log
2))^2$ and (\ref{Restriction on delta 111}) with $C_7=C_6/4$, we
obtain (\ref{restion on delta in theorem5a}) and (\ref{restion on
delta in theorem5b}). Then, it follows from
 Theorem
\ref{Theorem:rate for l} that with confidence $1-\delta$, there
holds
\begin{eqnarray*}
    &&\|\overline{f}_{D,\lambda}^{\ell}-
    f_{D,\lambda}\|_\rho
     \leq  2\tilde{C}_3
     |D|^\frac{-r}{2r+s}\log^{\ell+2}\frac{4}\delta.
\end{eqnarray*}
Hence, we obtain from (\ref{estimate for KRR}) and (\ref{bound
ADlambda}) that with confidence $1-\delta$, there holds
\begin{eqnarray*}
     &&\|\overline{f}_{D,\lambda}^{\ell}-
    f_\rho\|_\rho\leq  \|\overline{f}_{D,\lambda}^{\ell}-
    f_{D,\lambda}\|_\rho+\|f_{D,\lambda}-f_\rho\|_\rho\\
    &\leq&
     C_8|D|^\frac{-r}{2r+s}\log^{\ell+2}\frac{4}\delta,
\end{eqnarray*}
where $C_8=1+\tilde{C}_1+2\tilde{C}_3,$ which completes the proof of
Theorem \ref{Theorem:optimal l}. $\Box$

{\bf Proof of Corollary \ref{Corollary: almost surely convergence
l}.} The proof of   Corollary \ref{Corollary: almost surely
convergence l} is the same as that of  Corollary \ref{Corollary:
almost surely convergence} with (\ref{Restiction on m for DKRR in
prob})  replaced by (\ref{Restiction on m for l}).  $\Box$

\section{Experiments}\label{Sec.Experiments}
In this section, we report numerical results to verify our theoretical statements. We employ  {three} criteria for
comparisons. The first criterion is the global mean squared error (GMSE) which is the mean square error (MSE)
of a testing set with training all samples in a batch mode. GMSE provides a baseline to assess the  {performances of DKRR and DKRR($\ell$)}.
The second criterion is the average error (AE) which is the MSE of DKRR. The third criterion is the average
error with communications (AEC), which is the MSE of  DKRR ($\ell$). Regularization parameters in all experiments are selected by grid search.

We carry out three simulations to verify our theoretical statements. The first simulation is devoted to
illustrating the power of communications in DKRR. The second simulation is performed to
demonstrate the relation between the generalization ability of DKRR($\ell$) and the number of training samples for
fixed numbers of local machines and communications.
The third simulation focuses on  {comparisons on the training complexities of DKRR and DKRR($\ell$).}

Before carrying out simulations, we describe how the synthetic data is generated.
The inputs $\{\bm{x}_i\}_{i=1}^N$ of training samples are independently
drawn according to the uniform distribution on the (hyper-)cube $[0,1]^d$ with $d=1$ or $d=3$. The corresponding outputs
$\{y_i\}_{i=1}^N$ are generated from the regression models $y_i=g_j(\bm{x}_i)+\varepsilon_i$ for $i=1,2,\cdots,N$ and $j=1,2$, where
$\varepsilon_i$ is the independent Gaussian noise $\mathcal{N}(0,0.2)$,
\begin{equation}\label{g1}
g_1(x)=\left\{
\begin{array}{ll}
x & \quad\mbox{if} ~ 0<x\leq 0.5, \\
1-x & \quad \mbox{if} ~ 0.5 < 0 \leq 1,
\end{array}
\right.
\end{equation}
for 1-dimensional data, and
\begin{equation}\label{g2}
g_2(\bm{x})=\left\{
\begin{array}{ll}
(1-\|\bm{x}\|_2)^6(35\|\bm{x}\|_2^2+18\|\bm{x}\|_2+3) & \quad\mbox{if} ~ 0<\|\bm{x}\|_2\leq 1, \\
0 & \quad \mbox{if} ~ \|\bm{x}\|_2> 1,
\end{array}
\right.
\end{equation}
for 3-dimensional data. The inputs $\{\bm{x}_i'\}_{i=1}^{N'}$ of testing samples are also drawn independently according to
the uniform distribution on the (hyper-)cube $[0,1]^d$ \textcolor[rgb]{1,0,0}{but} the corresponding outputs $\{y_i'\}_{i=1}^{N'}$ are generated  by
$y_i'=g_j(\bm{x}_i')$. It can be found in \citep{Wu1995,Schaback2006} that $g_1\in W_1^1$ and $g_2\in W_3^4$, where $W_d^\alpha$ represents
the $\alpha$-order Sobolev space on $[0,1]^d$. If we define $K_1(x,x')=1+\min(x,x')$ and $K_2(\bm{x},\bm{x}')=h(\|\bm{x}-\bm{x}'\|_2)$
with
\begin{equation}\label{h}
h(r)=\left\{
\begin{array}{ll}
(1-r)^4(4r+1) & \quad\mbox{if} ~ 0<r \leq 1, \\
0 & \quad \mbox{if} ~ r> 1,
\end{array}
\right.
\end{equation}
then we know \citep{Wu1995,Schaback2006} that $K_1$ and $K_2$ are reproducing kernels for $W_1^1$ and $W_3^2$, respectively. Obviously,
$g_1\in\mathcal{H}_{K_1}$ and $g_2\in \mathcal{H}_{K_2}$. In the training process of DKRR and DKRR($\ell$), we uniformly distribute $N$ training samples
to $m$ local machines.
% ---------Figure 4----------------
% ---------The role of Communication----------------
\begin{figure*}[t]
    \centering
	 \subfigure{\includegraphics[width=6cm,height=5cm]{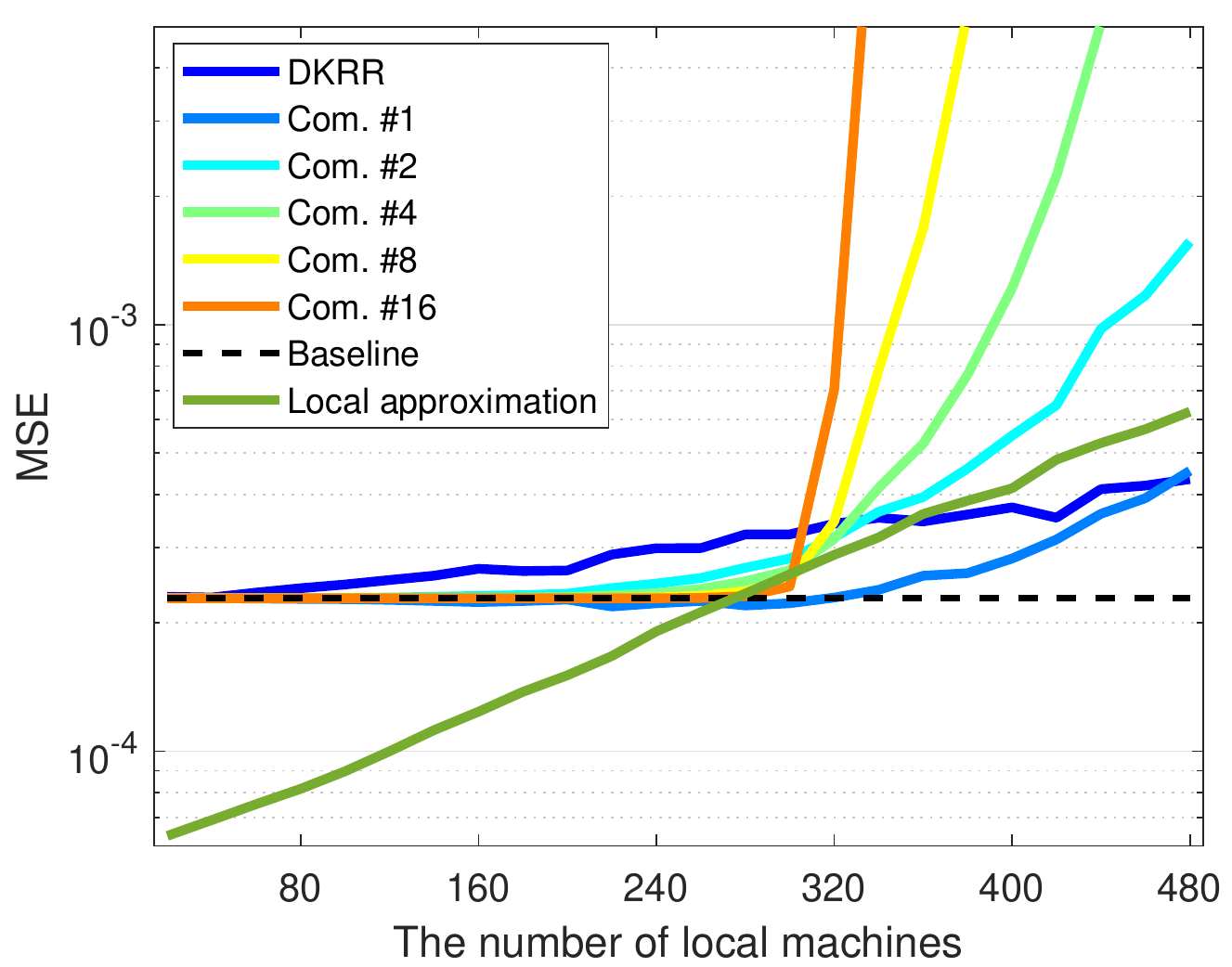}}
    \subfigure{\includegraphics[width=6cm,height=5cm]{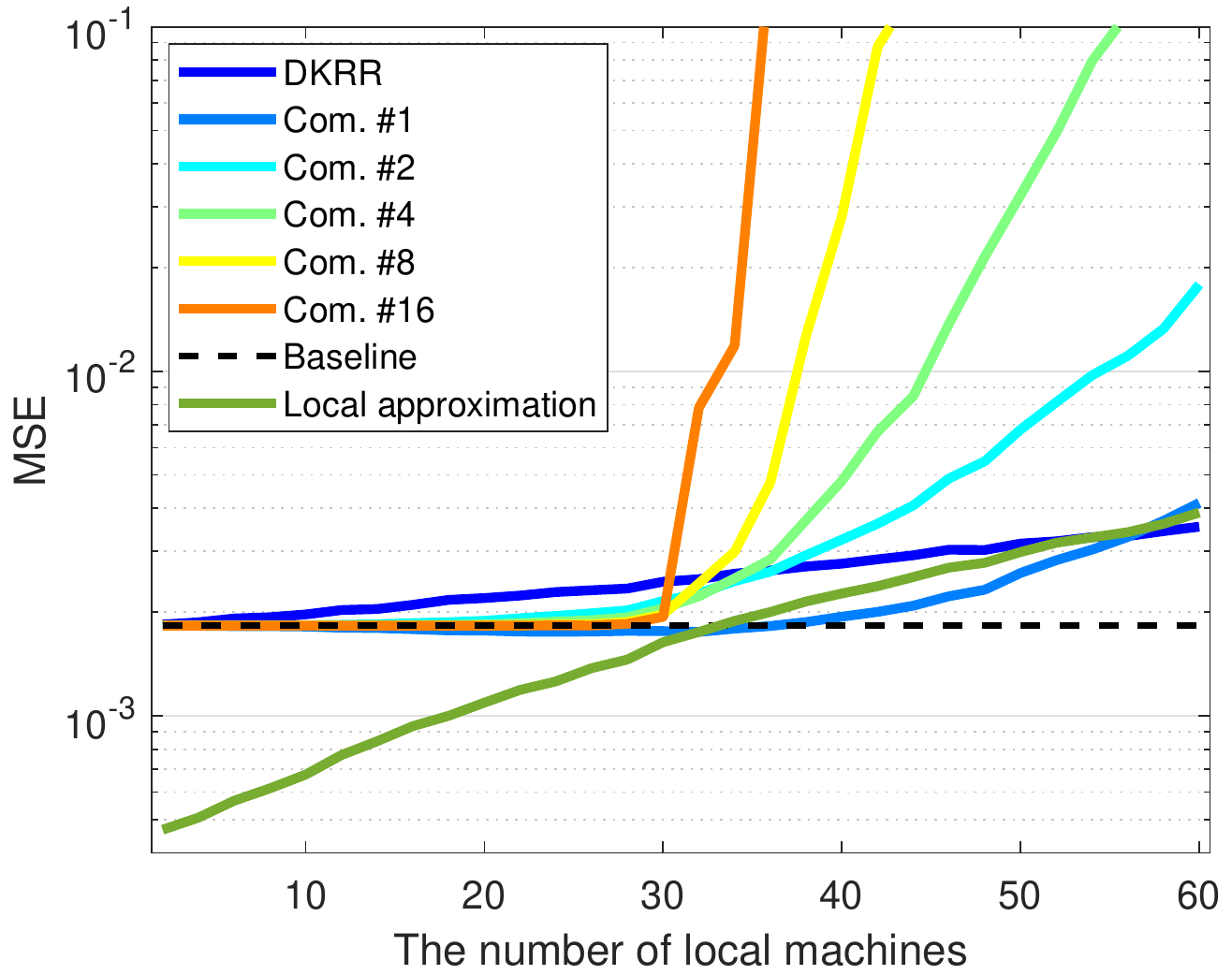}}
	\caption{The relation between MSE and the number of local machines for fixed  numbers of communications. The left figure and right figure
are respectively the results on the 1-dimensional data and the 3-dimensional data. `Com. \#' represents the number of communications. }\label{m_MSE}
\end{figure*}
% ---------Figure 5----------------
% ---------convergence of Communication----------------
\begin{figure*}[t]
    \centering
	 \subfigure{\includegraphics[width=6cm,height=5cm]{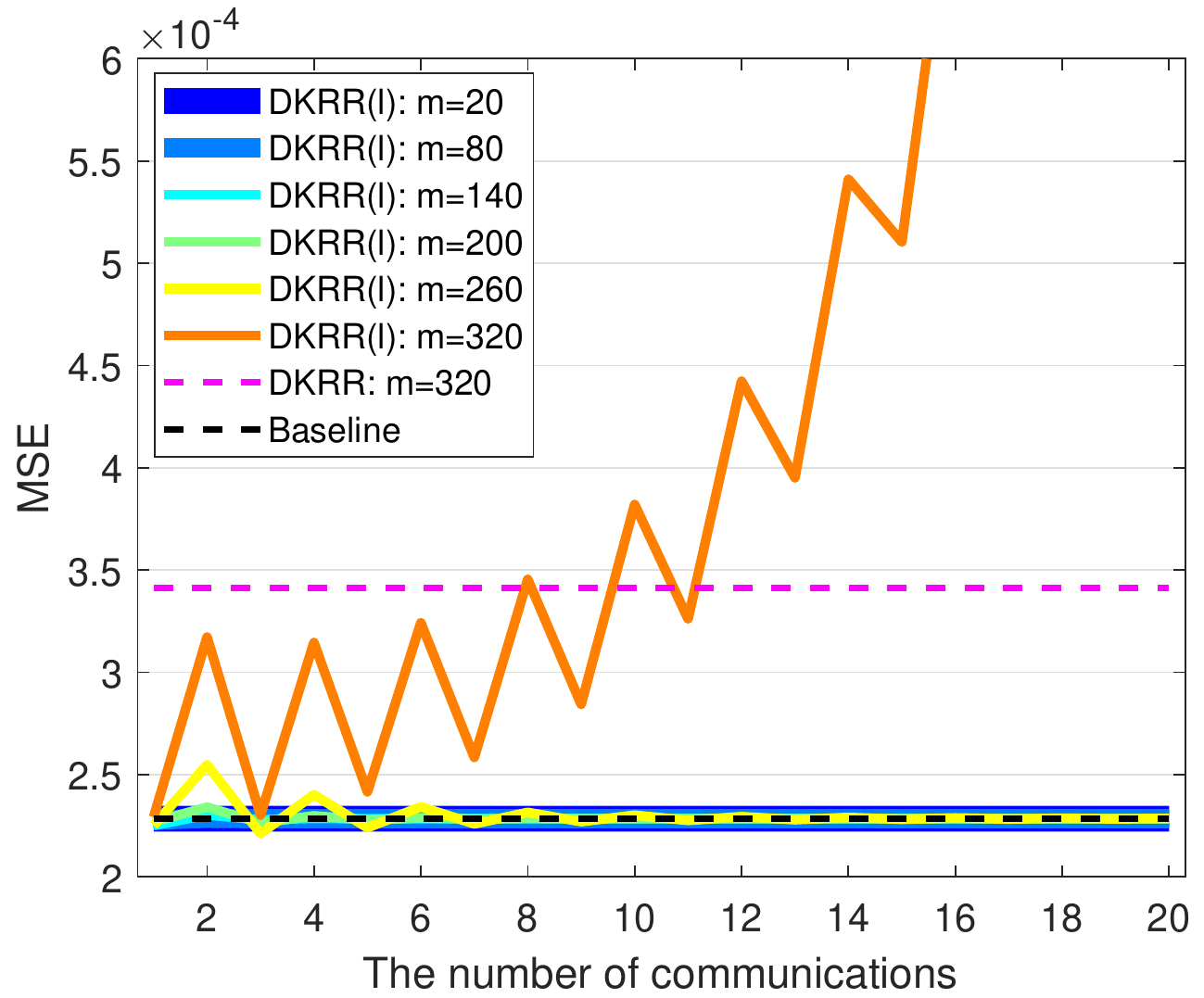}}
    \subfigure{\includegraphics[width=6cm,height=5cm]{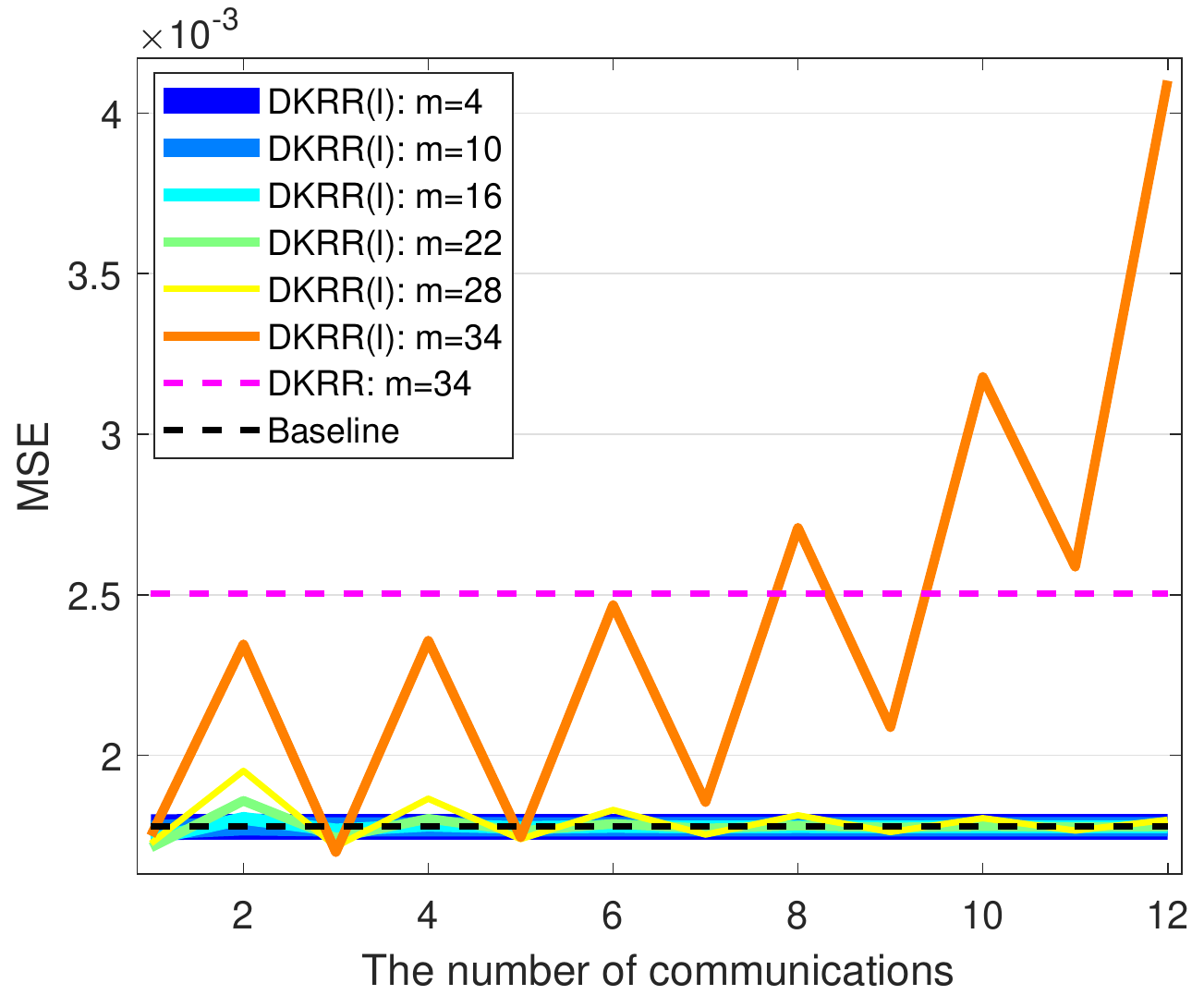}}
	\caption{ The relation between MSE and the number of communications for fixed numbers of local machines. The left figure and right figure
are respectively the results on the 1-dimensional data and the 3-dimensional data.}\label{Com_MSE}
\end{figure*}

{\bf Simulation 1:} We generate 10000 samples for training and 1000 samples for testing. The number $m$ of local machines
varies from  $\{20,40,60,\cdots,480\}$ for 1-dimensional data, and varies from $\{2,4,6,\cdots,60\}$ for 3-dimensional
data. The testing results are shown in Figure \ref{m_MSE} and Figure \ref{Com_MSE}.
Figure \ref{m_MSE} shows the relation between MSE and the number of local machines by different numbers of communications.
From Figure \ref{m_MSE}, we can conclude the following four assertions: 1) When $m$ is not too large, AEs are always comparable to GMSEs.
There exists an upper bound of $m$, denoted by $m_{\mbox{\scriptsize{B}}}$ (e.g., $m_{\mbox{\scriptsize{B}}}\approx 40$ for $d=1$ and $m_{\mbox{\scriptsize{B}}}\approx 6$ for $d=3$), larger than which AE curves
increase dramatically and far from the GMSE curves. This verifies {the theoretical statement in Theorem \ref{Theorem:optimal dkrr in probability}}. 2) AECs also have the upper bound, such as $m_{\mbox{\scriptsize{B}}}\approx 200$ for $d=1$ and $\ell=2$,
which is much larger than that of AEs. This result confirms  Theorem \ref{Theorem:optimal l} by showing the power of communications in the sense that communications in DKRR can help to relax
the restriction on $m$.
3) The upper bound $m_{\mbox{\scriptsize{B}}}$ of AECs increases with the number of communication increasing, which implies the necessity of communications and verifies Theorem \ref{Theorem:optimal l} by showing that  the upper bound in (\ref{Restiction on m for l}) is monotonically increasing with respect to the number of communications.
4) Different from DKRR which cannot sufficiently embody the best approximation ability of local estimator, DKRR($\ell$) succeeds in presenting an upper bound of $m$ which is  close to  $m_1$ in Figure \ref{Fig:Motivation}.

% ---------Figure 6------------
\begin{figure*}[t]
    \centering	 \subfigure{\includegraphics[width=4.7cm,height=3.7cm]{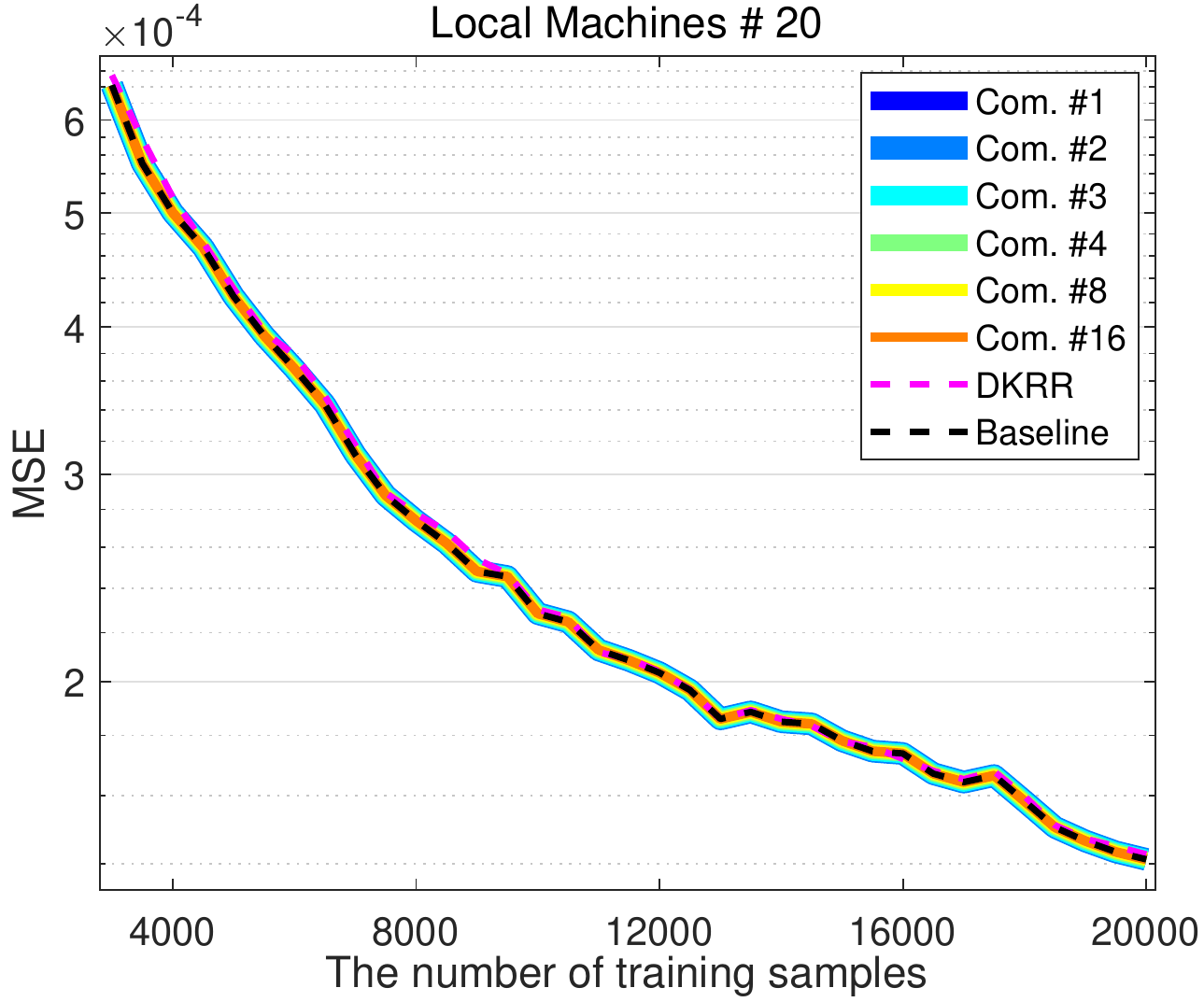}}
    \subfigure{\includegraphics[width=4.7cm,height=3.7cm]{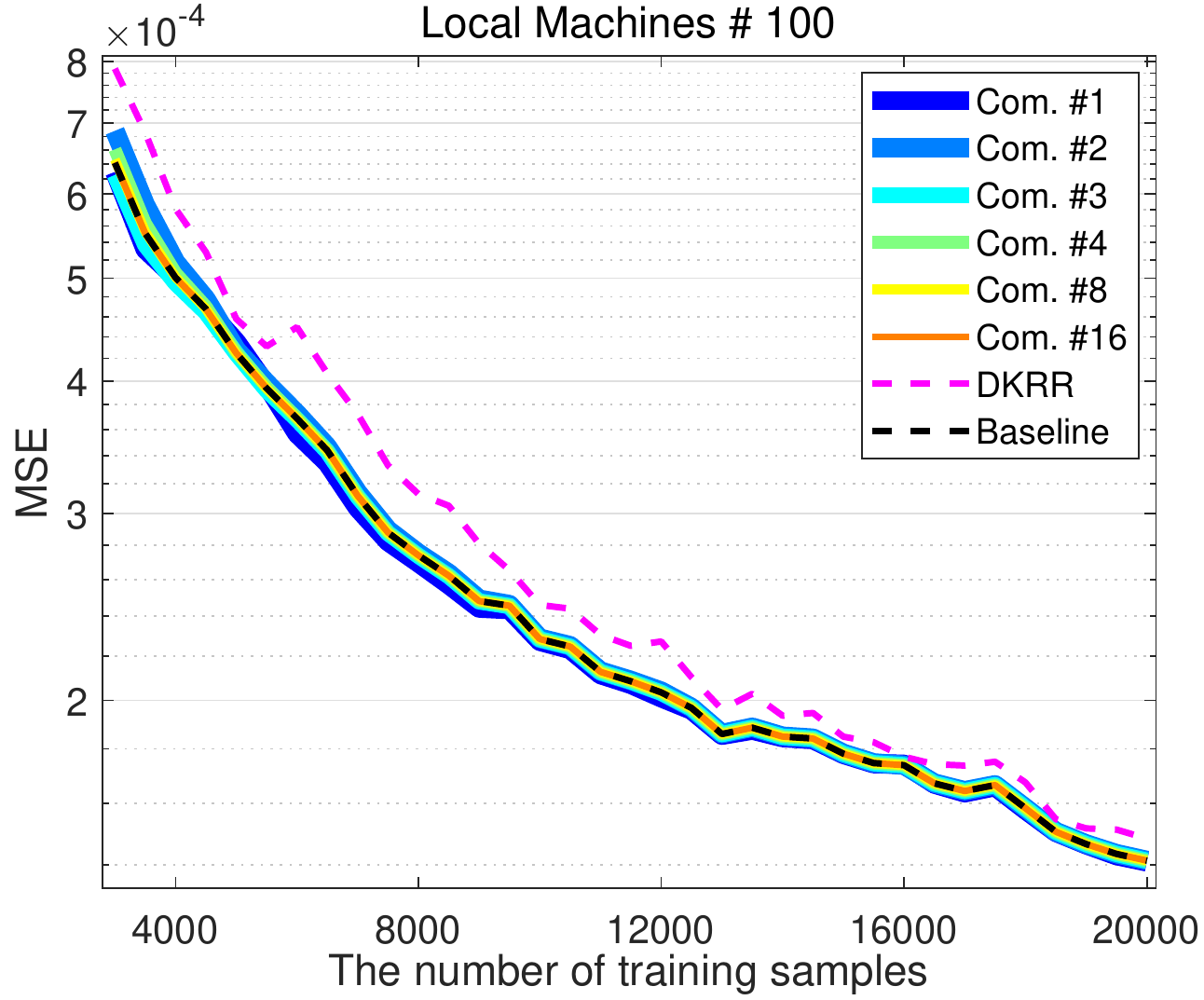}}
    \subfigure{\includegraphics[width=4.7cm,height=3.7cm]{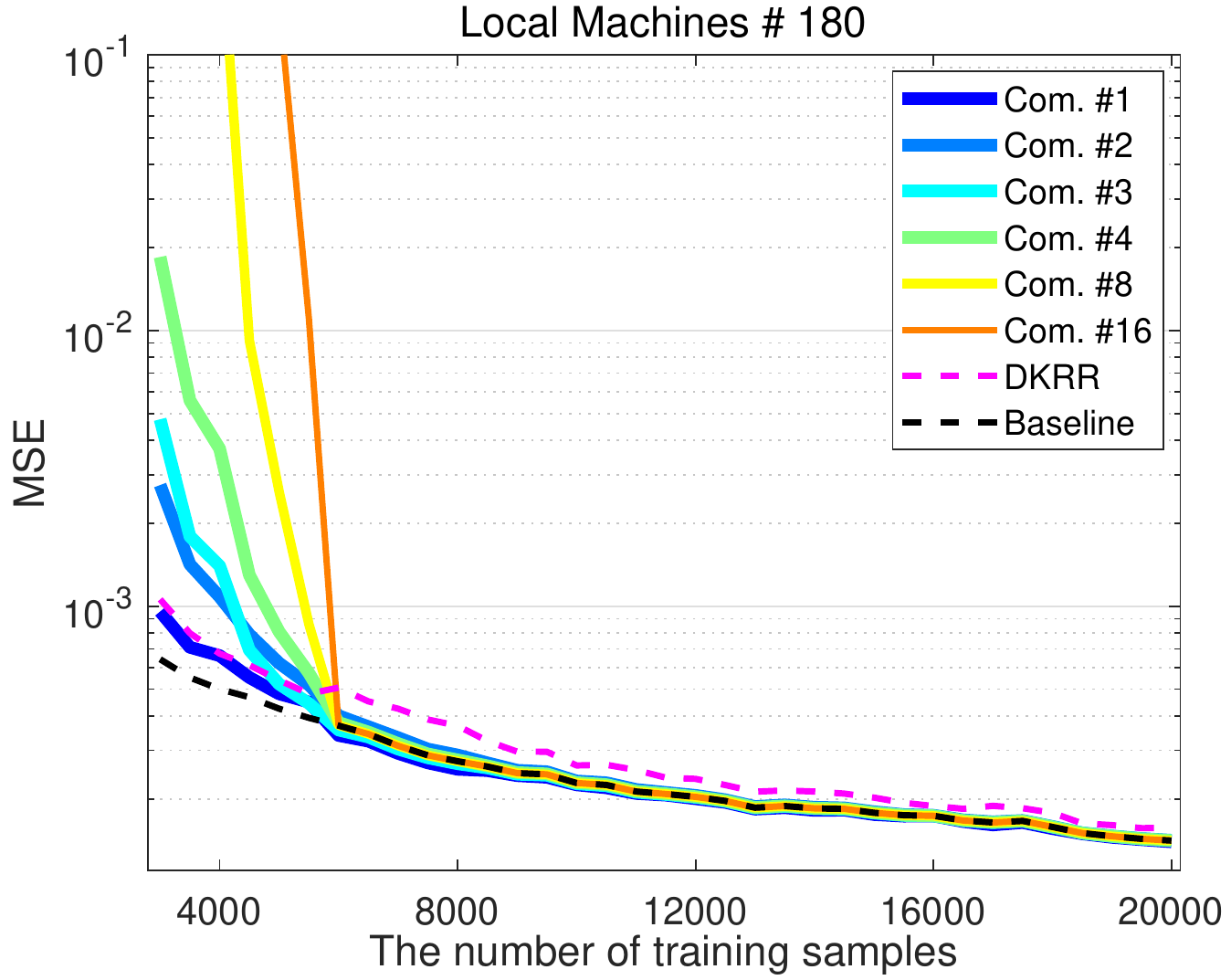}}\\
    \subfigure{\includegraphics[width=4.7cm,height=3.7cm]{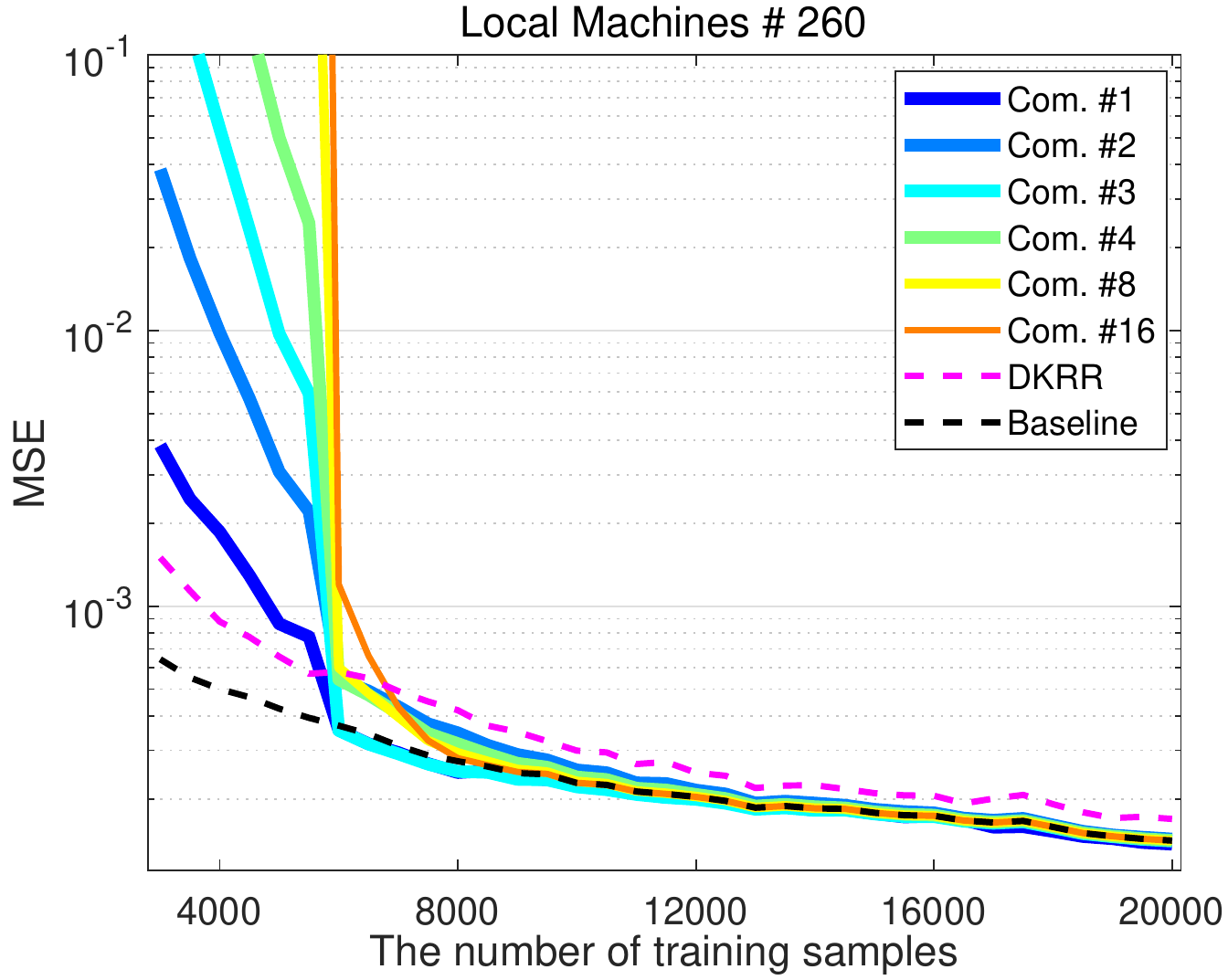}}
    \subfigure{\includegraphics[width=4.7cm,height=3.7cm]{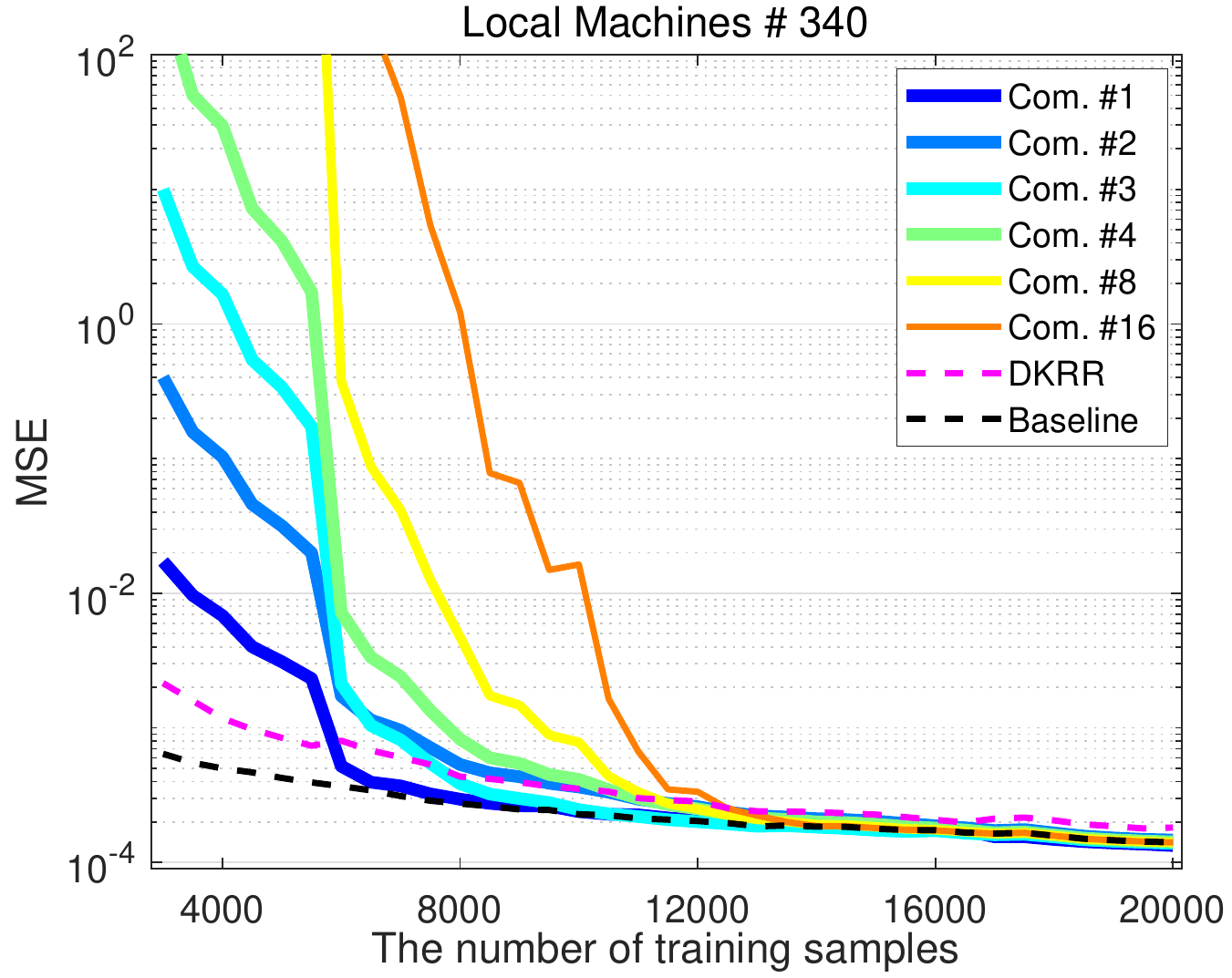}}
    \subfigure{\includegraphics[width=4.7cm,height=3.7cm]{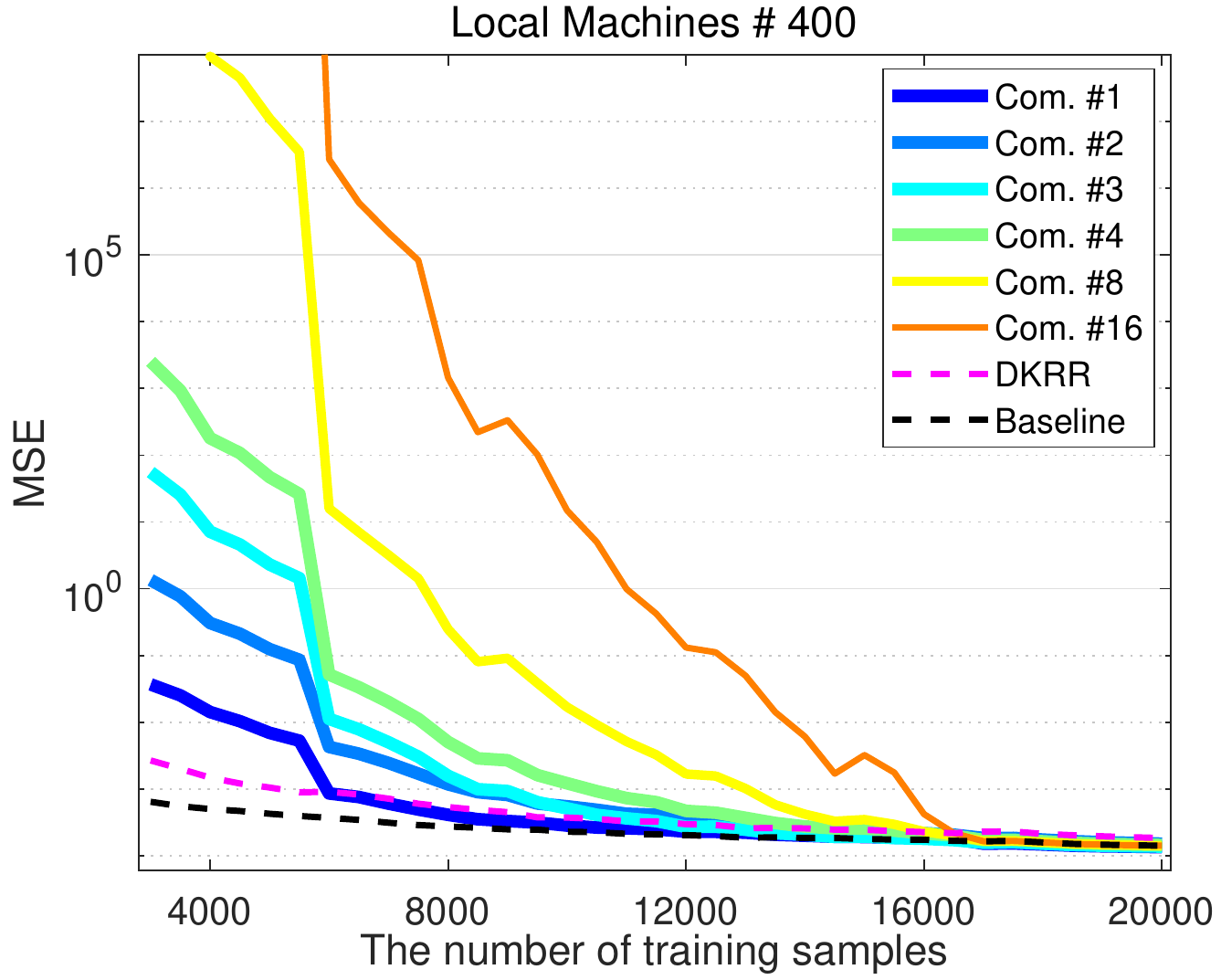}}
    \vspace{-0.1in}
	\caption{ The relation between the number of training samples and MSE on the 1-dimensional data.}\label{MSE_TrNum1}
\end{figure*}

% ---------Figure 7------------
\begin{figure*}[t]
    \centering
 \subfigure{\includegraphics[width=4.7cm,height=3.7cm]{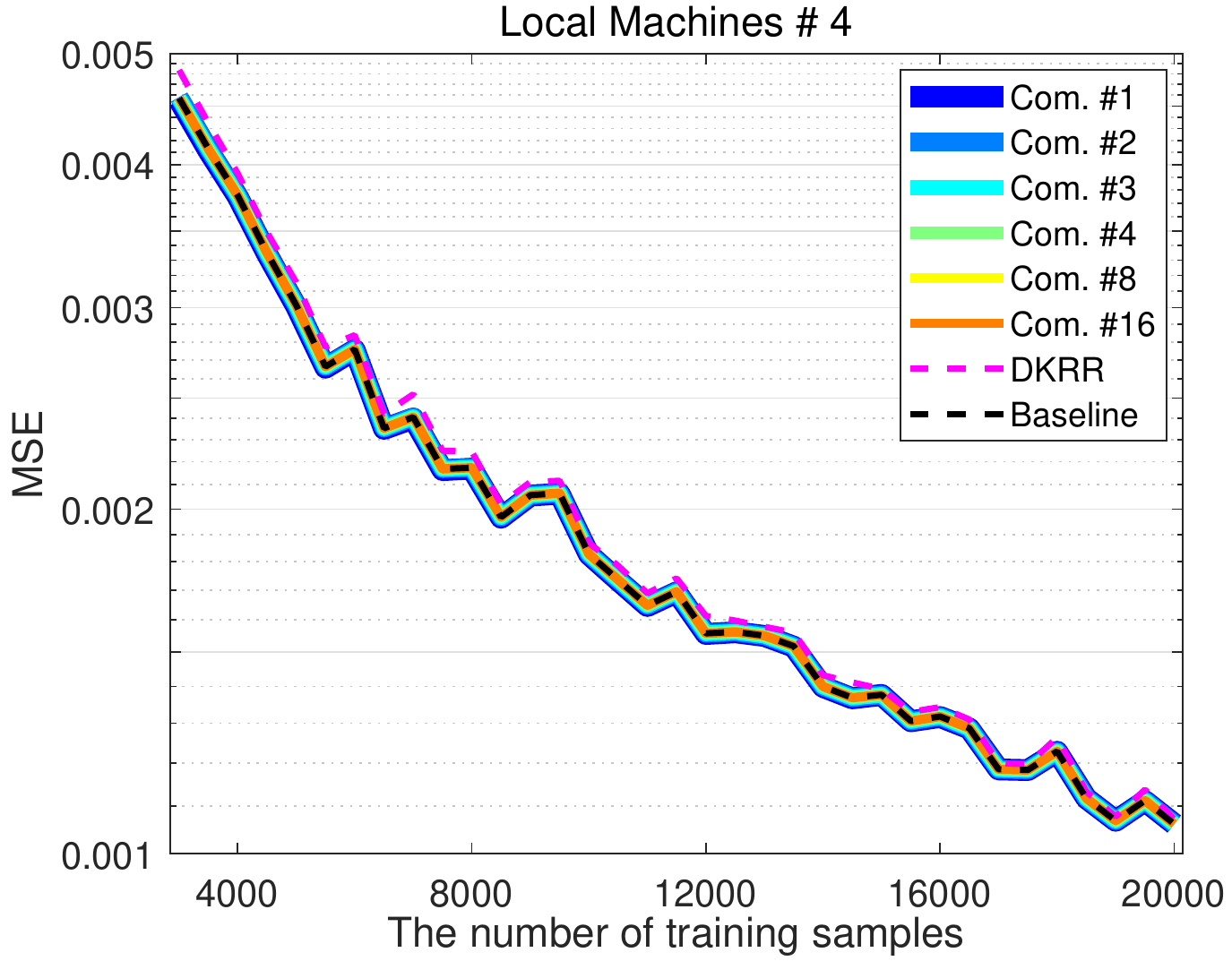}}
    \subfigure{\includegraphics[width=4.7cm,height=3.7cm]{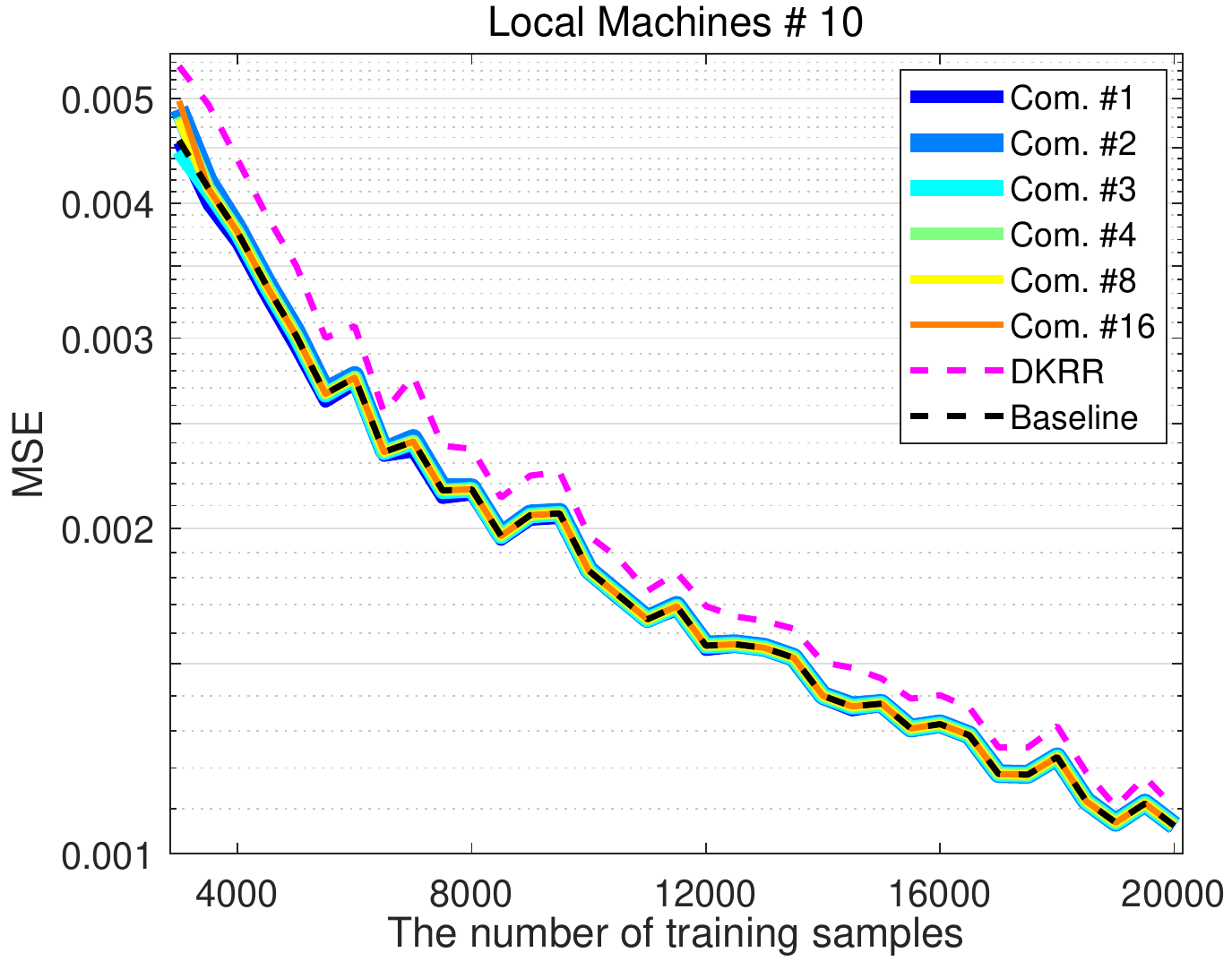}}
    \subfigure{\includegraphics[width=4.7cm,height=3.7cm]{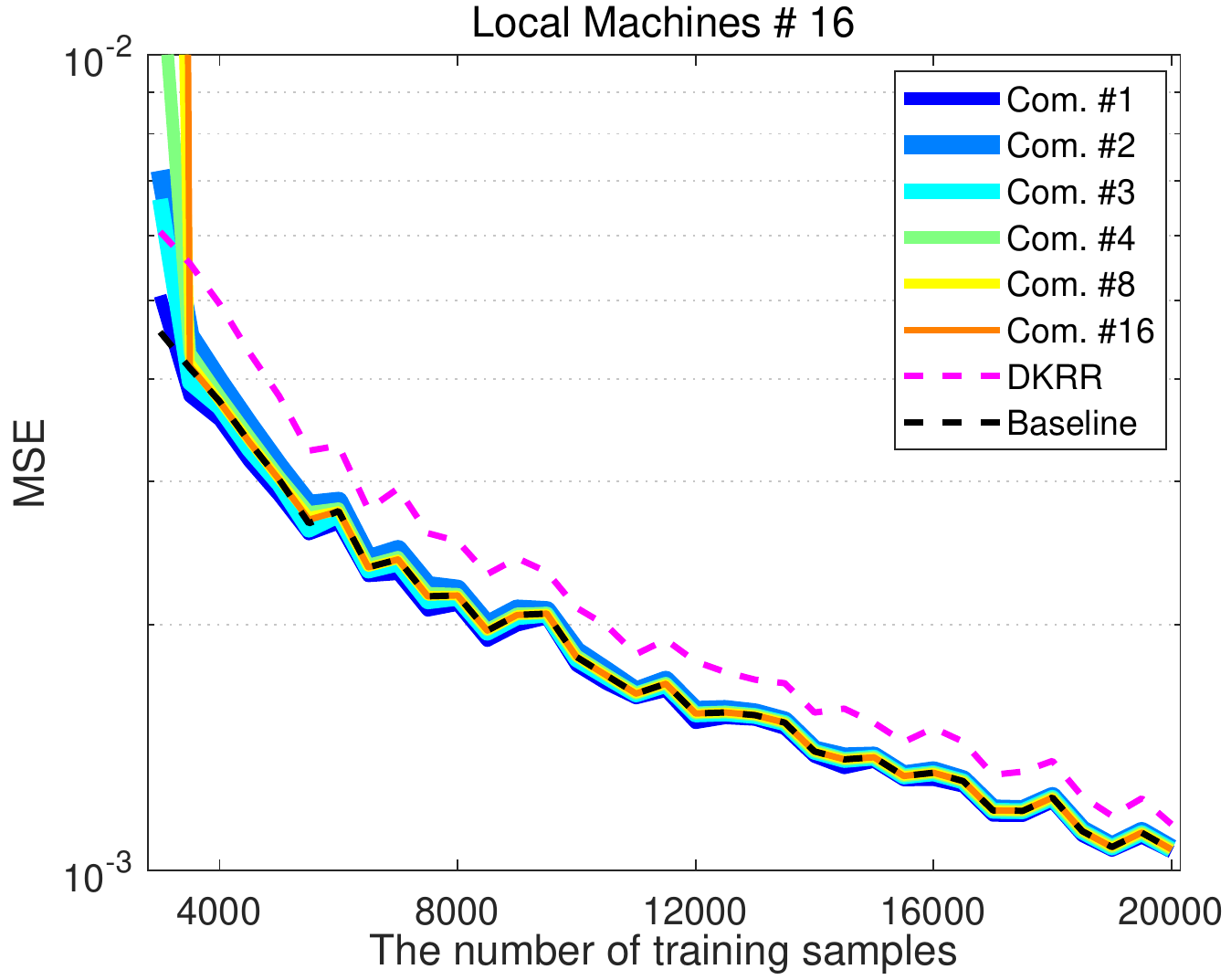}}\\
    \subfigure{\includegraphics[width=4.7cm,height=3.7cm]{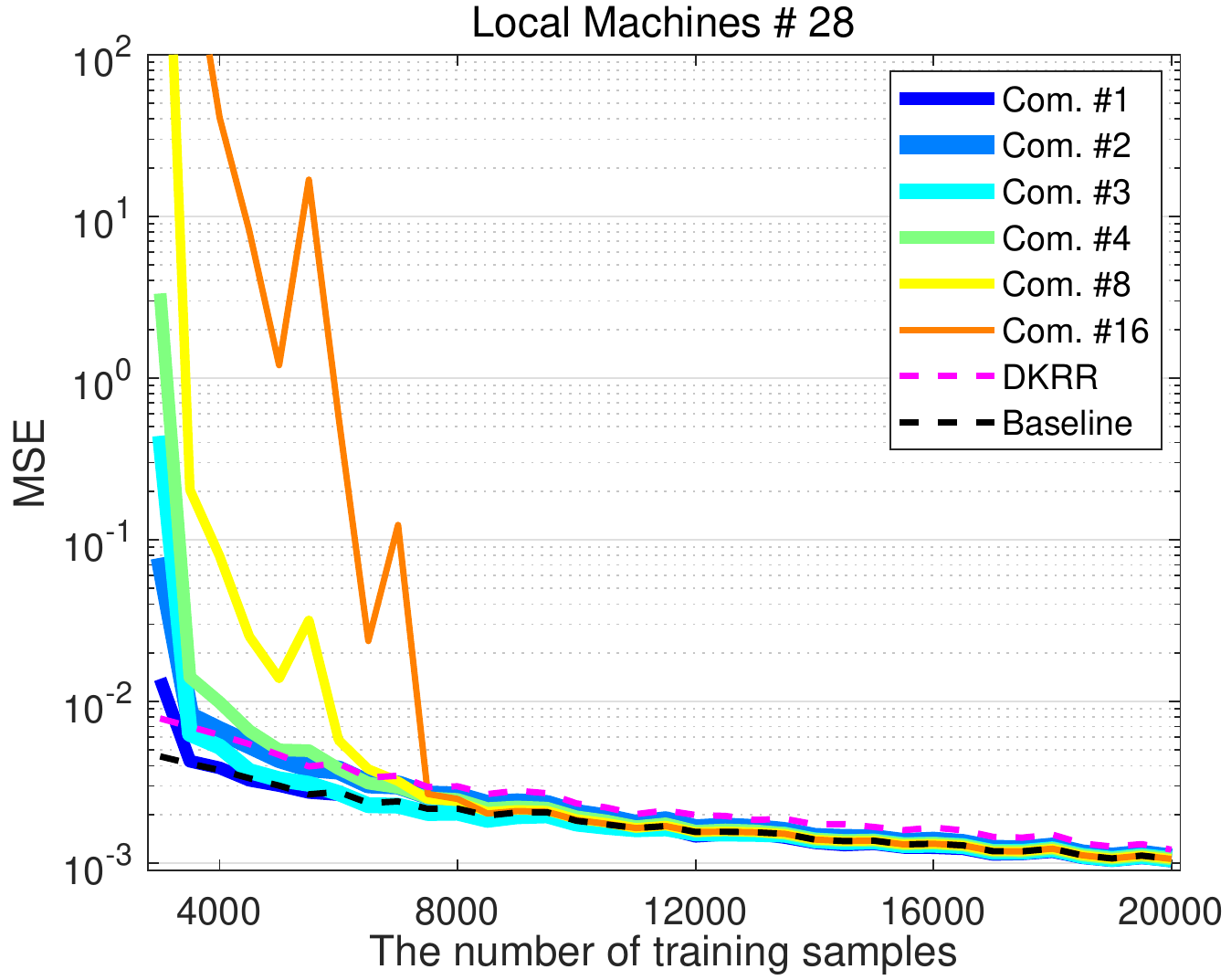}}
    \subfigure{\includegraphics[width=4.7cm,height=3.7cm]{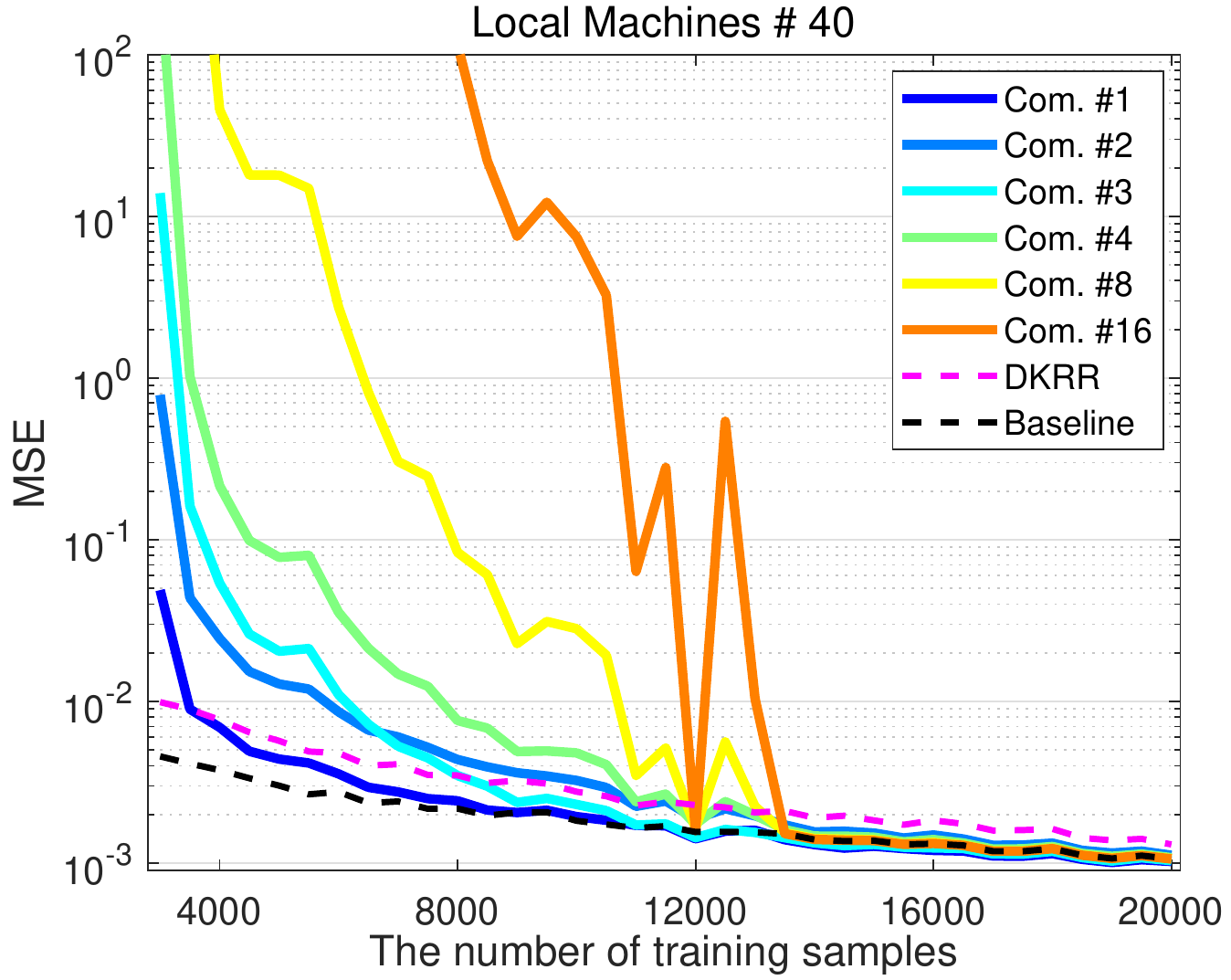}}
    \subfigure{\includegraphics[width=4.7cm,height=3.7cm]{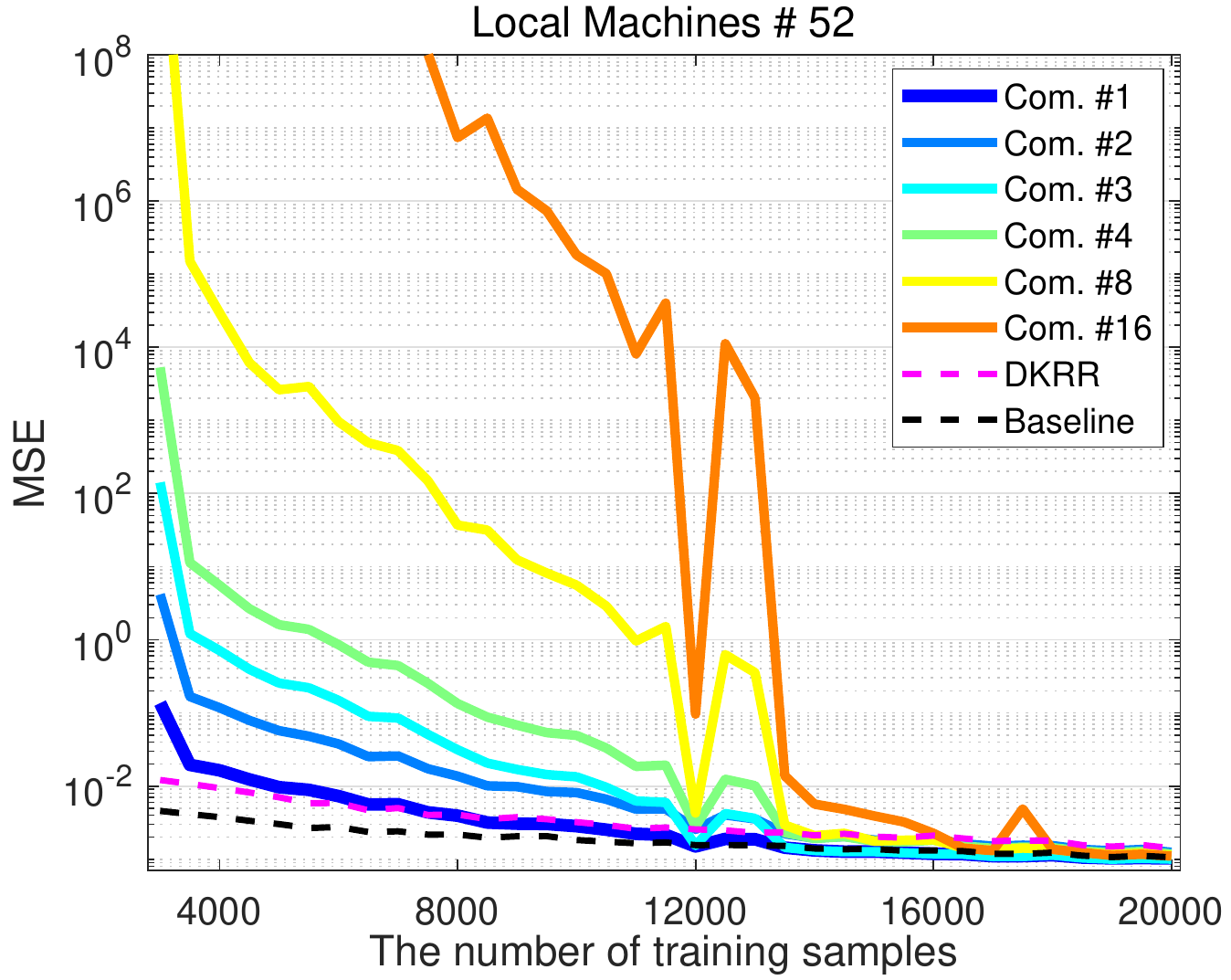}}
\vspace{-0.1in}	
\caption{ The relation between the number of training samples and MSE on the 3-dimensional data.}\label{MSE_TrNum2}
\end{figure*}

The relation between MSE and the number of communications by different  numbers of local machines is illustrated in Figure \ref{Com_MSE}.
We can see that there exists an upper bound of $m$ (e.g., about $280$ for 1-dimensional data and $30$ for 3-dimensional data), less than which AECs are
guaranteed to converge to GMSEs with a fast rate, and larger than which AECs diverge dramatically.
This verifies Proposition \ref{Proposition: error decomposition for l} by comparing
$\sum_{j=1}^m\frac{|D_j|}{|D|}\mathcal Q_{D_j,\lambda}^2 (\mathcal R_{D_j,\lambda}+\mathcal R_{D,\lambda})$ with 1.

{\bf Simulation 2:} This simulation investigates the generalization performance of DKRR and DKRR($\ell$) with
varying the number of training samples $N\in \{3000,3500,4000,$ $\cdots,20000\}$
and fixing the number of testing samples $N'=1000$. The regression results are shown
in Figure \ref{MSE_TrNum1} and Figure \ref{MSE_TrNum2}, from which it can be drawn the following conclusions:
1) For a fixed number of local machines and a fixed number  of communications, AE curves are getting closer and closer
to GMSE curves, and AEC curves are converging to GMSE curves with the number of training samples increasing.
This verifies our theoretical assertions of Theorem \ref{Theorem:optimal dkrr in probability} and Theorem \ref{Theorem:optimal l}.
2) For each fixed number of local machines, there exists a lower bound for the number of training
samples, denoted by $N_{\mbox{\scriptsize{B}}}$ (e.g., $N_{\mbox{\scriptsize{B}}}\approx 6000$ for $d=1$ and $m=180$, and $N_{\mbox{\scriptsize{B}}}\approx 13500$ for
$d=3$ and $m=40$), larger than which, the generalization
performance of DKRR($\ell$) is significantly superior than that of DKRR, and AECs converge to
GMSEs with the number of communications increasing. Obviously, the bound $N_{\mbox{\scriptsize{B}}}$ increases
as the number of local machines increases. This also confirms the conclusion
in Theorem \ref{Theorem:optimal l}.

% ---------Table 1------------
\begin{table*}[t]
\renewcommand\arraystretch{1.3}
%\scriptsize
\small
\centering
\caption{ {Computational complexity of the training flow in Appendix B. `INDV' represents the complexity of calculating
each individual item, and `TOT' represents the total complexity of Step 1, Step 2 and the loop from Step 4 to Step 7. $\tau$ represents the complexity of calculating a value of a kernel function $K(\cdot,\cdot)$.}
}
\label{Tab01}
\begin{tabular}{
                        *{1}{p{0.75cm}<{\centering}}
				        *{1}{|p{1.0cm}<{\centering}}
				        *{1}{|p{0.8cm}<{\centering}}
				        *{1}{|p{0.4cm}<{\centering}}
                        *{1}{|p{1.6cm}<{\centering}}
				        *{1}{|p{1.0cm}<{\centering}}
				        *{1}{|p{1.2cm}<{\centering}}
				        *{1}{|p{1.1cm}<{\centering}}
				        *{1}{|p{0.4cm}<{\centering}}
				        *{1}{|p{1.2cm}<{\centering}}
                        *{1}{|p{1.05cm}<{\centering}}
			           }
\toprule[1.5pt]
\multirow{2}{*}{Item} & \multicolumn{4}{c|}{Step 1}  & Step2 & Step4 & Step5  & \multicolumn{2}{c|}{Step6} & Step7 \\
\cline{2-11}
&  $\mathbb{K}_{D_k,D_j}$  &  $M_{D_j,\lambda}$ &   $\vec{\alpha}_{D_j}$ & $\mathbb{K}_{D_k,D_j}\vec{\alpha}_{D_j}$
&   $\vec{f}_{D,\lambda,D_k}^0$  &  $\vec{G}_{D_j,\lambda,k,\ell}$   &  $\vec{G}_{D,\lambda,k,\ell}$
&  $\vec{\beta}_{j,\ell}$  & $\vec{H}_{D_j,\lambda,k,\ell}$  &  $\vec{f}_{D,\lambda,D_k}^\ell$  \\
\hline
\hline
INDV & $\frac{N^2\tau}{m^2}$ & $\frac{N^3}{m^3}$ & $\frac{N^2}{m^2}$  & $\frac{N^2}{m^2}$ & $N$ & $\frac{N^2}{m^2}$
& $N$ & $\frac{N^2}{m^2}$ & $\frac{N^2}{m^2}$  & $N$ \\
\hline
TOT & \multicolumn{4}{c|}{$O\left( \frac{N^2\tau}{m}+\frac{N^3}{m^3}\right)$}
& $O(mN)$ &\multicolumn{5}{c}{$O\left(mN\ell+\frac{N^2\ell}{m}\right)$} \\
\bottomrule[1.5pt]
\end{tabular}
\end{table*}

{\bf Simulation 3:} {In this simulation, we compare the training complexities of DKRR and DKRR($\ell$).
The computational complexity analysis for the training flow in Appendix B is given in Table \ref{Tab01}, in which the computational
complexity of Step 1, Step 4 and Step 6 refers to the complexity of each local machine, provided that parallelization is
implemented. It should be noted that the time of data transmission in DKRR($\ell$) is not considered, and the computational complexity
of Step 1 is also that of DKRR. Thus, the training complexity of DKRR and DKRR($\ell$) are $O(N^2\tau/m+N^3/m^3)$
and $O(N^2\tau/m+N^3/m^3+N^2\ell/m+mN\ell)$, respectively. We define
\begin{equation*}
\Omega_{\mbox{\scriptsize{DKRR}}} = N^2\tau/m+N^3/m^3,~~\Omega_{\scriptsize{\mbox{DKRR}(\ell)}} = N^2\tau/m+N^3/m^3+N^2\ell/m+mN\ell~.
\end{equation*}
It can be seen that $\Omega_{\mbox{\scriptsize{DKRR}}}$ monotonously decreases with $m$ increasing; $\Omega_{\scriptsize{\mbox{DKRR}(\ell)}}$
first decreases and then increases with $m$ increasing, and the number $m$ with minimum of $\Omega_{\scriptsize{\mbox{DKRR}(\ell)}}$
is given by
\begin{equation}\label{mstar}
m^\star = \sqrt{\frac{\sqrt{(\tau+\ell)^2+12\ell}+(\tau+\ell)}{2\ell}}\sqrt{N}~.
\end{equation}
Once data and the kernel function are given, the complexity $\tau$ of calculating a kernel value can be considered as a constant.
Provided $\ell$ is large enough, we have $m^\star\approx\sqrt{N}$, from which a rough conclusion can be drawn as follows: the training complexity of DKRR($\ell$) mainly focuses on the local process (i.e., Step 1 of the training flow in Appendix B) when $m<\sqrt{N}$, while it mainly focuses on the communications (i.e., from Step 4 to Step 7 of the training flow in Appendix B)
when $m>\sqrt{N}$. In the following, some numerical results are reported to verify the above analysis of training complexity.}
% ---------Figure 8----------------
% ---------The number m_B----------------
\begin{figure*}[t]
    \centering \subfigure{\includegraphics[width=6cm,height=5cm]{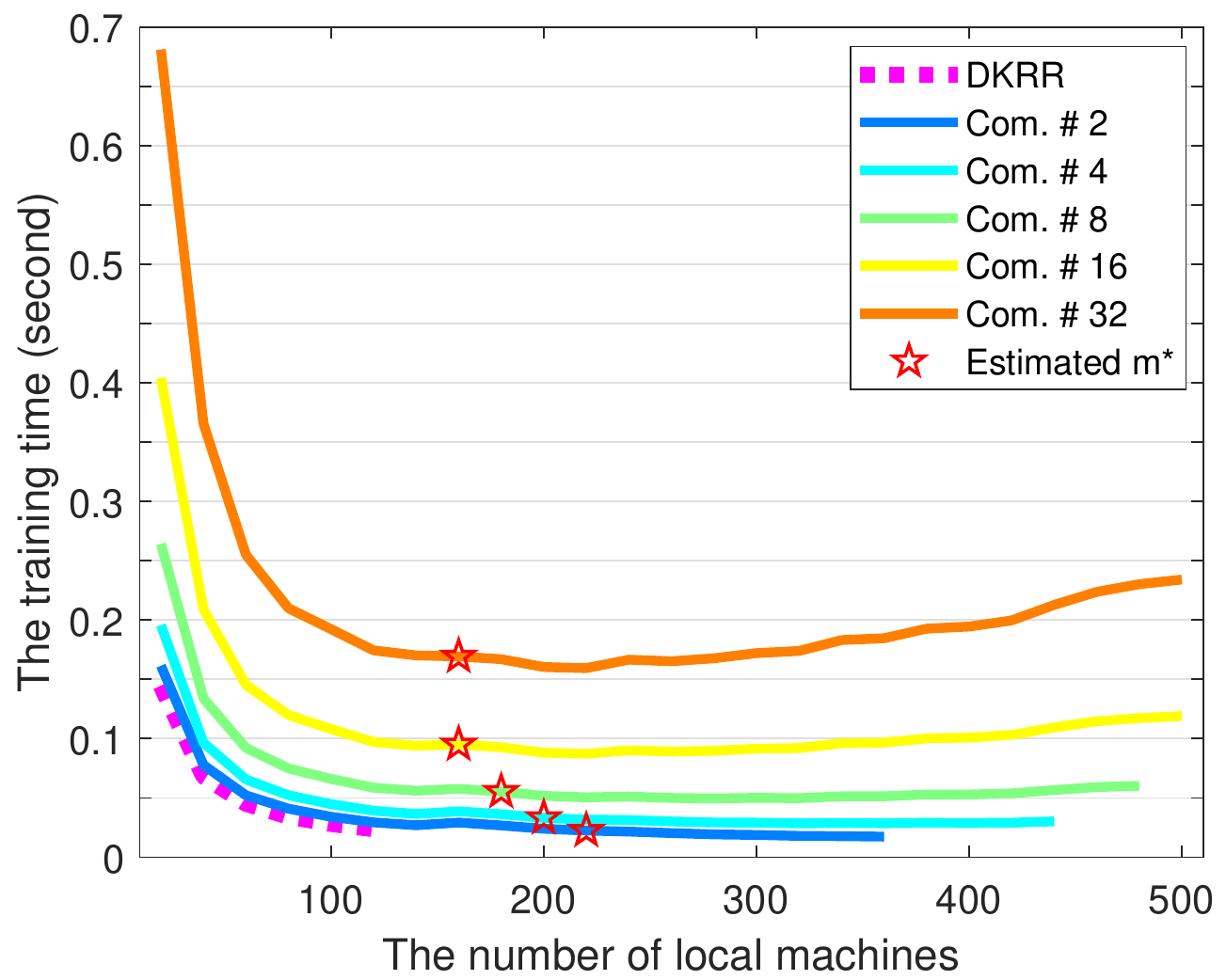}}
    \subfigure{\includegraphics[width=6cm,height=5cm]{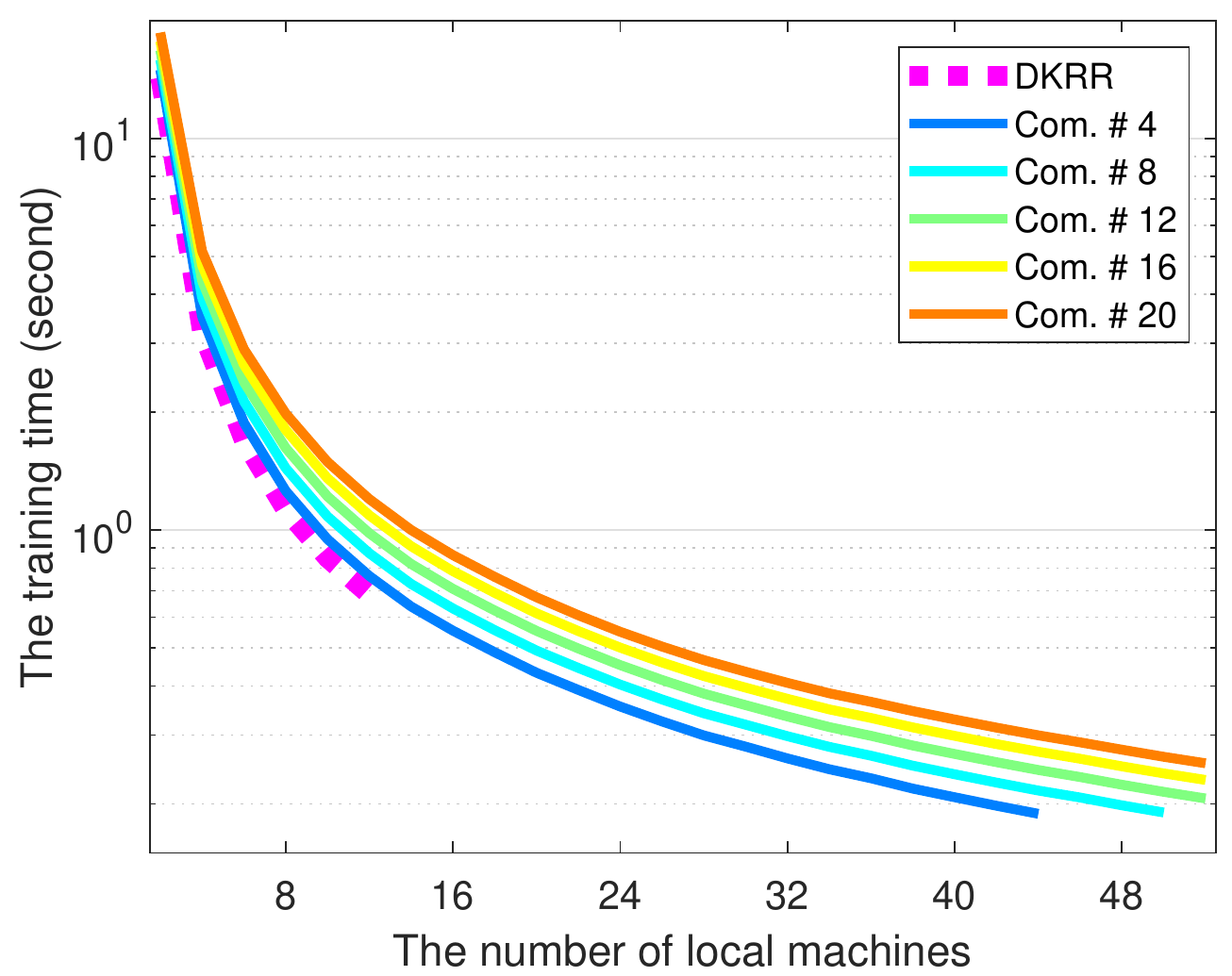}}
	\caption{ {The relation between the number of local machines and the training time for fixed numbers of communications.
The left figure and right figure are respectively the results on the 1-dimensional data and the 3-dimensional data with $N=20000$
training samples.}}\label{TrTime20000}
\end{figure*}

 {We start by introducing some notations that are used below.
$\mbox{AE}_{N,m}$ represents the MSE of DKRR with $m$ local machines for $N$ training samples.
% $\mbox{MSE}_{N,m,\ell}^{\scriptsize{\mbox{DKRR}(\ell)}}$
$\mbox{AEC}_{N,m,\ell}$ represents the MSE of DKRR($\ell$) with $m$ local machines and $\ell$ communications for $N$ training samples.
$\mbox{GMSE}_{N}$ represents the MSE of the model with training $N$ samples in a batch mode. The relative errors for DKRR and DKRR($\ell$)
are defined by
$\mbox{RE}_{N,m}=\frac{\left|\mbox{AE}_{N,m}-\mbox{GMSE}_{N}\right|}{\mbox{GMSE}_{N}}$
and
$\mbox{REC}_{N,m,\ell}=\frac{\left|\mbox{AEC}_{N,m,\ell}-\mbox{GMSE}_{N}\right|}{\mbox{GMSE}_{N}}$, respectively. In DKRR with $N$ training samples, we denote the maximum number $m$ of local machines that satisfies $\mbox{RE}_{N,m}<\varepsilon$ as $\bar{m}_{\mbox{\scriptsize{B}}}^N$, i.e.,
$\bar{m}_{\mbox{\scriptsize{B}}}^N = \max \{m| \mbox{RE}_{N,m}<\varepsilon\}$.
In DKRR($\ell$) with $N$ training samples and $\ell$ communications, we denote the maximum number $m$ of
local machines that satisfies $\mbox{REC}_{N,m,\ell}<\varepsilon$ as $\hat{m}_{\mbox{\scriptsize{B}}}^{N,\ell}$, i.e.,
$\hat{m}_{\mbox{\scriptsize{B}}}^{N,\ell} = \max \{m| \mbox{REC}_{N,m,\ell}<\varepsilon\}$.
In the experiments, $\varepsilon$ is set as $0.05$. The computing capability of each local machine is assumed to be the same
for simplicity.}

 {We first generate 20000 training samples and 1000 testing samples. The number $m$ varies from $\{20,40,60,\cdots,600\}$ for the 1-dimensional data, and varies from $\{2,4,6,\cdots,60\}$ for the 3-dimensional data. The training time of DKRR with the number $m$ changing from the minimum number (20 for 1-dimensional data and 2 for 3-dimensional data) to $\bar{m}_{\mbox{\scriptsize{B}}}^N$
and the training time of DKRR($\ell$) with the number $m$ changing from the minimum number to $\hat{m}_{\mbox{\scriptsize{B}}}^{N,\ell}$ by different numbers
$\ell\in\{4,8,12,16,20\}$ are shown in Figure \ref{TrTime20000}.
The following observations can be made from these results. 1) Because there are
no calculations for communication steps, DKRR naturally consumes the least time when $m<\bar{m}_{\mbox{\scriptsize{B}}}^N$. 2) For the two simulation datasets, we have $\bar{m}_{\mbox{\scriptsize{B}}}^N \ll \hat{m}_{\mbox{\scriptsize{B}}}^{N,\ell}$, resulting that DKRR($\ell$) can further reduce the least training time of DKRR with the maximum number $\bar{m}_{\mbox{\scriptsize{B}}}^N$ as the number $m$ approaches to $\hat{m}_{\mbox{\scriptsize{B}}}^{N,\ell}$. However, the improvement is very limited for the 1-dimensional data, because the number $\bar{m}_{\mbox{\scriptsize{B}}}^N$ of DKRR is sufficiently large ($\bar{m}_{\mbox{\scriptsize{B}}}^N = 120$), and thus the marginal utility for the reduction of training time is very small when DKRR($\ell$) further enlarges the maximum number of local machines. 3) With the number $m$ increasing, the training time curves of DKRR($\ell$) ($\ell=4,8,16,32$) first decrease quickly and then increase slowly for the 1-dimensional data, and monotonously decrease for the 3-dimensional data. This phenomenon coincides with the
aforementioned theoretical analysis of training complexity, because $\hat{m}_{\mbox{\scriptsize{B}}}^{N,\ell} \approx 450 > \sqrt{N}$ for the 1-dimensional
data, and $\hat{m}_{\mbox{\scriptsize{B}}}^{N,\ell} \approx 50 < \sqrt{N}$ for the 3-dimensional data. In practical applications, we estimate the optimal number of local machines by the formula
\begin{equation} \label{hatmstar}
\hat{m}^\star_{N,\ell} = \left\{
\begin{array}{ll}
 m^\star, \qquad & \mbox{if} \quad \hat{m}_{\mbox{\scriptsize{B}}}^{N,\ell}>m^\star, \\
\hat{m}_{\mbox{\scriptsize{B}}}^{N,\ell}, \qquad & \mbox{otherwise}
\end{array}
\right.
\end{equation}
to expect it to have the lowest time consumption. The estimated $m^\star$ and the corresponding training time for DKRR($\ell$) with $\ell=2,4,8,16,32$ on the 1-dimensional data are marked by red pentagrams in the left figure of Figure \ref{TrTime20000}. It can be seen that the training time with the estimated number $m^\star$ is very close to the minimum of training time, which validates the correctness and reliability of equation (\ref{hatmstar}).}
% ---------Figure 9----------------
% ---------The number m_B----------------
\begin{figure*}[t]
    \centering
	 \subfigure{\includegraphics[width=6cm,height=5cm]{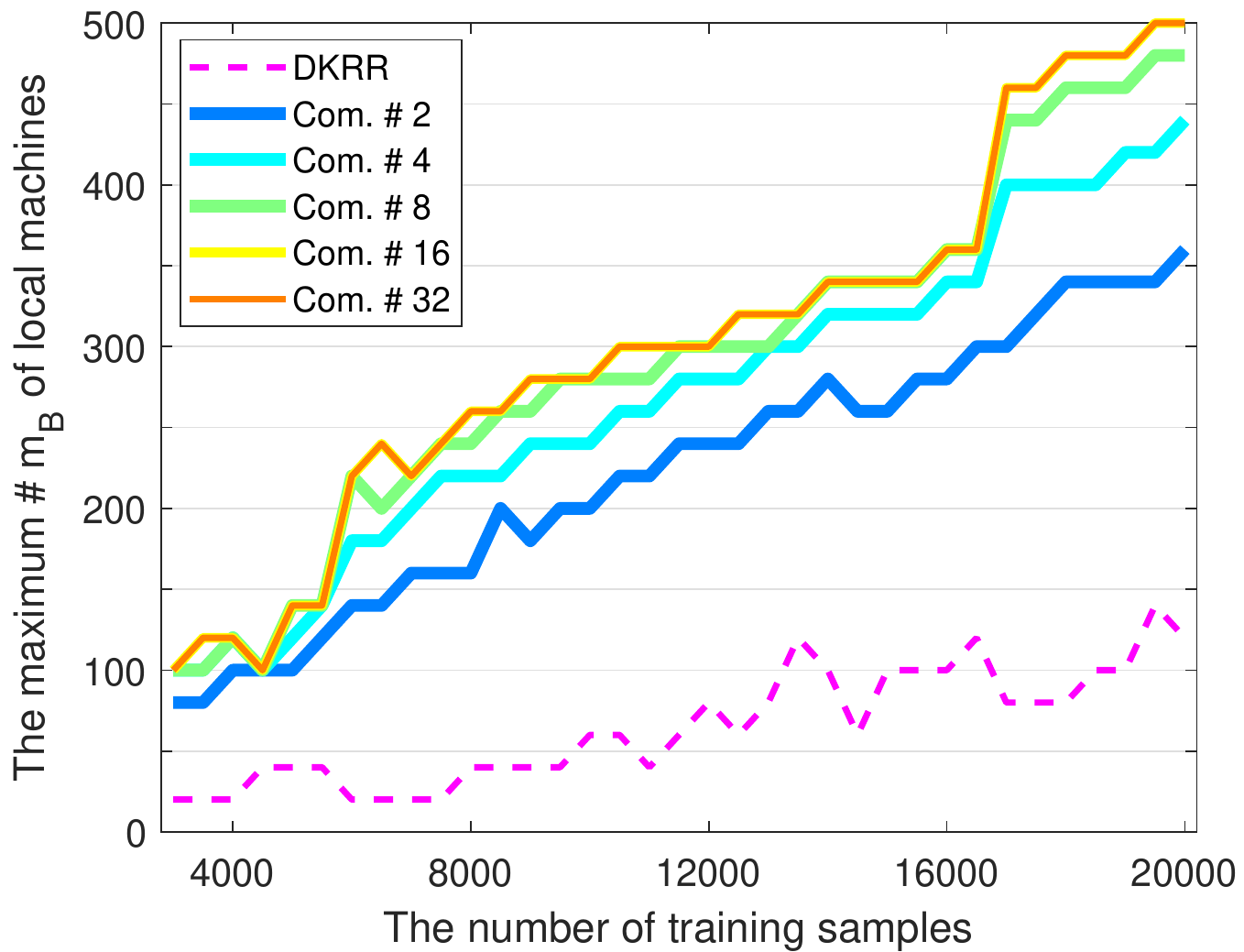}}
    \subfigure{\includegraphics[width=6cm,height=5cm]{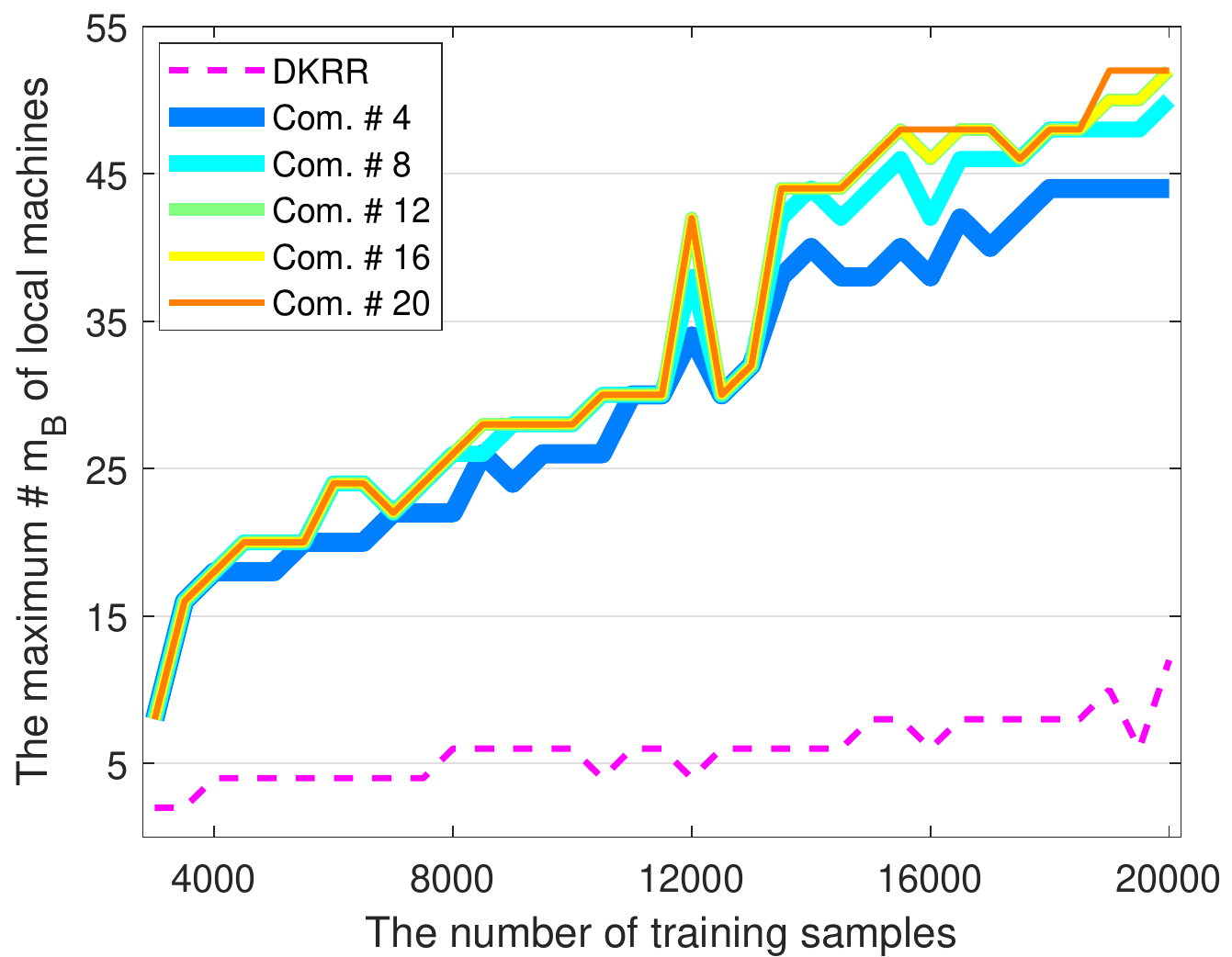}}
	\caption{ {1,0,0}{The relation between the number of training samples and the maximum number of local machines. The left figure and right figure
are respectively the results on the 1-dimensional data and the 3-dimensional data.}}\label{mB}
\end{figure*}
% ---------Figure 10----------------
% ---------The training time ----------------
\begin{figure*}[t]
    \centering
	 \subfigure{\includegraphics[width=6cm,height=5cm]{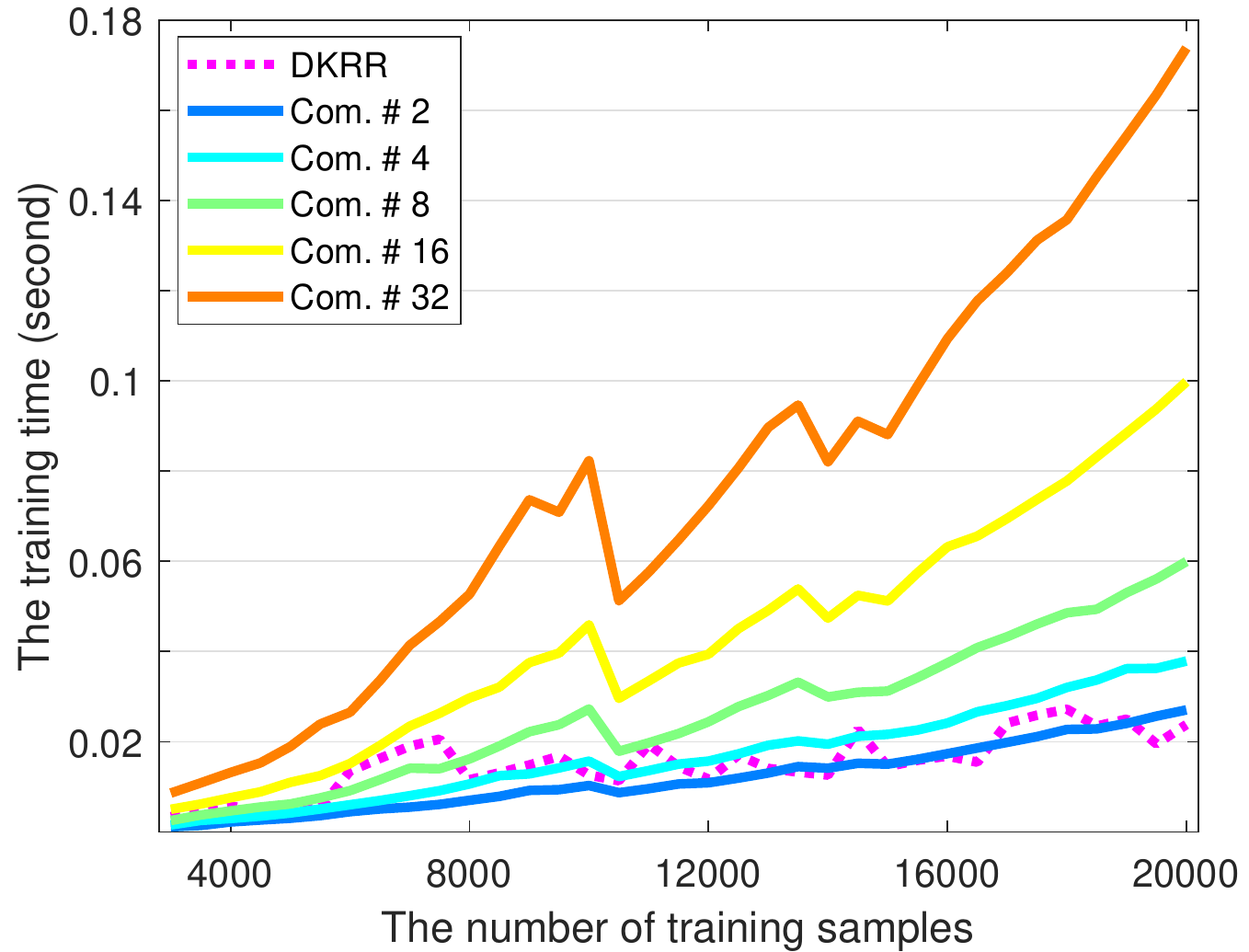}}
    \subfigure{\includegraphics[width=6cm,height=5cm]{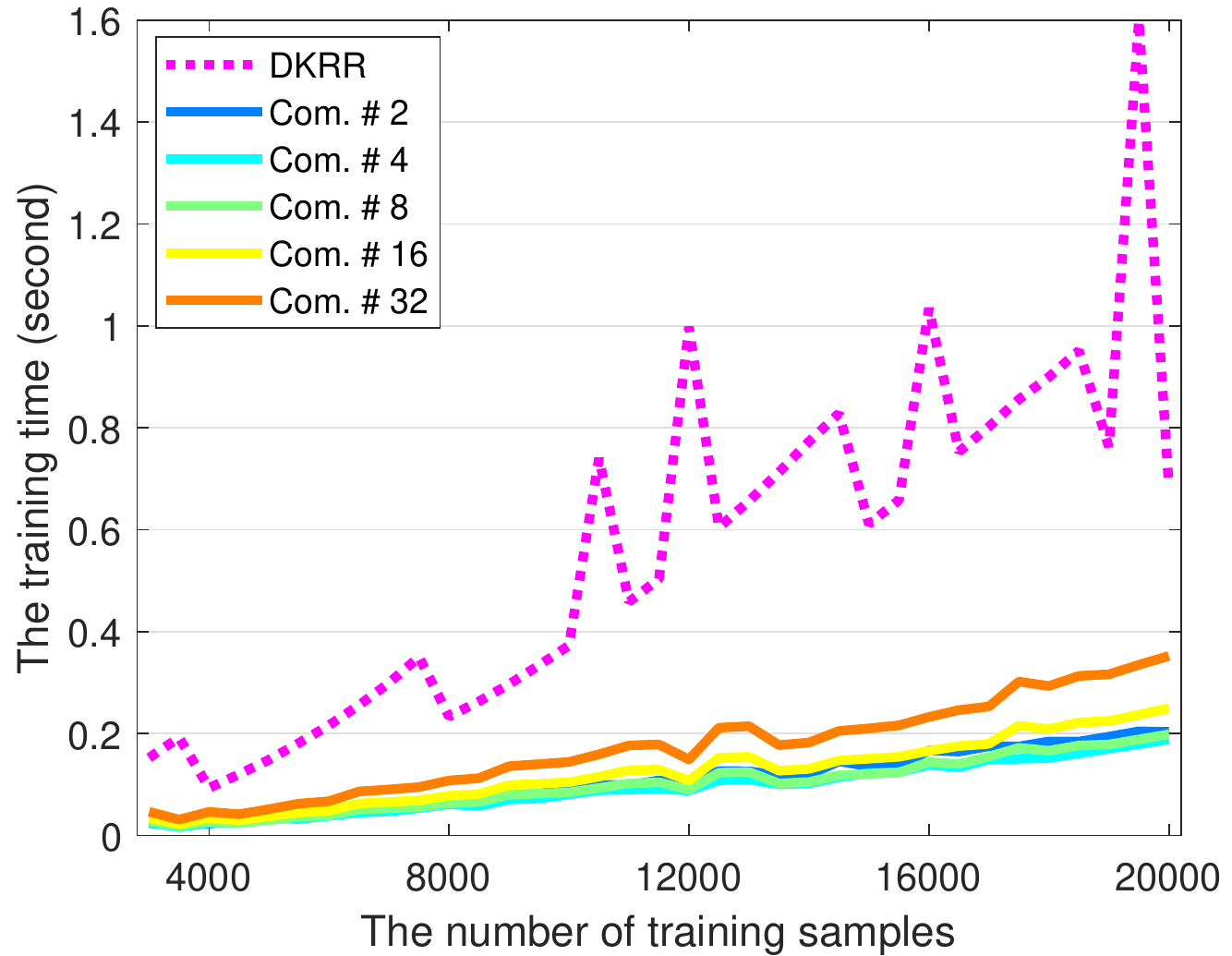}}
	\caption{ {The relation between the number of training samples and the training time with the optimal number of local machines. The left figure and right figure are respectively the results on the 1-dimensional data and the 3-dimensional data.}}\label{TrTime}
\end{figure*}

 {We also record the results with varying the number of training samples $N\in \{3000,3500,$ $4000,\cdots,20000\}$ and fixing the
number of testing samples $N'=1000$. Figure \ref{mB} shows the maximum numbers $\bar{m}_{\mbox{\scriptsize{B}}}^N$ of DKRR and
the maximum numbers $\hat{m}_{\mbox{\scriptsize{B}}}^{N,\ell}$ of DKRR($\ell$) with varying the numbers of training samples and communications, and Figure \ref{TrTime}
shows the training time of DKRR with $\bar{m}_{\mbox{\scriptsize{B}}}^N$ local machines and the training time of DKRR($\ell$) with $\hat{m}^\star_{N,\ell}$ local machines.
From these results, we have the following observations. 1) The maximum number of local machines increases as the number of training samples increases
for both DKRR and DKRR($\ell$), but the growth of DKRR($\ell$) is faster than that of DKRR.
2) The maximum number $\hat{m}_{\mbox{\scriptsize{B}}}^{N,\ell}$ of DKRR($\ell$) is robust to the number $\ell$ of communications when $\ell>4$. This verifies the fast convergence as shown in Figure \ref{Com_MSE}, and also implies that DKRR($\ell$) can achieve a satisfactory generalization
performance in a small number of communications. 3) For the 1-dimensional data, DKRR($\ell$) does not have much advantage in the training time, and is even worse than DKRR when the number of training samples $N\geq 10000$ and the number of communications $\ell\geq4$. Furthermore, the training time increases significantly as the number of communications increases. The main reason has two folds. On one hand, the marginal utility of enlarging the number of local machines diminishes when $\bar{m}_{\mbox{\scriptsize{B}}}^N$ of DKRR is large enough to efficiently deal with large-scale data. On the other hand, the training complexity of DKRR($\ell$) mainly focuses on the communications when the number $m$ is sufficiently large.
4) For the 3-dimensional data, the number $\bar{m}_{\mbox{\scriptsize{B}}}^N$ of DKRR is relatively small (e.g., $\bar{m}_{\mbox{\scriptsize{B}}}^N=12$ for $N=20000$ training samples). DKRR($\ell$) shows the significantly superior performance on training time when compared to DKRR, even for the large number $\ell=32$. This is the main focus of this paper, on an enlargement of the maximum number of local machines guaranteeing optimal learning rates when DKRR has a small number $\bar{m}_{\mbox{\scriptsize{B}}}^N$ of local machines.}

All these simulations verify our theoretical statements  and show the power
of  communications in distributed  learning.

\section*{Appendix A: Proofs of Lemmas in Section \ref{Sec.Operator representations}}

We use a new concentration inequality for positive operators
\citep{Minsker2017}, which was a refined estimate for the well known
Bernstein inequality for matrix \citep{Tropp2015}. This lemma has
been adopted in \citep{Dicker2017} and \citep{Guo2018} to derive
optimal learning rates for kernel-based spectral algorithms and
coefficient-regularization algorithms.

\begin{lemma}\label{Lemma:operator concerntration}
Let $R>0$ be a positive real constant and consider a finite sequence
of self-adjoint Hilbert-Schmidt operators $\{\xi_i\}_{i=1}^n$
satisfying $E[\xi_i]=0$ and $\|\xi_i\|\leq R$ almost surely. Suppose
there are constants $V,W>0$ such that
$\|E[(\sum_{i=1}^n\xi_i)^2]\|\leq V$ and
$\mbox{Tr}(E[(\sum_{i=1}^n\xi_i)^2])\leq VW$. For all $t\geq V^{1/2}
+R/3$,
$$
     P\left(\left\|\sum_{i=1}^n\xi_i\right\|\geq t\right)\leq
      4W\exp\left(-\frac{t^2}{2(V+Rt/3)}\right).
$$
\end{lemma}

{\bf Proof of Lemma \ref{Lemma:operator difference}.} Define the
random variable
$$
       \eta(x)=(L_K+\lambda I)^{-1/2}K_x\otimes K_x(L_K+\lambda
       I)^{-1/2}, \qquad x\in\mathcal X.
$$
It is easy to see that $\eta(x)$ is a self-adjoint operator for
 $x\in\mathcal X$. Furthermore,
$$
     E[\eta]=(L_K+\lambda I)^{-1/2}L_K(L_K+\lambda I)^{-1/2},
$$
and
$$
     \frac1{|D|}\sum_{(x,y)\in D} \eta(x)=(L_K+\lambda I)^{-1/2}L_{K,D}(L_K+\lambda
     I)^{-1/2}.
$$
Set
$$
      \xi(x)=\eta(x)- E[\eta],\qquad x\in\mathcal X.
$$
Then, we have $E[\xi]=0$.  If
$\{\sigma_\ell,\psi_\ell\}_{\ell\geq1}$ are  normalized
eigenparis of the integral operator $L_K$ on $L_{\rho_X}^2$, then
$\|\sqrt{\sigma_\ell}\psi_\ell\|_K=1$ when $\sigma_\ell>0$. Thus,
$\{\sqrt{\sigma_\ell}\psi_\ell:\sigma_\ell>0\}$ forms an orthonormal
basis of $\mathcal H_K$. This implies that
\begin{eqnarray*}
     \|\eta(x)\|^2
     &=&
     \sup_{\|f\|_K=1}\|\eta(x)f\|_K^2
     \leq
     \sum_{\ell}\|(L_K+\lambda I)^{-1/2}K_x\otimes K_x(L_K+\lambda
       I)^{-1/2}\sqrt{\sigma_\ell}\psi_\ell\|_K^2\\
       &=&
       \sum_{\ell}
       \frac{\sigma_\ell\psi_\ell^2(x)}{\lambda+\sigma_\ell}\|(L_K+\lambda
       I)^{-1/2}K_x\|_K^2=\left[\sum_{\ell}\frac{\sigma_\ell\psi_\ell^2(x)
       }{\lambda+\sigma_\ell}\right]^2.
\end{eqnarray*}
Then,
$$
  \|\xi(x)\|= \|\eta(x)- E[\eta]\|\leq  \max_{x\in\mathcal
  X}\|\eta(x)\|+1\leq \max_{x\in\mathcal
  X}\sum_{\ell}\frac{\sigma_\ell\psi_\ell^2(x)
       }{\lambda+\sigma_\ell}+1\leq  (\kappa^2+1)\lambda^{-1},
$$
where we use
$
     \|E[\eta]\|=\|(L_K+\lambda I)^{-1}L_K\|\leq 1
$ and $0<\lambda\leq 1$.
 For an arbitrary $f\in\mathcal H_K$, it follows from $[\eta(x)]^2f=\eta(x)[\eta(x)f]$ that
$$
    [\eta(x)]^2=\langle K_x,(L_K+\lambda I)^{-1}K_x\rangle_K\eta(x)
    =\sum_\ell\frac{\sigma_\ell\psi_\ell^2(x)}{\lambda+\sigma_\ell}\eta(x).
$$
Then, we get from
$\xi(x)=\eta(x)- E[\eta]$
that
\begin{eqnarray*}
   &&\left\|E\left[\left(\sum_{(x,y)\in
   D}\xi(x)\right)^2\right]\right\|\leq
   |D|(\|E[(\eta)^2]\|+\|(E[\eta])^2\|)    \nonumber\\
   &\leq&
   |D|(\kappa^2\lambda^{-1}\|E[\eta]\|+\|E[\eta]\|^2)
   \leq
    |D|\lambda^{-1}(\kappa^2+1).
\end{eqnarray*}
Moreover,
\begin{eqnarray*}
   &&\mbox{Tr}\left(E\left[\left(\sum_{(x,y)\in
   D}\xi(x)\right)^2\right]\right)\leq
   |D|Tr\left(E[(\eta)^2]-(E[\eta])^2  \right)   \nonumber\\
   &\leq&
   |D|(\kappa^2\lambda^{-1}\mbox{Tr}(E[\eta])+\mbox{Tr}(E[\eta])^2)
   \leq
    |D|\lambda^{-1}\mathcal N(\lambda)(\kappa^2+1).
\end{eqnarray*}
Plugging $R=\lambda^{-1}(\kappa^2+1)$,
$V=|D|\lambda^{-1}(\kappa^2+1)$ and $W=\mathcal N(\lambda)$, we have
that for all $ t\geq
\frac{\sqrt{(\kappa^2+1)}}{\sqrt{\lambda|D|}}+\frac{\kappa^2+1}{3\lambda|D|}$,
there holds
$$
      P[\mathcal R_{D,\lambda}\geq t]
     =
     P\left[\left\|\sum_{(x,y)\in D}\xi(x)\right\|\geq |D|t\right]
     \leq 4\mathcal
     N(\lambda)\exp\left(-\frac{\lambda|D|t^2}{2(\kappa^2+1)+(\kappa^2+1)t/3}\right).
$$
 Set
$4\mathcal
         N(\lambda)\exp\left\{-\frac{\lambda|D|\varepsilon^2}{(\kappa^2+1)(2+\varepsilon/3)}\right\}=\delta$.
 With confidence $1-\delta$,
$$
      \varepsilon\leq \frac{(\kappa^2+1)\log\frac{4\mathcal
      N(\lambda)}{\delta}}{3\lambda|D|}+ \sqrt{\frac{2(\kappa^2+1)\log\frac{4\mathcal
      N(\lambda)}{\delta}}{\lambda|D|}}.
$$
Setting
$$
    t=\frac{(\kappa^2+1) \log\frac{4\mathcal
      N(\lambda)}{\delta}}{3\lambda|D|}+ \sqrt{\frac{2(\kappa^2+1)\log\frac{4\mathcal
      N(\lambda)}{\delta}}{\lambda|D|}}\geq
      \frac{\sqrt{(\kappa^2+1)}}{\sqrt{\lambda|D|}}+\frac{\kappa^2+1}{3\lambda|D|},
$$
 we then get that    (\ref{operator difference concentration 1}) holds with confidence $1-\delta$. This
completes the proof of Lemma \ref{Lemma:operator difference}. $\Box$

We then prove Lemma \ref{Lemma:operator product} by using Lemma
\ref{Lemma:operator difference}.

{\bf Proof of Lemma \ref{Lemma:operator product}.}  Since
$\delta\geq 4\exp\{2C_1^*\mathcal B_{D,\lambda}\}$, it follows from
Lemma \ref{Lemma:operator difference} that $
         \mathcal R_{D,\lambda}<\frac12
$ holds with confidence $1-\delta$. Then,
\begin{eqnarray*}
      &&(L_K+\lambda I)^{1/2}(L_{K,D}+\lambda
         I)^{-1}(L_K+\lambda I)^{1/2}\\
         &=&
         (L_K+\lambda I)^{1/2}[(L_{K,D}+\lambda
         I)^{-1}-(L_{K}+\lambda
         I)^{-1}]   (L_K+\lambda I)^{1/2}+I
          = I\\
          &+&
         (L_K+\lambda I)^{-1/2}(L_K-L_{K,D})(L_K+\lambda I)^{-1/2}(L_K+\lambda I)^{1/2}(L_{K,D}+\lambda
         I)^{-1}(L_K+\lambda I)^{1/2}.
\end{eqnarray*}
Thus,
\begin{eqnarray*}
   && \|(L_K+\lambda I)^{1/2}(L_{K,D}+\lambda
         I)^{-1}(L_K+\lambda I)^{1/2}\|\\
         &\leq&
         1+ \frac12\|(L_K+\lambda I)^{1/2}(L_{K,D}+\lambda
         I)^{-1}(L_K+\lambda I)^{1/2}\|.
\end{eqnarray*}
This implies
$$
   \|(L_K+\lambda I)^{1/2}(L_{K,D}+\lambda
         I)^{-1}(L_K+\lambda I)^{1/2}\|\leq 2.
$$
Since $t^{1/2}$ is operator monotone \citep[Chap.4]{Bathia1997}, we
have
$$
   \mathcal Q_{D,\lambda}\leq \|[(L_K+\lambda I)^{1/2}(L_{K,D}+\lambda
         I)^{-1}(L_K+\lambda I)^{1/2}]^{1/2}\|\leq\sqrt{2}.
$$
  This completes the proof of Lemma
\ref{Lemma:operator product}. $\Box$

\section*{Appendix B:   training and testing flows for DKRR($\ell$) }

Generally speaking, it is difficult to communicate functions in practice. Thus, the implementation of DKRR($\ell$) requires communications of additional information. In this part, we numerically realize  DKRR($\ell$) by communicating the inputs of data for the sake of simplicity,  though there are numerous state-of-the-art approaches to approximate the gradient matrix via much less of communication loss.

{\bf Training Flow:}

Step 1 (local process). Run KRR (\ref{KRR}) on the $j$-th local
machine with data $D_j$ and communicate $D_j(x)=\{x:(x,y)\in D\}$ to the
$k$-th local machine for $k=1,\dots,m$. Store $m$ matrices of size
$|D_j|\times|D_k|$, $\mathbb K_{D_k,D_j}:=\{K(x,x')\}_{x\in
D_k(x),x'\in D_j(x)}$ with $k=1,\dots,m$, the matrix
$M_{D_j,\lambda}=(\mathbb K_{D_j,D_j}+\lambda |D_j| \mathbb
I_{D_j})^{-1}$, and the
 vector
$ \vec{\alpha}_{_{D_j}}=M_{D_j,\lambda}y_{D_j}$, where $\mathbb I$
is the unit matrix of size $|D_j|\times|D_j|$.

Step 2 (synthesization) On the $j$-th local machine, communicate $m$
vectors of size $|D_k|$, $K_{D_k,D_j}\vec{\alpha}_{_{D_j}}$ for
$k=1,\dots,m$ to the global machine. Synthesize $m$   global
    vectors
$\vec{f}_{D,\lambda,D_k}^0=\sum_{j=1}^m\frac{|D_j|}{|D|}\mathbb
K_{D_k,D_j}\vec{\alpha}_{_{D_j}}$ for $k=1,\dots,m$.

Step 3 (distributing) For  $\ell=1,2,\dots,$ distribute
$\vec{f}_{D,\lambda,D_j}^{\ell-1}$ to the $j$-th local machine.

Step 4 (local gradients) On the $j$-th local machine, compute $m$
gradient vectors $\vec{G}_{D_j,\lambda,k,\ell}=\frac{\mathbb
K_{D_k,D_j}}{|D_j|}(\overline{f}_{D,\lambda,D_j}^{\ell-1}-y_{D_j})+\lambda\vec{f}_{D,\lambda,D_k}^{\ell-1}$
for $k=1,\dots,m$ and communicate these $m$ vectors to the global
machine.

Step 5 (synthesizing gradients)  Synthesize $m$ global gradient
vectors,$\vec{G}_{D,\lambda,k,\ell}:=\sum_{j=1}^m\frac{|D_j|}{|D|}$ $\vec{G}_{D_j,\lambda,k,\ell}$
for $k=1,\dots,m$. Distribute $m$ vectors
$\vec{G}_{D,\lambda,k,\ell}$ with $k=1,\dots,m$ to each local
machine.

Step 6 (KRR on gradient data) On the $j$-th local machine,  compute
the vector
$\vec{\beta}_{j,\ell}=M_{D_j,\lambda}\vec{G}_{D,\lambda,j,\ell}$ and
 $m$ vectors $\vec{H}_{D_j,\lambda,k,\ell}:=
\vec{G}_{D,\lambda,k,\ell}
   -\mathbb K_{D_k,D_j}\vec{\beta}_{j,\ell}$ for $k=1,\dots,m$.
   Communicate $m$ vectors   $\vec{H}_{D_j,\lambda,k,\ell}$  with $k=1,\dots,m$ to the
global machine.

Step 7 (final estimator)  Generate $m$ vectors
$$
   \vec{f}_{D,\lambda,D_k}^{\ell}=\vec{f}_{D,\lambda,D_k}^{\ell-1}
   -\frac1\lambda\sum_{j=1}^m\frac{|D_j|}{|D|}\vec{H}_{D_j,\lambda,k,\ell},\qquad
   k=1,\dots,m
$$
and transmit $\vec{f}_{D,\lambda,D_j}^{\ell}$ to the $j$-th local
machine with $j=1,\dots,m$.

The above 7 steps provide a realization for DKRR($\ell$). However, it is difficult to get all  input  data in a local machine, due to the loss of communications.  In particular, input data of some local machines are unaccessible due to the data privacy. Under this circumstance, these local machines are only used to build $\overline{f}_{D,\lambda}^0$ and are excluded in building $\overline{f}_{D,\lambda}^\ell$ for $\ell\geq 1$. This provides a numerical realization for the algorithm in Simulation 3 in Section \ref{Sec.Experiments}.

{\bf Testing Flow:} given vector $D'(x):=(x_1',\dots,x_{|D'|}) $ which
consists  of $|D'|$ query points.

Step 1 (local estimates) Transmit $D'(x)$ to $m$ local machines. On
the $j$-th local machine, store vectors of size $|D'|$,
$\vec{K}_{D',D_j}\vec{\alpha}_{_{D_j}} $.

Step 2 (global estimates) Compute test vector of size $|D'|$,
$$
       \overline{f}_{D,\lambda}^0(D'):=\sum_{j=1}^m\frac{|D_j|}{|D|}\vec{K}_{D',D_j}\vec{\alpha}_{_{D_j}} .
$$

Step 3 (local gradients) For $\ell=1,2,\dots,$ distribute
$\overline{f}_{D,\lambda}^{\ell-1}(D')$ to $m$ local machines, and
compute
$$
      G_{D_j,\lambda,\ell}(D'):=\frac{\vec{K}_{D',D_j}}{|D_j|}(\vec{f}_{D,\lambda,D_j}^{\ell-1}-y_{D_j})
      +\lambda\overline{f}_{D,\lambda}^{\ell-1}(D'),
$$
where $\vec{f}_{D,\lambda,D_j}^{\ell-1}$ is obtained from the
training flow.

Step 4 (global gradients) Transmit $G_{D_j,\lambda,\ell}(D')$ with
$j=1,\dots,m$ to the global machine and get the global gradient
vector
$$
     G_{D,\lambda,\ell}(D')=\sum_{j=1}^m\frac{|D_j|}{|D|}G_{D_j,\lambda,\ell}(D').
$$

Step 5 (final estimator) Generate the vector of estimators
$$
    \overline{f}_{D,\lambda}^{\ell}(D')
    =
     \overline{f}_{D,\lambda}^{\ell-1}(D')-\frac1\lambda\sum_{j=1}^m\frac{|D_j|}{|D|}
     \left[G_{D,\lambda,\ell}(D')-\vec{K}_{D',D_j}\vec{\beta}_{j,\ell}\right].
$$

\section*{Acknowledgement}
The work of Shao-Bo Lin is supported
partially by the National Natural Science Foundation of China (Nos. 61876133, 11771021).
The work of Di Wang is supported partially by the National Natural Science Foundation of China (Grant No. 61772374)
and by Zhejiang Provincial Natural Science Foundation (Grants No. LY17F030004).
The work of Ding-Xuan Zhou is supported partially by the Research Grants Council
of Hong Kong [Project No. CityU 11306617] and by National Natural Science Foundation of China under Grant 11461161006.


\begin{thebibliography}{31}
\providecommand{\natexlab}[1]{#1}
\providecommand{\url}[1]{\texttt{#1}}
\expandafter\ifx\csname urlstyle\endcsname\relax
  \providecommand{\doi}[1]{doi: #1}\else
  \providecommand{\doi}{doi: \begingroup \urlstyle{rm}\Url}\fi

{ 
\bibitem[Bathis(1997)]{Bathia1997}
R.~Bathis.
\newblock \emph{Matrix {A}nalysis, Volume 169 of Graduate Texts in Mathematics}.
\newblock Springer, Berlin, 1997.
}

\bibitem[Bellet et~al.(2015)Bellet, Liang, Garakani, Balcan, and Sha]{Bellet2015}
A.~Bellet, Y.~Liang, A.~B.~Garakani, M.~F.~Balcan, and F.~Sha.
\newblock A distributed frank-wolfe algorithm for communication-efficient sparse learning.
\newblock \emph{In Proceedings of the fifteenth SIAM International Conference on Data Mining, Society for Industrial and
Applied Mathematics}, 478-486, 2015.

{ 
\bibitem[Blanchard and Kr{\"a}mer(2016)]{Blanchard2016}
G.~Blanchard and N.~Kr{\"a}mer.
\newblock Convergence rates of kernel conjugate gradient for random design regression.
\newblock \emph{Anal. Appl.}, 14:\penalty0 763-794, 2016.
}


{ 
\bibitem[Caponnetto and DeVito(2007)]{Caponnetto2007}
A.~Caponnetto and E.~DeVito.
\newblock Optimal rates for the regularized least squares algorithm.
\newblock \emph{Found. Comput. Math.}, 7:\penalty0 331-368, 2007.
}

\bibitem[Chang et~al.(2017)Chang, Lin, and Zhou]{Chang2017}
X.~Chang, S.~B.~Lin,  and D.~X.~Zhou.
\newblock Distributed semi-supervised learning with kernel ridge regression.
\newblock \emph{J. Mach. Learn. Res.}, 18\penalty0 (46):\penalty0 1-22, 2017.

\bibitem[Chang et~al.(2017)Chang, Lin, and Wang]{Chang2017a}
X.~Chang, S.~B.~Lin, and Y.~Wang.
\newblock Divide and conquer local average regression.
\newblock \emph{Electronic J. Stat.}, 11:\penalty0 1326-1350, 2017.


\bibitem[Cucker and Zhou(2007)]{Cucker2007}
F. Cucker and D. X. Zhou.
\newblock Learning {T}heory: {A}n {A}pproximation {T}heory {V}iewpoint.
\newblock Cambridge University Press, Cambridge, 2007.

\bibitem[Guo et~al.(2017)Guo, Lin, and Zhou]{Guo2017}
Z.~C.~Guo, S.~B.~Lin, and D.~X.~Zhou.
\newblock Learning theory of distributed spectral algorithms.
\newblock \emph{Inverse Problems}, 33:\penalty0 074009, 2017.


\bibitem[Guo and Shi(2019)]{Guo2018}
Z.~C.~Guo and L.~Shi.
\newblock Optimal rates for coefficient-based regularized regression.
\newblock \emph{Appl. Comput. Harmonic Anal.},  47: 662-701, 2019.

\bibitem[Guo et~al.(2019)Guo, Lin, and Shi]{Guo2019a}
Z.~C.~Guo, S.~B.~Lin, and L.~Shi.
\newblock Distributed learning with multi-penalty regularization.
\newblock \emph{Appl. Comput. Harmonic Anal.}, 46:\penalty0 478-499, 2019.


\bibitem[Dicker et~al.(2017)Dicker, Foster, and Hsu]{Dicker2017}
L.~H.~Dicker, D.~P.~Foster, and D.~Hsu.
\newblock Kernel ridge vs. principal component regression: Minimax bounds and the qualification of regularization operators.
\newblock \emph{Electron. J. Stat.}, 11:\penalty0 1022-1047, 2017.


\bibitem[Dudley(2002)]{Dudley2002}
R.~M.~Dudley.
\newblock Real Analysis and Probability, Cambridge Studies in Advanced Mathematics {\bf 74}.
\newblock Cambridge University Press, Cambridge, 2002.

\bibitem[Gy\"{o}rfy et~al.(2002)Gy\"{o}rfy, Kohler, Krzyzak, and Walk]{Gyorfi2002}
L.~Gy\"{o}rfy, M.~Kohler, A.~Krzyzak, and H.~Walk.
\newblock A Distribution-Free Theory of Nonparametric Regression.
\newblock Springer-Verlag, Berlin, 2002.


\bibitem[Hu et~al.(2019)Hu, Wu, and Zhou]{Hu2019}
T.~Hu, Q.~Wu, and D.~X.~Zhou.
\newblock Distributed kernel gradient descent algorithm for minimum error entropy principle.
\newblock \emph{Appl. Comput. Harmon. Anal.}, In Press, 2019.



\bibitem[Huang and Huo(2017)]{Huang2015}
C.~Huang and X.~M.~Huo.
\newblock A distributed one-step estimator.
\newblock arXiv preprint arXiv:1511.01443, 2015.


\bibitem[Li et~al.(2014)Li, Andersen, Smola, and Yu]{Li2014}
M.~Li, D.~G.~Andersen, A.~J.~Smola, and K.~Yu.
\newblock Communication efficient distributed machine learning with the parameter server.
\newblock \emph{ Advances in Neural Information Processing Systems}, 19-27, 2014.


{ 
\bibitem[Lin et~al.(2018)Lin, Rudi, Rosasco, and Cevher]{Linj2018}
J.~Lin, A.~Rudi, L.~Rosasco and V.~Cevher.
\newblock  Optimal rates for spectral algorithms with least-squares regression over Hilbert spaces.
\newblock \emph{Appl. Comput. Harmonic Anal.}, In Press, 2018.
}

\bibitem[Lin et~al.(2017)Lin, Guo, and Zhou]{Lin2017}
S.~B.~Lin, X.~Guo, and D.~X.~Zhou.
\newblock Distributed learning with regularized least squares.
\newblock \emph{J. Mach. Learn. Res.}, 18\penalty0 (92):\penalty0  1-31, 2017.


\bibitem[Lin and Zhou(2018)]{Lin2018}
S.~B.~Lin and D.~X.~Zhou.
\newblock Distributed kernel-based gradient descent algorithms.
\newblock \emph{Constr. Approx.}, 47:\penalty0 249-276, 2018.

\bibitem[Lin et~al.(2019)Lin, Lei, and Zhou]{Lin2019}
S.~B.~Lin, Y.~Lei, and D.~X.~Zhou.
\newblock  Boosted kernel ridge regression: optimal learning rates and early stopping.
\newblock \emph{J.  Mach. Learn. Res.}, 20:\penalty0  1-36, 2019.




\bibitem[Liu et~al.(2018)Liu, Shang, and Cheng]{Liu2018}
M.~Liu, Z.~Shang, and G.~Cheng.
\newblock  How many machines can we use in parallel computing for kernel ridge regression?
\newblock arXiv preprint arXiv:1805.09948, 2018.

\bibitem[Mann et~al.(2009)Mann, McDonald, Mohri, Silberman, and Walker]{Mann2009}
G.~Mann, R.~McDonald, M.~Mohri, N.~Silberman, and D.~Walker.
\newblock Efficient large-scale distributed training of conditional maximum entropy models.
\newblock \emph{ Advances in Neural Information Processing Systems}, 19-27, 2009.


\bibitem[Minsker(2017)]{Minsker2017}
S.~Minsker.
\newblock On some extensions of Bernstein's inequality for self-adjoint operators.
\newblock \emph{Stat. Prob. Letters}, 127:\penalty0 111-119, 2017.

{ 
\bibitem[M\"{u}cke and Blanchard(2018)]{Blanchard2018}
N.~M\"{u}cke and G.~Blanchard.
\newblock Parallelizing spectrally regularized kernel algorithms.
\newblock \emph{J. Mach. Learn. Res.}, 19:\penalty0 1-29, 2018.
}

{ 
\bibitem[Pang and Sun(2018)]{Pang2018}
M.~Pang and H.~Sun.
\newblock Distributed regression learning with coefficient regularization.
\newblock \emph{J. Math. Anal. Appl.}, 466\penalty0 (1):\penalty0  676-689, 2018.}

\bibitem[Pinelis(1994)]{Pinelis1994}
I.~Pinelis.
\newblock Optimum bounds for the distributions of martingales in Banach spaces.
\newblock \emph{Ann. Prob.}, 22:\penalty0 1679-1706, 1994.



%\bibitem{Pinkus1985}
%A. Pinkus. $n$-Widths in Approximation Theory.
%  Springer-Velag, Berlin Heidelberg, 1985.


\bibitem[Schaback and Wendland(2006)]{Schaback2006}
R.~Schaback and H.~Wendland.
\newblock Kernel techniques: from machine learning to mesheless methods.
\newblock \emph{Acta Numerica}, 15:\penalty0  543-639, 2006.

\bibitem[Shamir et~al.(2014)Shamir, Srebro, and Zhang]{Shamir2014}
O.~Shamir, N.~Srebro, and T.~Zhang.
\newblock Communication-efficient distributed optimization using an approximate newton-type method.
\newblock \emph{In Proceedings of the 31st International Conference on International Conference on Machine Learning}, 32:\penalty0 1000-1008, 2014.


\bibitem[Shang and Cheng(2017)]{Shang2017}
Z.~Shang and G.~Cheng.
\newblock Computational limits of a distributed algorithm for smoothing spline.
\newblock \emph{J.  Mach. Learn. Res.}, 18:\penalty0  3809-3845, 2017.

{ 
\bibitem[Shi(2019)]{Shi2018}
L.~Shi.
\newblock Distributed learning with indefinite kernels.
\newblock \emph{Anal. Appl.}, In Press, 2019.
}

\bibitem[Smale and Zhou(2007)]{Smale2007}
S.~Smale and D.~X.~Zhou.
\newblock Learning theory estimates via integral operators and their approximations.
\newblock \emph{Constr. Approx.}, 26:\penalty0  153-172, 2007.


\bibitem[Steinwart et~al.(2009)Steinwart, Hush, and Scovel]{SteinwartHS}
I.~Steinwart, D.~Hush, and C.~Scovel.
\newblock Optimal rates for regularized least squares regression.
\newblock In S.~Dasgupta and A.~Klivan, editors, \emph{Annual Conference on Learning Theory}, 79-93, 2009.


%  \bibitem{Steinwart2012}
%I. Steinwart, C. Scovel, Mercer's theorem on general domains: On the
%interaction between measures, kernels, and RKHSs, Constr. Approx.,
%35 (2012), 363-417.

\bibitem[Tropp(2015)]{Tropp2015}
J.~A.~Tropp.
\newblock An introduction to matrix conecentation inequalities.
\newblock \emph{Foundations and Trends in Machine Learning}, 8:\penalty0 1-230, 2015.


\bibitem[Wu(1995)]{Wu1995}
Z.~M.~Wu.
\newblock Compactly supported positive definite radial functions.
\newblock \emph{Adv. Comput. Math.}, 4:\penalty0 283-292, 1995.

\bibitem[Zeng and Ying(2018)]{Zeng2018}
J. Zeng and W. Ying.
\newblock On nonconvex decentralized gradient descent.
\newblock \emph{IEEE Trans. Signal Process}, 66:\penalty0  2834-2848, 2018.

\bibitem[Zhang et~al.(2013)Zhang, Duchi, and Wainwright]{Zhang2013}
Y.~C.~Zhang, J.~Duchi, and M.~Wainwright.
\newblock Communication-efficient algorithms for statistical optimization.
\newblock \emph{J. Mach. Learn. Res.}, 14:\penalty0 3321-3363, 2013.


\bibitem[Zhang et~al.(2015)Zhang, Duchi, and Wainwright]{Zhang2015}
Y.~C.~Zhang, J.~Duchi, and M.~Wainwright.
\newblock Divide and conquer kernel ridge regression: A distributed algorithm with minimax optimal rates.
\newblock \emph{J. Mach. Learn. Res.}, 16:\penalty0 3299-3340, 2015.

%\bibitem{Zhou2014}
%Z. H. Zhou, N. V. Chawla, Y. Jin,  G. J. Williams, Big data
%opportunities and challenges: Discussions from data analytics
%perspectives, IEEE Comput. Intel. Mag., 9 (2014), 62-74.
%
\bibitem[Zhou(2018a)]{Zhou2018Deep}
D.~X.~Zhou.
\newblock Deep distributed convolutional neural networks: Universality.
\newblock \emph{Anal. Appl.}, 16:\penalty0 895-919, 2018.

\bibitem[Zhou(2018b)]{Zhou2018Uni}
D.~X.~Zhou.
\newblock Universality of deep convolutional neural networks.
\newblock \emph{Appl. Comput. Harmonic Anal.}, 48:\penalty0 787-794, 2020.

\bibitem[Zhou(2018c)]{Zhou2018Dis}
D.~X.~Zhou.
\newblock Distributed approximation with deep convolutional neural networks.
\newblock preprint, 2018.

\end{thebibliography}
\end{document}